\documentclass[twoside,table,xcdraw]{article}

\usepackage[accepted]{aistats2023}
\usepackage{breakcites}

\usepackage[utf8]{inputenc} % allow utf-8 input
\usepackage[T1]{fontenc}    % use 8-bit T1 fonts
\usepackage[hidelinks]{hyperref}       % hyperlinks
\usepackage{url}            % simple URL typesetting
\usepackage{booktabs}       % professional-quality tables
\usepackage{amsfonts}       % blackboard math symbols
\usepackage{nicefrac}       % compact symbols for 1/2, etc.
\usepackage{microtype}      % microtypography
\usepackage{xcolor}         % colors
\usepackage{wrapfig}
\usepackage{enumitem}
\usepackage{microtype} 
\usepackage{marvosym} 
\usepackage[english]{babel}
\usepackage{bm}
\usepackage{subcaption}
\usepackage{multirow}
\usepackage{algorithm}
\usepackage{algpseudocode}
\usepackage{setspace}
\usepackage{pythonhighlight}
\usepackage{natbib}

\addto\captionsenglish{% Replace "english" with the language you use
  \renewcommand{\contentsname}%
    {Table of contents}%
}

\newtheorem{theorem}{Theorem}
\newtheorem{proposition}{Proposition}

\newtheorem{definition}{Definition}

\newtheorem{assumption}{Assumption}

\newcommand{\our}[0]{\textsc{Projector}~}
\newcommand*{\oure}{\textsc{Projector}}

\usepackage{tikz}
\newcommand*\circled[1]{\tikz[baseline=(char.base)]{
            \node[shape=circle,draw,inner sep=0.3pt] (char) {#1};}}
            
\begin{document}



\twocolumn[

\aistatstitle{Efficiently Forgetting What You Have Learned in Graph Representation Learning via Projection}

\aistatsauthor{ Weilin Cong \And Mehrdad Mahdavi}

\aistatsaddress{ The Pennsylvania State University \And  The Pennsylvania State University} ]

% \footnote{Linear-GNN is a graph neural network (GNN) without non-linearity and only has a single weight matrix, e.g., SGC.}

\begin{abstract}
As privacy protection receives much attention, unlearning the effect of a specific node from a pre-trained graph learning model has become equally important. However, due to the \emph{node dependency} in the graph-structured data,  representation unlearning in Graph Neural Networks (GNNs) is challenging and  less well explored. In this paper, we fill in this gap by first studying the unlearning problem in  linear-GNNs, and then introducing its extension to  non-linear structures. Given a set of nodes to unlearn, we propose \textsc{Projector} that unlearns by projecting the weight parameters of the pre-trained model onto a subspace that is irrelevant to features of the nodes to be forgotten. \textsc{Projector} could overcome the challenges caused by node dependency and enjoys a perfect data removal, i.e., the unlearned model parameters do not contain any information about the unlearned node features which is guaranteed by algorithmic construction. Empirical results on real-world datasets illustrate the effectiveness and efficiency of \textsc{Projector}.
[\href{https://github.com/CongWeilin/Projector}{\textcolor{blue}{Code}}].
\end{abstract}

\section{Introduction}
As graph representation learning has achieved great success in real-world applications (e.g., social networks~\cite{kipf2016semi,hamilton2017inductive}, knowledge graphs~\cite{wang2019knowledge,wang2019kgat}, and recommender system~\cite{berg2017graph}), privacy protection in graph representation learning has become equally important. Recently, as ``\emph{Right to be forgotten}'' gradually implemented in multiple jurisdictions, users are empowered with the right to request any organization or company to remove the effect of their private data from a machine learning model, which is known as ``\emph{machine unlearning}''. For example, when a Twitter user deletes a post, the user not only may require Twitter to permanently remove the post from their database, but also might require Twitter to eliminate its impact on any machine learning models pre-trained on the deleted post, so as to prevent the private information in the deleted post be inferred by any malicious third party. 

Existing unlearning approaches can be roughly classified into exact unlearning and approximate unlearning. The goal of ``\emph{exact unlearning}'' is to exactly produce the model parameters trained without the deleted data. The most straightforward unlearning approach is to retrain the model from scratch using the remaining data, which could be computationally prohibitive when the dataset is large or even infeasible if not all the data are available to retrain. To avoid re-training on large data,  \textsc{SISA}~\cite{bourtoule2021machine} proposes to split the original dataset into multiple shards and train a model on each data shard, then aggregate their prediction during inference. Upon receiving unlearning requirements, they only need to re-train the specific shard model that the unlearned data belongs to. While being more efficient compared to  retraining from scratch, the model performance suffers because each model has fewer data to be trained on and data heterogeneity also deteriorates the performance. To further reduce the computation overhead, ``\emph{approximate unlearning}'' is proposed to trade-off between the unlearning efficiency and the data removal effectiveness. For example,~\textsc{Influence}~\cite{guo2020certified} proposes to approximate the unlearned model using first-order Taylor approximation and~\textsc{Fisher}~\cite{golatkar2020eternal} proposes to directly fine-tune with Newton's method on the remaining data. Since approximate unlearning methods lack guarantee on whether all information associated with the deleted data is eliminated, it is necessary to inject random noise to model parameters or objective functions to amplify privacy, which could significantly hurt the  performance of unlearned model. Employing these methods in graph-structured data  is even more challenging due to the dependency among nodes. Motivated by the importance  of unlearning graph-structured data, we aim at answering the following questions in the context of GNNs:

\begin{figure*}[t]
    \centering
    \includegraphics[width=0.8\textwidth]{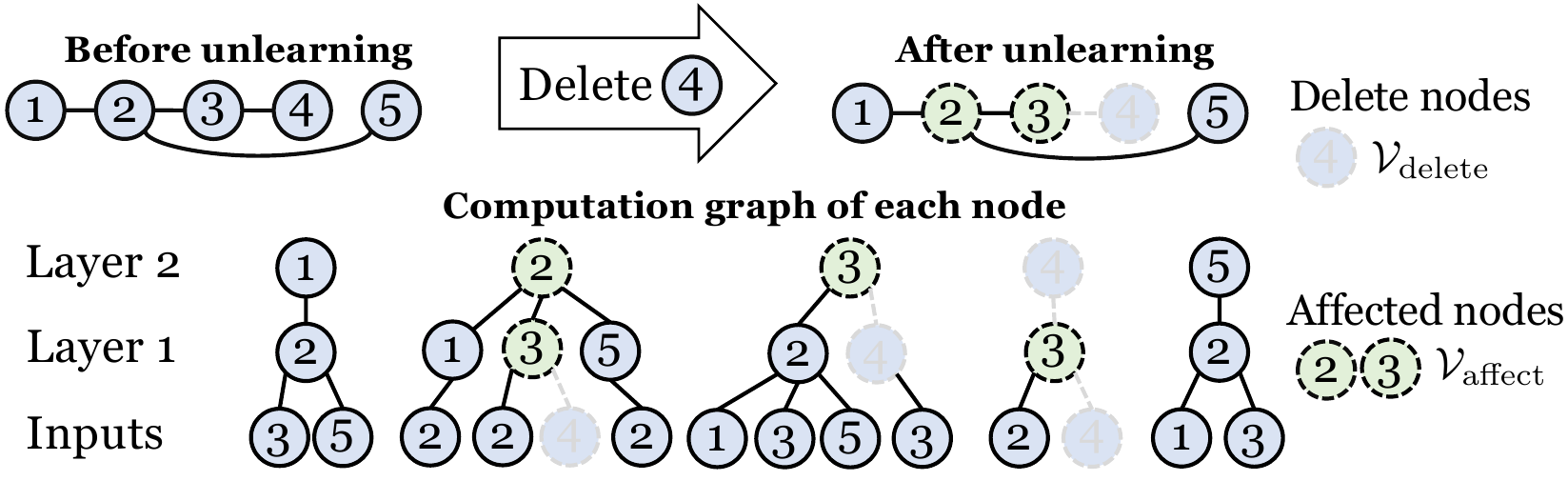}
    \vspace{-3mm}
    \caption{
    An illustration of how the node representations of a $2$-layer GNN (with neighbor average aggregation) are affected after deleting the node $v_4$ from the graph. After removing node $v_4$, the node representation of nodes $\{ v_2, v_3\}$ are also affected since these nodes require node $v_4$ to compute their representations.  Such dependency grows exponentially with respect to the number of GNN layers.
    }
    \label{fig:which_node_get_affected_combined}
    % \vspace{-3mm}
\end{figure*}

\noindent\textit{\textbf{Q1.~Can existing machine unlearning methods be utilized to solve graph unlearning problem?}}
Most of the existing methods are designed for  settings where the loss function can be decomposed over individual training samples, and the node dependency in graph-structured data render these methods inapplicable  to GNNs and makes them sub-optimal. For example, exact graph unlearning method \textsc{GraphEraser}~\cite{chen2021graph} extends~\cite{bourtoule2021machine} by partitioning the original graph into multiple subgraphs. However, graph partitioning will result in loosing part of the structure information due to ignorance of the edges that span subgraphs, which could further hurt the model performance. Moreover, the data heterogeneity issue on homophily graphs is more severe because nodes with similar properties/categories are more likely to be partitioned into the same subgraph. 
Applying approximate unlearning for graph structured data is also non-trivial.
For example, most of these methods require ``\emph{the objective function before data deletion}'' could be formulated as a summation of ``\emph{the objective after data deletion}''  and ``\emph{the loss on deleted data}''. However, this is not the case on graph-structured data because the representation of the deleted nodes' multi-hop neighbors $\mathcal{V}_\text{affect}$ are also affected after node deletion. Please refer to Figure~\ref{fig:which_node_get_affected_combined} on how node dependency would affect the GNN models' output after deleting a single node from graph, refer to Appendix~\ref{section:dependency_issue_in_finite_sum} for a detailed mathematical explanation on node dependency. To overcome this issue, we need to update all affected nodes $\mathcal{V}_\text{affect}$ in parallel, which results in massive computation overhead because $|\mathcal{V}_\text{affect}|$ grows exponentially with the number of layers.

\begin{figure}
    \centering
    \includegraphics[width=0.45\textwidth]{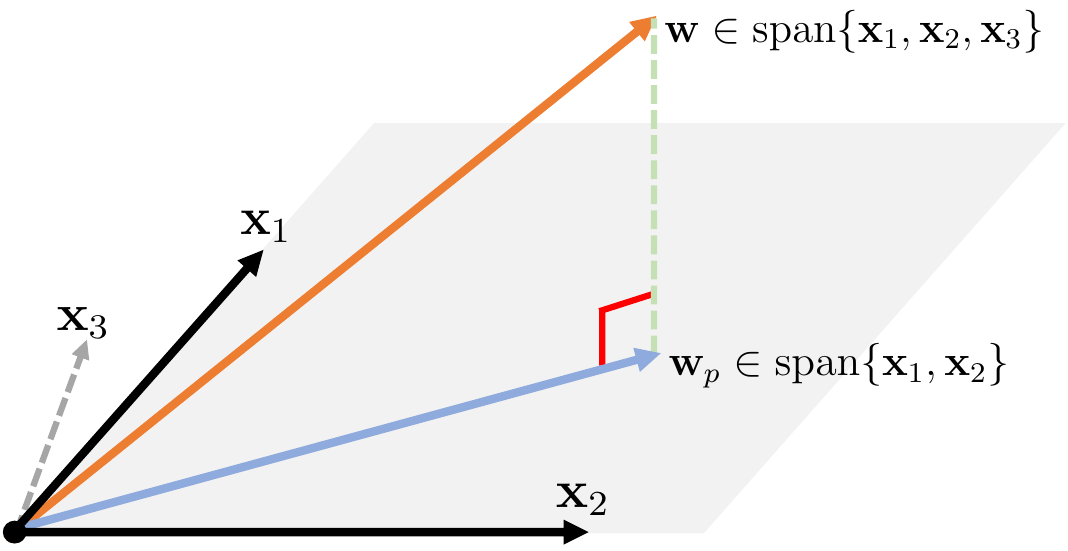}
    \vspace{-3mm}
    \caption{The orthogonal projection unlearning in \oure. The original weight $\mathbf{w}$ exists inside the subspace defined by node feature vectors $\{\mathbf{x}_1,\mathbf{x}_2,\mathbf{x}_3\}$. We can unlearn $\mathbf{x}_3$ and obtain the new weight $\mathbf{w}_p$ by projecting $\mathbf{w}$ onto the subspace defined without $\mathbf{x}_3$.}
    \label{fig:orthogonal_proj_weight}
    \vspace{-5mm}
\end{figure}

\noindent\textit{\textbf{Q2.~If not, can we effectively unlearn representations in GNNs in a computationally efficient manner?}}
We propose a projection-based unlearning approach for linear-GNNs that not only ``\emph{bypasses the node dependency issue}'' but also ``\emph{enjoys a perfect data removal guarantee}''. More specifically, we propose to unlearn node features by orthogonal projecting linear-GNN's weight parameters to a subspace that is irrelevant to the unlearned node features (Section~\ref{section:method}). The projection step guarantees our weight parameters do not carry any information about the deleted node features, please refer to Figure~\ref{fig:orthogonal_proj_weight} for an illustration of our main idea. \our could bypass the node dependency issue because the graph convolutions in linear-GNN can be re-formulated as a linear combination of the input node features and the projection-step is directly applied to the node features (Section~\ref{section:application_to_graph}). Notice that this is different from most approximate unlearning approaches because their gradient and Hessian are computed on the output of GNN models, therefore they are affected by the node dependency. 
% Besides, we propose unlearning-friendly non-linearity extension and adaptive-diffusion graph convolution to further boost \oure's performance (Section~\ref{section:extension}).

\noindent\textit{\textbf{Q3.~How to assess the effectiveness of unlearning in GNNs?}}
We consider two criteria to evaluate the effectiveness of unlearning.
Our first criterion is ``\emph{the distance between the unlearned weights to the exactly retrained weights}''. We evaluate this criterion by theoretically upper bound the distance of two models. We show that \our enjoys a tighter upper bound than approximate unlearning methods~\cite{guo2020certified,golatkar2020eternal} (Section~\ref{section:effectiveness_of_ours}).
Although this criterion has become the de facto way to measure the success of unlearning for approximate unlearning methods, it has been pointed out by~\cite{thudi2021necessity} that we cannot infer ``whether the data have been deleted'' solely from it. Our theoretical explanation on this point is deferred to the Appendix~\ref{section:proof_unlearn_without_change}. 
Therefore, we introduce our second criterion by checking ``\emph{whether unlearned weights contain the deleted node features}''. To achieve this, we introduce ``feature injection test'' in the experiment section to  rigorously verify this criterion. 

\textbf{Contributions.} The main contributions of the present paper are summarized as follows:
\begin{itemize} [noitemsep,topsep=0pt,leftmargin=5mm]
    \item We propose an efficient graph representation unlearning method \oure, which could overcome the node dependency issue and is guaranteed to remove the trace of the deleted node features (Section~\ref{section:application_to_graph}). \vspace{1mm}
    \item We theoretically show that \emph{unlearned model} of \our is closer to the \emph{model retrained from scratch} than other approximate unlearning methods, which indicates that \our is more preferred if only approximate unlearning is required (Section~\ref{section:effectiveness_of_ours}). \vspace{1mm}
    \item  To improve the expressive of the linear-GNN used with \oure, we introduce two unlearning-favorable extension, i.e., non-linearity extension and adaptive diffusion graph convolution (Section~\ref{section:extension}). \vspace{1mm}
    \item We introduce the ``feature injection test'' to rigorously verify whether an unlearning method could perfectly remove the trace of the deleted node features. Our results show that \our could perfectly remove the trace of the deleted node features, however, other approximate unlearning methods cannot, which emphasizes the importance of \our (Section~\ref{section:experiments}). \vspace{1mm}
    \item Empirical results on large-scale real-world  datasets of different sizes illustrate the effectiveness, efficiency, and robustness of \our (Section~\ref{section:experiments} and Appendix~\ref{section:more_experiment_results}).
\end{itemize}
% \circled{1} We propose an efficient graph representation unlearning method \oure,
% which could overcome the node dependency issue and is guaranteed to remove the trace of the deleted node features (Section~\ref{section:method}). 
% \circled{2} We theoretically show that ``\oure's unlearned solution'' is closer to the ``retraining from scratch model'' than other approximate unlearning methods (Section~\ref{section:effectiveness_of_ours}), which indicates that \our is more preferred if only approximate unlearning is required. To further rigorously verify whether an unlearning method could perfectly remove the trace of the deleted node features, we introduce the ``feature injection test'' in the experiment section. Our results show that \our could perfectly remove the trace of the deleted node features but other approximate unlearning methods cannot, which emphasizes the importance of \oure.  
% \circled{3} To improve the expressive of the linear-GNN used with \oure, we introduce two unlearning-favorable extension, i.e., non-linearity extension and adaptive diffusion graph convolution (Section~\ref{section:extension}).
% \circled{4} Empirical results on large-scale datasets illustrate the effectiveness, efficiency, and robustness of \our and our unlearning-favorable extensions (Section~\ref{section:experiments} and Appendix~\ref{section:more_experiment_results}).

\section{Related work and backgrounds}\label{section:existing_works}

%%%%%%%%%%%%%%%%%%%%%%%%%%%%%%%%%%%%%%%%%%%%%%%%%%%%%%%%%%%%%%%%%%%%%%%%%%%%%%
\noindent\textbf{Exact unlearning.}
The most straightforward way is to retrain the model from scratch, which is computationally demanding, except for some model-specific problems such as SVM~\cite{cauwenberghs2000incremental}, K-means~\cite{ginart2019making}, and decision tree~\cite{brophy2021machine}. 
To reduce the computation cost, \cite{bourtoule2021machine} proposes to split the dataset into multiple shards and train an independent model on each data shard, then aggregate their prediction during inference. A similar idea is explored in~\cite{aldaghri2021coded,he2021deepobliviate}. 
\textsc{GraphEraser}~\cite{chen2021graph} extends~\cite{bourtoule2021machine} to graph-structured data by proposing a graph partition method that can preserve the structural information as much as possible and weighted prediction aggregation for evaluation. \cite{chen2022recommendation} further generalize~\cite{chen2021graph} to the recommender system. Although the data partition schema allows for a more efficient retrain of models on a smaller fragment of data, the model performance suffers because each model has fewer data to be trained on and data heterogeneity can also deteriorate the performance. Moreover, if a large set of deleted nodes are selected at random, it could still result in massive retraining efforts. \cite{ullah2021machine} proposes to retrain at the iteration that deleted data the first time appears, which is not suitable if it requires iterating the full dataset multiple rounds. \cite{neel2020descent,ullah2021machine,sekhari2021remember} study the unlearning from the generalization theory perspective,~\cite{fu2022knowledge,nguyen2022markov} study unlearning for Bayesian inference, which is orthogonal to the main focus of this paper. 
% A concurrent work \textsc{GraphEditor}~\cite{cong2022grapheditor} proposes to reformulate linear-GNN training as Ridge regression and unlearn by using closed-form solution. However, it has a cubic computation complexity the multi-hop neighbors of deleted nodes, which is less efficient when graphs are dense or large.

%%%%%%%%%%%%%%%%%%%%%%%%%%%%%%%%%%%%%%%%%%%%%%%%%%%%%%%%%%%%%%%%%%%%%%%%%%%%%%
\noindent\textbf{Approximate unlearning.}
The main idea is to approximate the model trained without the deleted data in the parameter space. For example, \cite{guo2020certified} proposes to unlearn by removing the influence of the deleted data on the model parameters by first-order Taylor approximation, where the Hessian is computed on the remaining data and gradient is computed on the deleted data. \cite{chien2022certified} generalize the analysis in~\cite{guo2020certified} to graph. A similar idea has been explored in~\cite{wu2022puma} but requires an objective function as a finite-sum formulation, which is non-trivial to extend onto graph-structured data. \cite{golatkar2020eternal} performs Fisher forgetting by taking a single step of Newton's method on the remaining training data. \cite{golatkar2021mixed} generalizes the idea to deep neural networks by assuming a subset of training samples are never forgotten, which can be used to pre-train a neural network as a feature extractor and only unlearn the last layer. \cite{izzo2021approximate} speeds up~\cite{guo2020certified} by using the leave-one-out residuals for the linear model update, which reduces the time complexity to linear in the dimension of the deleted data and is independent of the size of the dataset. \cite{wu2020deltagrad} proposes to first save all the intermediate weight parameters and gradients during training, then utilize such information to efficiently estimate the optimization path. Similar idea have been explored in~\cite{wu2020priu} for logistic regression. Notice that due to the nature of approximate unlearning, these methods only approximately unlearn the information of deleted data, require adding random noise, and lack of perfect data removal guarantee in practice~\cite{thudi2021necessity}.

\noindent\textbf{Linearity requirement in unlearning.}
Linearity is required in most unlearning methods ~\cite{guo2020certified,golatkar2020eternal,wu2020deltagrad} to verify whether the trace of deleted data has been perfectly unlearned. Unless re-training from scratch, it is still an open problem to theoretically or rigorously empirically verify this in the non-linear models~\cite{thudi2021necessity,guo2020certified}. Therefore, we initiate our study on linear-GNNs in Section~\ref{section:application_to_graph} and provide its non-linearity extension in Section~\ref{section:extension}. We will rigorously test whether the information is perfectly unlearned on linear-GNNs and demonstrate the application of using \our with non-linear GNNs.

\noindent\textbf{Relation between unlearning and differential privacy.}
Unlearning and differential privacy (DP) are two concepts that could be used in parallel. More specifically, \textit{DP} aims to prevent the privacy leakage issue, while \textit{unlearning} seeks to remove some data points' effect on the pre-trained model. Recently, a number of approximate unlearning methods~\cite{guo2020certified,golatkar2020eternal,chien2022certified} are inspired by DP to unlearn by injecting random noises and derive an approximate unlearning DP-like upper bound. However, not all unlearning methods require using random noises and could be evaluated under a DP-like framework. For example, \cite{ullah2021machine,chen2021graph} unlearn by re-training from scratch and \our unlearns by orthogonal projection, therefore adding random noise is not required. Please refer to Appendix~\ref{section:differental_privacy_vs_unlearning} for more details.
In this paper, we only consider fully removing the trace of data from the model by unlearning, but do not consider preventing the privacy leakage issue with DP.

%%%%%%%%%%%%%%%%%%%%%%%%%%%%%%%%%%%%%%%%%%%%%%%%%%%%%%%%%%%%%%%%%%%%%%%%%%%%%%
%%%%%%%%%%%%%%%%%%%%%%%%%%%%%%%%%%%%%%%%%%%%%%%%%%%%%%%%%%%%%%%%%%%%%%%%%%%%%%
%%%%%%%%%%%%%%%%%%%%%%%%%%%%%%%%%%%%%%%%%%%%%%%%%%%%%%%%%%%%%%%%%%%%%%%%%%%%%%

\section{Graph representation unlearning} \label{section:method}

We first introduce backgrounds on graph learning and unlearning in Section~\ref{section:preliminary}. Then, we introduce our graph representation unlearning approach \our on linear-GNN in Section~\ref{section:application_to_graph} and theoretically analyzing its effectiveness in Section~\ref{section:effectiveness_of_ours}. Finally, we introduce \oure's non-linearity extension in Section~\ref{section:extension}.

\subsection{Backgrounds} \label{section:preliminary}
%%%%%%%%%%%%%%%%%%%%%%%%%%%%%%%%%%%%%%%%%%%%%%%%%%%%%%%%%%%%%%%%%%%%%%%%%%%%%%
We consider solving semi-supervised binary node classification using the linear-GNN, which could be easily extended to multi-class classification. More specifically, given a graph $\mathcal{G}(\mathcal{V},\mathcal{E})$ with $n=|\mathcal{V}|$ nodes and $|\mathcal{E}|$ edges, let us suppose each node $v_i \in \mathcal{V}$ is associated with a node feature vector $\mathbf{x}_i \in \mathbb{R}^{d}$. Let $\mathbf{A}, \mathbf{D} \in \mathbb{R}^{n\times n}$ denote the adjacency matrix and its associated degree matrix. Then, an $L$-layer linear-GNN\footnote{Non-linear GNNs usually add activation function and weight matrix after each graph convolution. For example, the GCN's hidden representation is computed by $\mathbf{H}^{(\ell)}=\sigma(\mathbf{P} \mathbf{H}^{(\ell-1)} \mathbf{W}^{(\ell)})$.} computes the node representation $\mathbf{H} = \mathbf{P}^L \mathbf{X} \in \mathbb{R}^{n\times d}$ by applying $L$ propagation matrices $\mathbf{P} = \mathbf{D}^{-1/2}\mathbf{A}\mathbf{D}^{-1/2}$ to the node features matrix $\mathbf{X} \in \mathbb{R}^{n \times d}$. During training, only training set nodes $\mathcal{V}_\text{train}\subset \mathcal{V}$ are labeled by a binary label $y_i\in\{-1,+1\}$, our goal is to estimate the label of the unlabeled nodes $\mathcal{V}_\text{eval}=\mathcal{V}\setminus \mathcal{V}_\text{train}$. More specifically, we want to find the weight parameters $\mathbf{w} \in\mathbb{R}^d$ that minimize
\begin{equation} \label{eq:ovr_logistic_regression}
    \begin{aligned}
    F(\mathbf{w}) &= \frac{\lambda}{2} \| \mathbf{w} \|_\mathrm{2}^2 + \frac{1}{\mathcal{V}_\text{train}}\sum_{v_i\in\mathcal{V}_\text{train}} f_i(\mathbf{w}),\\
    f_i(\mathbf{w}) &= \log\left(1 + \exp(- y_i \mathbf{w}^\top \mathbf{h}_i) \right),\mathbf{h}_i = [\mathbf{P}^L \mathbf{X}]_i.
    \end{aligned}
\end{equation}
For graph representation unlearning, let $\mathcal{V}_\text{delete}\subset \mathcal{V}_\text{train}$ denote the set of deleted nodes and $\mathcal{V}_\text{remain} = \mathcal{V}_\text{train} \setminus \mathcal{V}_\text{delete}$ denote the remaining nodes. Our goal is to unlearn the node feature information $\{ \mathbf{x}_i~|~v_i \in \mathcal{V}_\text{delete}\}$ of the deleted nodes $\mathcal{V}_\text{delete}$. 
% We assume any node feature cannot be represented by the linear combination of other node features, which is also implicitly required by other unlearning methods (e.g., to compute Hessian inverse) and can be guaranteed by adding small random noises to features that without hurting model performance. 
In terms of the notations, we denote $\mathbf{w}$ as the solution before unlearning, $\mathbf{w}_p$ as the solution obtained by \oure, and $\mathbf{w}_u$ as the solution obtained by re-training from scratch on the dataset without the deleted nodes.

\subsection{Graph representation unlearning via \oure} \label{section:application_to_graph}

The main idea behind \our is as follows: ``\emph{If the weight parameters of linear-GNN are located inside the linear span of all node features (precondition), then we can unlearn a set of node features by projecting the weight parameters onto a subspace that is irrelevant to the node features that we want to unlearn (how to unlearn).}'' In the following, we will first explain why the precondition holds in linear-GNNs, then introduce how to unlearn, and explain why \our can bypass the node dependency. 

\noindent\textit{\textbf{Why precondition holds in linear-GNN?}}
The precondition holds because the graph convolution in linear-GNN is a linear operator on node features. As a result, all gradients are inside the linear span of all node features. Therefore, if we optimizing linear-GNN (Eq.~\ref{eq:ovr_logistic_regression}) using SGD with weight initialization satisfying  $\mathbf{w}_\text{init} \in\text{span}\{\mathbf{x}_1,\ldots,\mathbf{x}_n\} $, regardless of how many steps of gradient updates, we still have $\mathbf{w}\in\text{span}\{\mathbf{x}_1,\ldots,\mathbf{x}_n\}$ holds. To see this, let us first recall that the gradient of Eq.~\ref{eq:ovr_logistic_regression} with respect to any $\mathbf{w}$ is 
\begin{equation}\label{eq:grad_linear_gnn_logistic_regression}
    \begin{aligned}
    \nabla F(\mathbf{w}) &= \lambda \mathbf{w} + \frac{1}{|\mathcal{V}_\text{train}|}\sum_{j\in\mathcal{V}_\text{train}} \nu_j \mathbf{x}_j,\\
    \nu_j &\underset{(a)}{=}  \sum_{i\in\mathcal{V}_\text{train}} \mu_i [\mathbf{P}^L]_{ij},~
    \mu_i = - y_i \sigma(-y_i \mathbf{w}^\top \mathbf{h}_i),
    \end{aligned}
\end{equation}
where $[\mathbf{P}^L]_{ij}$ denotes the $i$-th row $j$-th column of $\mathbf{P}^L$ and $\sigma(\cdot)$ is the Sigmoid function. Then, Eq.~\ref{eq:grad_linear_gnn_logistic_regression} implies that the gradient $F(\mathbf{w})$ is inside the linear span of all node features, i.e., $\nabla F(\mathbf{w}) \in \text{span}\{\mathbf{x}_1,\ldots,\mathbf{x}_n\}$. Therefore, when using gradient update rule $\mathbf{w}_{t+1} = \mathbf{w}_t - \eta \nabla F(\mathbf{w}_t)$, the weight after gradient updates still stays inside the same subspace defined by the linear span of all node features.

\noindent\textit{\textbf{How to unlearn?}} 
Recall that our goal is to unlearn node features $\mathbf{X}_\text{delete} = \{\mathbf{x}_i~|~v_i\in\mathcal{V}_\text{delete}\}$ of size $m=|\mathcal{V}_\text{delete}|$ by making sure the unlearned solution does not carry any information about $\mathbf{X}_\text{delete}$. This can be achieved by finding an alternative solution $\mathbf{w}_p$ from a subspace that is irrelevant to $\mathbf{X}_\text{delete}$. Meanwhile, we hope $\mathbf{w}_p$ is close to $\mathbf{w}$ because small changes in the input data are expected to lead to small changes in the optimal solutions. Formally, let us define $\mathcal{U}=\text{span}\{ \mathbf{x}_i~|~v_i\in\mathcal{V}_\text{remain}\}$ as the linear subspace spanned by all remaining samples and our goal is to find $\mathbf{w}_p = {{\arg\min}_{\mathbf{v}\in \mathcal{U}}}~\| \mathbf{v} - \mathbf{w} \|_2^2$. Because the vertical distance is the shortest, we can obtain $\mathbf{w}_p$ by orthogonal projecting $\mathbf{w}$ onto the subspace $\mathcal{U}$. Knowing that any projection $\Pi_\mathcal{U}(\mathbf{w})$ onto $\mathcal{U}$ is necessarily an element of $\mathcal{U}$, i.e., $\Pi_\mathcal{U}(\mathbf{w}) \in \mathcal{U}$, the results after orthogonal projection can be represented as a weighted combination of all remaining node features $\mathbf{w}_p = {\Pi_\mathcal{U}(\mathbf{w}) = \sum_{v_i\in\mathcal{V}_\text{remain}} \alpha_i \mathbf{x}_i}$, where the coefficients of the orthogonal projection $\boldsymbol{\alpha}$ is derived in Proposition~\ref{proposition:projection_step}. An illustration of the projection-based unlearning is shown in Figure~\ref{fig:orthogonal_proj_weight} and the proof is provided in Appendix~\ref{section:proof_alternative_orthogonal_projection}.

\begin{proposition}\label{proposition:projection_step}
The coefficients of the orthogonal projection is computed as $\bm{\alpha} = \mathbf{X}_\text{remain} (\mathbf{X}_\text{remain}^\top \mathbf{X}_\text{remain} )^\dagger \mathbf{w}$, where $\mathbf{X}_\text{remain} = \{\mathbf{x}_j~|~v_j\in\mathcal{V}_\text{remain}\}$ is the remaining node features and $\dagger$ is the pseudo-inverse operator. 
\end{proposition}

The significant computation required in Proposition~\ref{proposition:projection_step} includes 
computing $\mathbf{X}_\text{remain}^\top \mathbf{X}_\text{remain}\in \mathbb{R}^{d \times d}$ and its inverse with $\mathcal{O}(r d^2)$ and $\mathcal{O}(d^3)$ computation complexity, where $r=|\mathcal{V}_\text{remain}|$ is the size of remaining nodes and $d$ is node feature dimension. However, if we could pre-computed $\mathbf{X}^\top \mathbf{X}$ before the unlearning requests arrive, then we could efficiently compute $\mathbf{X}_\text{remain}^\top \mathbf{X}_\text{remain} = \mathbf{X}^\top \mathbf{X} - \mathbf{X}_\text{delete}^\top \mathbf{X}_\text{delete}$ and compute $(\mathbf{X}_\text{remain}^\top \mathbf{X}_\text{remain})^\dagger$ by applying the Woodbury identity~\cite{golub2013matrix} on $\mathbf{X}_\text{delete}^\top \mathbf{X}_\text{delete},~\mathbf{X}^\top \mathbf{X}$, which leads to a lower computation complexity of $\mathcal{O}(\max\{m^3, m d^2\})$ if $m < \max\{r,d\}$. After obtaining $\bm{\alpha}$, \our computes the unlearned weight parameters by $\mathbf{w}_p = \mathbf{X}_\text{remain}^\top \bm{\alpha}$. 
Intuitively, the projection step in \our could be thought of as a re-weighting on the remaining nodes, which allows our model to behave as close to the model before unlearning as possible, but without carrying any information about the deleted node features.
Therefore, the output of \our could be interpreted as re-training on the remaining graph under some unknown importance sampling distribution.

\begin{algorithm} [t]
\caption{\our to unlearning linear-GNN}\label{alg:projector}
\begin{algorithmic}
    \Require The pre-trained parameters $\mathbf{w}$, (\textcolor{cyan}{\textit{Option 1}}) remain nodes' features $\mathbf{X}_\text{remain}$, (\textcolor{teal}{\textit{Option 2}}) deleted node features $\mathbf{X}_\text{delete}$, pre-computed $\mathbf{M} = \mathbf{X}^\top  \mathbf{X}$ and $\mathbf{M}^{\dagger} = (\mathbf{X}^\top  \mathbf{X})^{\dagger}$
    \Ensure Unlearned weight parameters $\mathbf{w}_p$

    \If{(\textcolor{cyan}{\textit{Option 1}}) $\mathbf{X}_\text{remain}$ is available}
        \State Compute $\mathbf{M}_\text{remain}$ by $$\mathbf{M}_\text{remain} = \mathbf{X}_\text{remain}^\top \mathbf{X}_\text{remain}$$ 
        \State Compute $\mathbf{M}_\text{remain}^{\dagger}$ by $$\mathbf{M}_\text{remain}^{\dagger} = (\mathbf{X}_\text{remain}^\top \mathbf{X}_\text{remain})^{\dagger}$$
    \ElsIf{(\textcolor{teal}{\textit{Option 2}}) $\mathbf{X}_\text{delete},\mathbf{M},~\mathbf{M}^{\dagger}$ are available}
        \State Compute $\mathbf{M}_\text{remain}$ by $$\mathbf{M}_\text{remain} = \mathbf{M} - \mathbf{X}_\text{delete}^\top \mathbf{X}_\text{delete}$$
        \State Compute $\mathbf{M}_\text{remain}^{\dagger}$ by
        \begin{equation*}
            \begin{aligned}
                    &\mathbf{S} = \mathbf{X}_\text{delete}^\top [\mathbf{I} - \mathbf{X}_\text{delete} \mathbf{X}_\text{delete}^\top] \mathbf{X}_\text{delete} \\
                    &\mathbf{M}_\text{remain}^{\dagger} = 
                    \mathbf{M}^{\dagger} +  \mathbf{M}^{\dagger} \mathbf{S} \mathbf{M}^{\dagger}
            \end{aligned}
        \end{equation*}
        % $$\mathbf{M}_\text{remain}^{\dagger} = 
        %     \mathbf{M}^{\dagger} +  \mathbf{M}^{\dagger} \mathbf{X}_\text{delete}^\top [\mathbf{I} - \mathbf{X}_\text{delete} \mathbf{X}_\text{delete}^\top] \mathbf{X}_\text{delete} \mathbf{M}^{\dagger}$$
    \EndIf
    
    \State Compute $\mathbf{w}_p = \mathbf{M}_\text{remain} \mathbf{M}_\text{remain}^{\dagger} \mathbf{w}$ as final output
\end{algorithmic}
\end{algorithm}

To this end, we summarize \our in Algorithm~\ref{alg:projector}, where two different types of input options are available that lead to identical results. More specifically, we can use \textcolor{cyan}{\textit{option 1}} if only remaining node features are available, otherwise we can use \textcolor{teal}{\textit{option 2}} if only the features of deleted nodes are available but pre-computing is feasible. Besides,  due to the similarity between logistic regression and SVM, \our could also be used in primal-based SVM unlearning~\cite{chu2015warm} to alleviate the high computation cost of the dual-based SVM unlearning approach~\cite{cauwenberghs2000incremental}. Readers could refer to Appendix~\ref{section:connection_to_svm} for more details on its application to SVM.

\noindent\textit{\textbf{Why node dependency is bypassed?}} From Eq.~\ref{eq:grad_linear_gnn_logistic_regression} (a), we could tell that node dependencies in $\mathbf{P}$ are included inside the finite sum weight $\mu_j$, which is a constant that is multiplied with its features $\mathbf{x}_j$. \our could bypass the node dependency because our projection-step is directly applied to the input node features, instead of the final outputs of GNNs. This is not the case for most approximate unlearning methods, e.g.,~\cite{guo2020certified,golatkar2020eternal,wu2020deltagrad}, because their unlearning requires computing the gradient or Hessian on the final layer outputs.

\noindent\textit{\textbf{Extension to multi-class classification.}}
Please notice that \our also works with cross-entropy loss for multi-class classification. To see this, let us consider $C$ categories and $N$ data but without considering the node dependency for simplicity, i.e.,
optimizing ${\mathbf{W} = [\mathbf{w}_1,\ldots,\mathbf{w}_C] \in\mathbb{R}^{C \times d}}$ on $\{\mathbf{x}_1,\ldots,\mathbf{x}_N\}$ where $\mathbf{w}_c$ is the $c$-th row of $\mathbf{W}$.
Then, the softmax's $c$-th class probability computed on $\mathbf{x}_n$ is  
\begin{equation*}
    p_{n,c} = \frac{\exp(a_{n,c})}{\sum_{i=1}^C \exp(a_{n,i})},~a_{n,c}=\mathbf{w}_c^\top \mathbf{x}_n.
\end{equation*}
We define the objective function as 
\begin{equation*}
    L_\mathbf{W} = - \sum_{n=1}^N \sum_{c=1}^C y_{n,c} \log (p_{n,c}),
\end{equation*}
then its gradient is 
\begin{equation*}
    \frac{\partial L_\mathbf{W}}{\partial \mathbf{w}_c} = \sum_{n=1}^N \sum_{i=1}^C \frac{\partial L_\mathbf{W}}{\partial a_{n,i}} \frac{\partial a_{n,i}}{\partial \mathbf{w}_c} = \sum_{n=1}^N (p_{n,c} - y_{n,c}) \mathbf{x}_n
\end{equation*}
because 
\begin{equation*}
    \frac{\partial L_\mathbf{W}}{\partial a_{n,i}} = p_{n,i} - y_{n,i}
    ~\text{and}~
    \frac{\partial a_{n,i}}{\partial \mathbf{w}_c} = \begin{cases}
        \mathbf{x}_n & \text{ if } i=c \\ 
        \mathbf{0} & \text{ if } i\neq c. 
        \end{cases}
\end{equation*} 
As a result, for any $j\in[C]$ we have 
\begin{equation*}
    \frac{\partial L_\mathbf{W}}{\partial \mathbf{w}_j} \in \text{span}\{\mathbf{x}_1,\ldots,\mathbf{x}_N\},
\end{equation*}
which means each row of the $\mathbf{W}$ is in the span of all node features, and we can apply \our on each row of $\mathbf{W}$ independently to unlearn. 

\subsection{On the effectiveness of \our}\label{section:effectiveness_of_ours}
In this section, we study the effectiveness of \our by measuring the $\ell_2$-norm on the difference between \oure's unlearned solution $\mathbf{w}_p$ to the solution obtained by retraining from scratch $\mathbf{w}_u$ on the dataset without the deleted nodes, and we are expecting $\|\mathbf{w}_p -  \mathbf{w}_u \|_2$ to be small for good unlearning methods. For unlearning, we suppose a random subset of nodes $\mathcal{V}_\text{delete} \subset \mathcal{V}_\text{train}$ are selected and the remaining nodes are denoted as $\mathcal{V}_\text{remain} = \mathcal{V}_\text{train}  \setminus \mathcal{V}_\text{delete}$. Since removing the nodes $\mathcal{V}_\text{delete}$ is the same as updating the propagation matrix from $\mathbf{P}$ to $\mathbf{P}_u$, where all edges that are connected to node $v_i\in\mathcal{V}_\text{delete}$ are removed in $\mathbf{P}_u$, we can write down the objective after data deletion $F^u(\mathbf{w}_u) $ as
\begin{equation} \label{eq:closeness_after_delete_objective}
    \begin{aligned}
    F^u(\mathbf{w}_u) &= \frac{1}{|\mathcal{V}_\text{remain}|}\sum_{v_i\in\mathcal{V}_\text{remain}} f_i^u(\mathbf{w}_u), \\
    f_i^u(\mathbf{w}_u) &= \log\left( 1 + \exp(-y_i \mathbf{w}_u^\top \mathbf{h}^u_i) \right) + \lambda \| \mathbf{w}_u\|_2,~
    \end{aligned}
\end{equation}
where $\mathbf{h}^u_i = [\mathbf{P}_u^L \mathbf{X}]_i$. 

Before proceeding to our result, we make the following customery assumptions on graph propagation matrices, node features, and weight parameters in Assumption~\ref{assumption:upper_bound_on_node_feats}, on the variance of stochastic gradients in Assumption~\ref{assumption:expectation_random_delete_nodes}, and on the correlation between node feature in Assumption~\ref{assumption:upper_bound_on_approximation_error}. Please notice that Assumption~\ref{assumption:upper_bound_on_node_feats},~\ref{assumption:expectation_random_delete_nodes} are standard assumptions in GNN's theoretical analysis~\cite{cong2021provable,ramezani2021learn} and Assumption~\ref{assumption:upper_bound_on_approximation_error} is a mild assumption that could be empirically verified in Table~\ref{table:delta_on_real_world} on real-world dataset, where $\delta$ could be think of as a measurement on the closeness of the subspace defined with and without the deleted node features. In practice, $\delta$ is small if only a small amount of nodes are removed from the original graph.

\begin{assumption}\label{assumption:upper_bound_on_node_feats}
We assume each row of the propagation matrices before and after node deletion is bounded by $P_s\geq0$, i.e., $\max_j \left\| [ \mathbf{P}^L ]_j \right\|_2 \leq P_s, ~\max_j \left\| [ \mathbf{P}^L_u ]_j \right\|_2 \leq P_s$.
Besides, we assume each row of the difference of the propagation matrices before and after data deletion is bounded by $P_d\geq0$, i.e., $\max_j \left\| [ \mathbf{P}^L_u - \mathbf{P}^L ]_j \right\|_2 \leq P_d$.
Furthermore, we assume the norm of any node features $\mathbf{x}_i,~v_i\in\mathcal{V}$ and weight parameters $\mathbf{w}$ are bounded by $B_x, B_w \geq 0$, i.e., $\| \mathbf{x}_i \|_2 \leq B_x,~\| \mathbf{w}\|_2 \leq B_w$.
\end{assumption}

\begin{assumption} \label{assumption:expectation_random_delete_nodes}
For any deleted nodes $\mathcal{V}_\text{delete}$, the gradient variance computed on the remaining nodes $\mathcal{V}_\text{remain} = \mathcal{V} \setminus \mathcal{V}_\text{delete}$ can be upper bounded by $G\geq0$, i.e.,  we have $\mathbb{E}_{\mathcal{V}_\text{delete}}[\| \mathbf{g} - \tilde{\mathbf{g}}\|_2] \leq G$, where $\mathbf{g} = \frac{1}{|\mathcal{V}_\text{train}|}\sum\nolimits_{v_i\in\mathcal{V}_\text{train}} \nabla f_i^u(\mathbf{w})$ and $\tilde{\mathbf{g}} = \frac{1}{|\mathcal{V}_\text{remain}|}\sum\nolimits_{v_i\in\mathcal{V}_\text{remain}} \nabla f_i^u(\mathbf{w})$ for any $\mathbf{w}$.
\end{assumption}
\begin{assumption} \label{assumption:upper_bound_on_approximation_error}
For any node $v_j\in\mathcal{V}_\text{delete}$, its node feature $\mathbf{x}_j$ can be approximated by the linear combination of all node features in the remaining node set $\{\mathbf{x}_i~|~v_i\in\mathcal{V}_\text{remain}\}$ up to an error $\delta \geq 0$. Formally, we have $\max_{v_j\in\mathcal{V}_\text{delete}} \min_{\bm{\alpha}} \left\| \sum_{i\in\mathcal{V}_\text{remain}} \alpha_i \mathbf{x}_i - \mathbf{x}_j \right\|_2 \leq \delta$.
\end{assumption}

To this end, let us introduce our main results. From Theorem~\ref{theorem:ell2_norm_of_weight_projections}, we know that $\| \mathbf{w}_p - \mathbf{w}_u \|_2$ is mainly controlled by three key factors: \circled{1} the difference between the propagation matrices before and after data deletion, which is captured by $P_d$ in Assumption~\ref{assumption:upper_bound_on_node_feats}; 
\circled{2} the variance of stochastic gradient computed on the remaining nodes, which is captured by $G$ in Assumption~\ref{assumption:expectation_random_delete_nodes};
\circled{3} the closeness of any deleted node features that could be approximated by weighted combination of all node features in the remaining node sets, which is captured by $\delta$ in Assumption~\ref{assumption:upper_bound_on_approximation_error}.
By reducing the number of nodes in $\mathcal{V}_\text{delete}$, all $P_d, \delta, G$ are expected to decrease.
At an extreme case with $|\mathcal{V}_\text{delete}|=0$, we have $P_d =\delta = G=0$ and  $\mathbf{w}_p=\mathbf{w}=\mathbf{w}_u$.
The proof is deferred to Appendix~\ref{section:proof_of_ell2_norm_of_weight_projections}.
\begin{theorem} \label{theorem:ell2_norm_of_weight_projections}
Let us suppose Assumptions~\ref{assumption:upper_bound_on_node_feats},\ref{assumption:expectation_random_delete_nodes},\ref{assumption:upper_bound_on_approximation_error} hold. Let us define $\mathbf{w}_p$ as the solution obtained by \oure, $\mathbf{w}_u$ is the solution obtained by re-training from scratch with objective function $F^u(\mathbf{w})$, and we assume $\mathbf{w}_u$ is well trained such that $\mathbf{w}_u\approx\arg\min_\mathbf{w} F^u(\mathbf{w})$.  Then, the closeness of $\mathbf{w}_p$ to the weight parameters $\mathbf{w}_u$ can be bounded by
\begin{equation}\label{eq:weight_diff_bound}
    \begin{aligned}
    &\mathbb{E}_{\mathcal{V}_\text{delete}}[\| \mathbf{w}_u - \mathbf{w}_p \|_2 ]
    \leq \Delta = \\
    &Q \sum\nolimits_{t=1}^T \left(1+\eta (\lambda+B_x^2P_s^2)\right)^{t-1} + \delta \eta T \times |\mathcal{V}_\text{delete}|,
    \end{aligned}
\end{equation}
where $Q = \eta \big( (1+B_x B_w P_s) B_x P_d + G  \big)$ and $\eta$ is the learning rate used to pre-train the weight $\mathbf{w}$ for $T$ steps of gradient descent updates.
After projection, we can fine-tune $\mathbf{w}_p$ for $K$ iterations with learning rate $(\lambda+B_x^{2} P_s^{2})^{-1}$ to obtain $\tilde{\mathbf{w}}_p$ that has an error
$F^u(\tilde{\mathbf{w}}_p) - \min_\mathbf{w} F^u(\mathbf{w}) \leq \mathcal{O}((\lambda+B^2_x P_s^2) \Delta / K)$.
\end{theorem}

Besides,  we know the solution of \our is probably closer to the model retrained from scratch compared to~\cite{guo2020certified,golatkar2020eternal} if $\delta$ satisfies the condition in Proposition~\ref{proposition:closeness_of_projector_vs_inflence}. In practice, the condition is very likely to be satisfied because learning rate $\eta$, regularization term $\lambda$, and the ratio of deleted nodes $|\mathcal{V}_\text{delete}|/|\mathcal{V}|$ are usually very small. For example, a common choice of learning rate and regularization is $\eta=0.01, \lambda=10^{-6}$ for most model training. Moreover, we empirically validate the difference between the weight before and after unlearning in the experiment section to validate our theoretical results. The proof of Proposition~\ref{proposition:closeness_of_projector_vs_inflence} is deferred to Appendix~\ref{section:proof_of_closeness_of_projector_vs_inflence}.
\begin{proposition} \label{proposition:closeness_of_projector_vs_inflence}
If the approximation error in Assumption~\ref{assumption:upper_bound_on_approximation_error} satisfying $\delta  < \left((\lambda \eta T)^{-1} + 1 \right) B_x \times \frac{|\mathcal{V}|}{|\mathcal{V}_\text{delete}|}$, then \oure's output is provably closer to re-training from scratch then using approximate unlearning \textsc{Influence}~\cite{guo2020certified} and \textsc{Fisher}~\cite{golatkar2020eternal}.
\end{proposition}

%%%%%%%%%%%%%%%%%%%%%%%%%%%%%%%%%%%%%%%%%%%%%%%%%%%%%%%%%%%%%%%%%%%%%
%%%%%%%%%%%%%%%%%%%%%%%%%%%%%%%%%%%%%%%%%%%%%%%%%%%%%%%%%%%%%%%%%%%%%
%%%%%%%%%%%%%%%%%%%%%%%%%%%%%%%%%%%%%%%%%%%%%%%%%%%%%%%%%%%%%%%%%%%%%

\subsection{Toward a more powerful structure} \label{section:extension}
To boost the model performance \oure, we first introduce an unlearning-favorable non-linearity extension to help \our better leverage node feature information, then we introduce an unlearning favorable adaptive diffusion graph convolution to help \our better leverage the graph structure information.

\noindent\textbf{An extension from linear to non-linear.}
Recall that the geometric view of solving logistic regression is finding a hyperplane to linearly separate the node representations $\mathbf{H}$ computed by linear-GNN. However, node representations computed by linear-GNNs might not be linearly separable.  To overcome this issue, we propose to first apply a \textit{MLP} on all node features,  then apply \textit{linear-GNN} onto the output of the MLP before classification, i.e., \smash{$\mathbf{Z}=\sigma(\sigma(\mathbf{X} \mathbf{W}^{(1)}_\text{mlp}) \mathbf{W}^{(2)}_\text{mlp}),~\mathbf{H}=\mathbf{P}^L \mathbf{Z} \mathbf{W}_\text{gnn}$}. The above extension can be interpreted as finding a non-linear separation in the input space. During training, we could first pre-train on a public dataset with training samples that do not need to be forgotten, then we only need to unlearn the linear-GNN model by applying \our onto the output of the MLP.
By doing so, \our enjoys both the separation power brought by the non-linearity of MLP and the efficiency brought by the projection-based unlearning.

\noindent\textbf{Adaptive diffusion graph convolution.}
To help the linear-GNN fully take advantages of the graph structure, we propose an unlearning favorable adaptive diffusion graph convolution operation that take the similarity of both node feature and node label category information into consideration. To achieve this, let us first initialize the node features as $\smash{\mathbf{h}_i^{(0)} = \mathbf{x}_i}$, initialize node labels as $\smash{\mathbf{z}_i^{(0)} = \mathbf{y}_i}$ if $\smash{i\not\in\mathcal{V}_\text{test}}$ and $\smash{\mathbf{z}_i^{(0)} = \mathbf{0}}$ if $\smash{i\in\mathcal{V}_\text{test}}$. Then, the forward propagation of the adaptive diffusion graph convolution operation is computed as $$[\mathbf{H}^{(\ell+1)}, \mathbf{Z}^{(\ell+1)}] = \big( (1-\gamma) \mathbf{I} + \gamma \mathbf{D}^{(\ell)}_\mathcal{G} \big) [\mathbf{H}^{(\ell)}, \mathbf{Z}^{(\ell)}],$$ where we denote $[\cdot,\cdot]$ as the feature channel concatenation operation and the $i$-th row $j$-th column of the $\ell$-th diffusion operator is defined by $$[ \mathbf{D}^{(\ell)}(\mathcal{G}) ]_{i,j} = \frac{1}{Z}\exp( - \sigma_h^2\| \mathbf{h}_i^{(\ell)} - \mathbf{h}_j^{(\ell)} \|_2^2 - \sigma_z^2\| \mathbf{z}_i^{(\ell)} - \mathbf{z}_j^{(\ell)} \|_2^2 ),$$
where $\sigma_h, \sigma_z \in \mathbb{R}$ are learned during training. Intuitively, our diffusion operator assign a higher neighbor aggregation weight to a node if it has a similar node feature and label information. Then, we set $\mathbf{H} = [\mathbf{H}^{(1)}, \mathbf{Z}^{(1)}, \ldots, \mathbf{H}^{(L)}, \mathbf{Z}^{(L)}]$ as the final node representation for prediction. During unlearning, we do not have to modify $\sigma_h, \sigma_z$ since these scalars will not leak the node feature information.

To this end, we conclude this section by showing in Proposition~\ref{prop:linear_as_expressive_as_nonlinear} that under mild conditions on $\mathbf{X}$ and $\mathbf{P}$, the linear-GNN used in \our could approximate any function defined on the graph. Since non-linearity extension and adaptive diffusion graph convolution could potentially alleviate the conditions on $\mathbf{X}$ and $\mathbf{P}$, these extensions could improve the expressive power of linear-GNN. 
\begin{proposition} \label{prop:linear_as_expressive_as_nonlinear}
Let us define $\mathbf{U},\bm{\lambda}$ as the eigenvectors and eigenvalues of graph propagation matrix $\mathbf{P}$, $g_{\mathbf{w}}(\mathbf{L}, \mathbf{X}) = \sum_{\ell=1}^n (\mathbf{P}^{\ell-1} \mathbf{X}) \mathbf{w}_\ell $ as the linear-GNN, and $f(\mathbf{P}, \mathbf{X}) \in \mathbb{R}^{n\times 1}$ as the target function we want to approximate by linear-GNN. If no elements in $\bm{\lambda}$ are identical and no rows of $\Tilde{\mathbf{X}} = \mathbf{U} \mathbf{X}$ are zero vectors, then 
there is always exists a set of $\mathbf{w}_\ell^\star \in \mathbb{R}^d$ such that $g_{\mathbf{w}^\star}(\mathbf{P}, \mathbf{X}) = f(\mathbf{P}, \mathbf{X})$. 
Replacing $\mathbf{P}$ with adaptive diffusion graph convolution and replace $\mathbf{X}$ as the output of MLP model could potentially alleviate our requirement on the $\bm{\lambda}$ and $\Tilde{\mathbf{X}}$ since their values are learned by training, therefore improving its expressiveness.
\end{proposition}

The intuition behind above proposition is that the expressive power of the linear-GNN $g_{\mathbf{w}}(\mathbf{L}, \mathbf{X})$ mainly comes from its graph convolution.
Given a dataset with $n$ nodes, using graph convolutions with polynomial from $0$ to $n-1$ allows us map each node feature to its desired value with $n$ different weight parameters, therefore it could approximate any function defined on graph. Proof deferred to Appendix~\ref{section:linear_as_expressive}.

\section{Experiments} \label{section:experiments}

\begin{figure*}[t]
    \centering
    \includegraphics[width=0.99\textwidth]{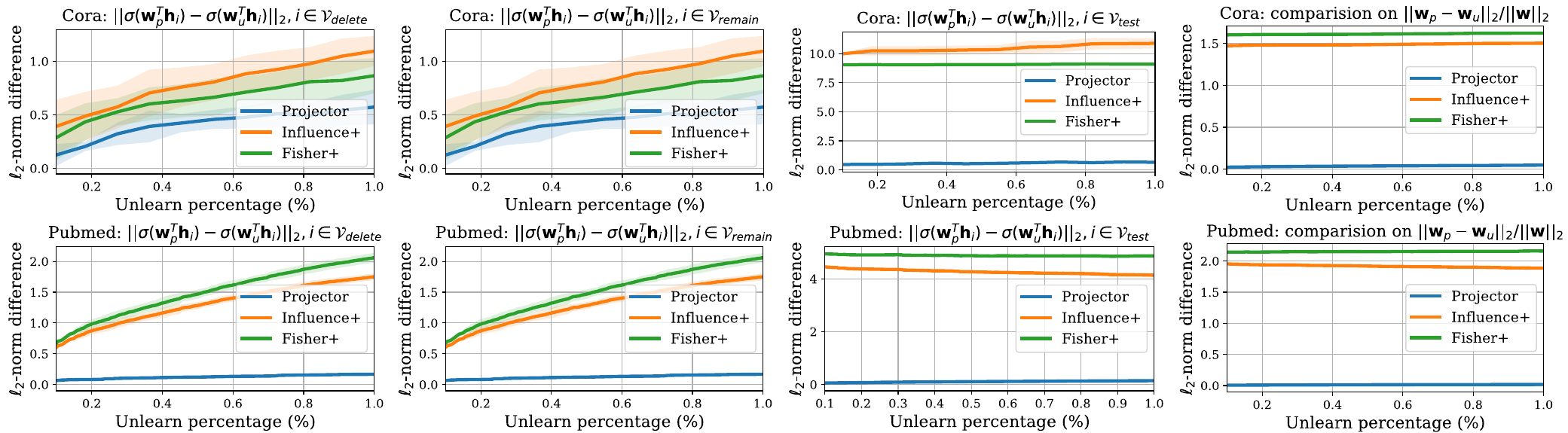}
    \vspace{-4mm}
    \caption{Comparison on the weight difference and model prediction (after the final layer activation function) before and after the unlearning process.}
    \label{fig:weight_diff_cora_citeseer_pubmed}
    \vspace{-4mm}
\end{figure*}

We consider \textsc{GraphEraser} as our exact graph unlearning baseline.
For approximate graph unlearning baselines, we extend \textsc{Influence} and \textsc{Fisher} to graph structured data by taking the node dependency into consideration and rename them as \textsc{Influence}+ and \textsc{Fisher}+. The details on the baselines are introduced in Appendix~\ref{section:baseline_details}.
Moreover, since each experiment is designed to evaluate different aspect of unlearning, the setup of each experiment could be slightly different (e.g., linear or non-linear, different deleted node size, different datasets, etc). Therefore, we choose to provide a brief introduction on the experiment design and setup at the beginning of each experiment paragraph, but defer the detailed descriptions to Appendix~\ref{section:experiment_details}.

\subsection{Experiment results}\label{section:experient_results}
\noindent\textbf{Feature injection test.}
This experiment is designed to verify whether \our and baselines could perfectly unlearn the trace of deleted node features from the weight parameters. To achieve this goal, we append an extra binary feature to all nodes and set the extra binary feature as $1$ for the deleted nodes and as $0$ for other nodes. To make sure this extra binary feature is an important feature and is heavily used during training, we add an extra category and change all deleted nodes to this extra category, then pre-train on the modified dataset. We measure the effectiveness of unlearning by checking \circled{1} whether unlearning method can \textcolor{cyan}{\emph{fully unlearn}} by comparing weight norm of the injected channel before and after unlearning\footnote{Since the weight parameters of logistic regression are weighted combination of all input features used during training, the weight norm of the injected channel before unlearning is expected to be positive if $\mathcal{V}_\text{delete}$ are used before unlearning.  However, if an unlearning method could perfectly remove the trace of $\mathcal{V}_\text{delete}$, the weight norm of the injected channel after unlearning should be zero because the features of $\mathcal{V}_\text{delete}$ does not belong to the support vectors of weight parameters. 
% Another way to understand this is from Eq.~\ref{eq:grad_linear_gnn_logistic_regression}, where the weight norm of the injected channel is not equal to zero if deleted nodes are used during training, but is zero if deleted nodes are not used during training. 
}; \circled{2} whether unlearning method \textcolor{brown}{\emph{hurt the model performance}} by comparing the accuracy before and after unlearning; \circled{3} the \textcolor{teal}{\emph{computation cost}} by comparing the time required for unlearning. We randomly select $5\%, 10\%$ of the nodes from the training set as deleted nodes. We have the following observations from Table~\ref{table:feature_injection_test}: \circled{1} By comparing the weight norm of the injected channel, we observe that \textsc{GraphEraser} and \our can perfectly unlearn the deleted nodes and setting the extra-feature channel as zero. However, \textsc{Influence}+, \textsc{Fisher}+ cannot fully unlearn the correlation because they are approximate unlearning methods; \circled{2} By comparing the wall-clock time, \our requires less time to unlearn because it is one-shot unlearning approach with the least computation cost, whereas baselines either require re-training for multiple iterations (e.g., \textsc{GraphEraser}) or require a larger computation cost to compute Hessian inversion (e.g., \textsc{Influence} and \textsc{Fisher}); \circled{3} By comparing the accuracy before and after unlearning, \textsc{Influence}+ and \textsc{Fisher}+ have around $2\%/7\%$ performance degradation on OGB-Arxiv/Products dataset than re-training because a stronger regularization is required to stabilize the unlearning process (to make sure the Hessian inverse is bounded), and \textsc{GraphEraser} have around $4\%/9\%$ performance degradation on OGB-Arxiv/Products dataset due to graph partitioning; \circled{4} By comparing the performance of \our with and without adaptive diffusion, we know that adaptive diffusion provides consistent performance boosting to linear-GNN models; \circled{5} When comparing with re-training from scratch, \our is around $0.04\sim0.2\%$ slightly better than re-training because \our could be thought of as a re-weighting on the remaining nodes, which allows our model to behave similar to the model before unlearning, but without carrying information about the deleted nodes.

\begin{table}[t]
\centering
\caption{Comparison on the \textit{F1-score accuracy} (\textcolor{brown}{Acc}),
and the norm of extra-feature weight channel (\textcolor{cyan}{WN}) before unlearning and after unlearning (denoted as \textit{before} $\rightarrow$ \textit{after}), and wall-clock time (\textcolor{teal}{T}) using linear GNN. }
\label{table:feature_injection_test}
\vspace{-4mm}
\centering\setlength{\tabcolsep}{1mm}
\scalebox{0.8}{
\begin{tabular}{l l l l l}
\hline\hline \rule{0pt}{2ex}   
                                       & Method                & Metrics  & Delete $5\%$ nodes           & Delete $10\%$ nodes        \\ \hline\hline \rule{0pt}{2ex}   
\multirow{12}{*}{\rotatebox[origin=c]{90}{\textbf{OGB-Arxiv}}}    
                                       & \cellcolor[HTML]{FEDEDC} & \cellcolor[HTML]{FEDEDC}\textcolor{brown}{Acc} (\%)      & \cellcolor[HTML]{FEDEDC}$73.33\rightarrow 73.39$       & \cellcolor[HTML]{FEDEDC}$73.25\rightarrow 73.39$     \\ 
                                       & \cellcolor[HTML]{FEDEDC}\multirow{-2}{*}{\our}  & \cellcolor[HTML]{FEDEDC}\textcolor{cyan}{WN} (\textcolor{teal}{T})   & \cellcolor[HTML]{FEDEDC}$21.7 \rightarrow 0$~($0.07$ s)   & \cellcolor[HTML]{FEDEDC}$56.8 \rightarrow 0$~($0.07$ s) \\
                                       \cline{2-5} \rule{0pt}{2ex}   
                                       %%%%%%%%%%%%%%%%%%%%%%%%%%%%%%%%%%%%%%%%%%%%%%%%%%%%%%%%%%%%%%%%%%%%%%%%%%%%%%%%%%%%%%%%%%%%%%%%%%%%%%%%%%%%%%%%%%%%%%%%%%%%%%%%%%%%%%%%%%%%
                                       %%%%%%%%%%%%%%%%%%%%%%%%%%%%%%%%%%%%%%%%%%%%%%%%%%%%%%%%%%%%%%%%%%%%%%%%%%%%%%%%%%%%%%%%%%%%%%%%%%%%%%%%%%%%%%%%%%%%%%%%%%%%%%%%%%%%%%%%%%%%
                                       %%%%%%%%%%%%%%%%%%%%%%%%%%%%%%%%%%%%%%%%%%%%%%%%%%%%%%%%%%%%%%%%%%%%%%%%%%%%%%%%%%%%%%%%%%%%%%%%%%%%%%%%%%%%%%%%%%%%%%%%%%%%%%%%%%%%%%%%%%%%
                                       & \cellcolor[HTML]{C4F8F6} & \cellcolor[HTML]{C4F8F6}\textcolor{brown}{Acc} (\%)         & \cellcolor[HTML]{C4F8F6}$73.42\rightarrow 73.48$         & \cellcolor[HTML]{C4F8F6}$73.34\rightarrow 73.44$     \\ 
                                       & \cellcolor[HTML]{C4F8F6}\multirow{-2}{*}{\cellcolor[HTML]{C4F8F6}\begin{tabular}[c]{@{}l@{}} \our \\ (+ adapt diff) \end{tabular}}  & \cellcolor[HTML]{C4F8F6}\textcolor{cyan}{WN} (\textcolor{teal}{T})  & \cellcolor[HTML]{C4F8F6}$24.3 \rightarrow 0$~($0.07$ s)   & \cellcolor[HTML]{C4F8F6}$25.6 \rightarrow 0$~($0.07$ s) \\ \cline{2-5} \rule{0pt}{2ex}  
                                       %%%%%%%%%%%%%%%%%%%%%%%%%%%%%%%%%%%%%%%%%%%%%%%%%%%%%%%%%%%%%%%%%%%%%%%%%%%%%%%%%%%%%%%%%%%%%%%%%%%%%%%%%%%%%%%%%%%%%%%%%%%%%%%%%%%%%%%%%%%%
                                       %%%%%%%%%%%%%%%%%%%%%%%%%%%%%%%%%%%%%%%%%%%%%%%%%%%%%%%%%%%%%%%%%%%%%%%%%%%%%%%%%%%%%%%%%%%%%%%%%%%%%%%%%%%%%%%%%%%%%%%%%%%%%%%%%%%%%%%%%%%%
                                       %%%%%%%%%%%%%%%%%%%%%%%%%%%%%%%%%%%%%%%%%%%%%%%%%%%%%%%%%%%%%%%%%%%%%%%%%%%%%%%%%%%%%%%%%%%%%%%%%%%%%%%%%%%%%%%%%%%%%%%%%%%%%%%%%%%%%%%%%%%%
                                       & \multirow{2}{*}{\begin{tabular}[c]{@{}l@{}} \textsc{GraphEraser} \\ ($\times 8$ subgraphs) \end{tabular}} 
                                                               & \textcolor{brown}{Acc} (\%)     & $70.59\rightarrow 70.56$       & $70.55\rightarrow 70.23$      \\ 
                                       &                       & \textcolor{cyan}{WN} (\textcolor{teal}{T}) & $22.3 \rightarrow 0$~($1,866$ s)   & $30.6 \rightarrow 0$~($1,866$ s) \\\cline{2-5} \rule{0pt}{2ex}    
                                      %%%%%%%%%%%%%%%%%%%%%%%%%%%%%%%%%%%%%%%%%%%%%%%%%%%%%%%%%%%%%%%%%%%%%%%%%%%%%%%%%%%%%%%%%%%%%%%%%%%%%%%%%%%%%%%%%%%%%%%%%%%%%%%%%%%%%%%%%%%%
                                       %%%%%%%%%%%%%%%%%%%%%%%%%%%%%%%%%%%%%%%%%%%%%%%%%%%%%%%%%%%%%%%%%%%%%%%%%%%%%%%%%%%%%%%%%%%%%%%%%%%%%%%%%%%%%%%%%%%%%%%%%%%%%%%%%%%%%%%%%%%%
                                       %%%%%%%%%%%%%%%%%%%%%%%%%%%%%%%%%%%%%%%%%%%%%%%%%%%%%%%%%%%%%%%%%%%%%%%%%%%%%%%%%%%%%%%%%%%%%%%%%%%%%%%%%%%%%%%%%%%%%%%%%%%%%%%%%%%%%%%%%%%%
                                      & \multirow{2}{*}{\textsc{Influence}+} 
                                                              & \textcolor{brown}{Acc} (\%)      & $71.90\rightarrow 72.73$       & $ 70.40 \rightarrow 72.65 $     \\ 
                                      &                       & \textcolor{cyan}{WN} (\textcolor{teal}{T}) & $29.2 \rightarrow 14.1$~($1.1$ s)   & $21.1 \rightarrow 12.1$~($1.1$ s) \\\cline{2-5} \rule{0pt}{2ex}    
                                    %   %%%%%%%%%%%%%%%%%%%%%%%%%%%%%%%%%%%%%%%%%%%%%%%%%%%%%%%%%%%%%%%%%%%%%%%%%%%%%%%%%%%%%%%%%%%%%%%%%%%%%%%%%%%%%%%%%%%%%%%%%%%%%%%%%%%%%%%%%%%%
                                    %   %%%%%%%%%%%%%%%%%%%%%%%%%%%%%%%%%%%%%%%%%%%%%%%%%%%%%%%%%%%%%%%%%%%%%%%%%%%%%%%%%%%%%%%%%%%%%%%%%%%%%%%%%%%%%%%%%%%%%%%%%%%%%%%%%%%%%%%%%%%%
                                    %   %%%%%%%%%%%%%%%%%%%%%%%%%%%%%%%%%%%%%%%%%%%%%%%%%%%%%%%%%%%%%%%%%%%%%%%%%%%%%%%%%%%%%%%%%%%%%%%%%%%%%%%%%%%%%%%%%%%%%%%%%%%%%%%%%%%%%%%%%%%%
                                      & \multirow{2}{*}{\textsc{Fisher}+} 
                                                              & \textcolor{brown}{Acc} (\%)      & $72.29\rightarrow 72.73$       & $ 71.71 \rightarrow 72.65 $     \\ 
                                      &                       & \textcolor{cyan}{WN} (\textcolor{teal}{T})    & $29.2 \rightarrow 14.1$~($0.4$ s)   & $35.4 \rightarrow 15.6$~($0.3$ s) \\ \cline{2-5} \rule{0pt}{2ex} 
                                      %   %%%%%%%%%%%%%%%%%%%%%%%%%%%%%%%%%%%%%%%%%%%%%%%%%%%%%%%%%%%%%%%%%%%%%%%%%%%%%%%%%%%%%%%%%%%%%%%%%%%%%%%%%%%%%%%%%%%%%%%%%%%%%%%%%%%%%%%%%%%%
                                    %   %%%%%%%%%%%%%%%%%%%%%%%%%%%%%%%%%%%%%%%%%%%%%%%%%%%%%%%%%%%%%%%%%%%%%%%%%%%%%%%%%%%%%%%%%%%%%%%%%%%%%%%%%%%%%%%%%%%%%%%%%%%%%%%%%%%%%%%%%%%%
                                    %   %%%%%%%%%%%%%%%%%%%%%%%%%%%%%%%%%%%%%%%%%%%%%%%%%%%%%%%%%%%%%%%%%%%%%%%%%%%%%%%%%%%%%%%%%%%%%%%%%%%%%%%%%%%%%%%%%%%%%%%%%%%%%%%%%%%%%%%%%%%%
                                      & \multirow{2}{*}{\begin{tabular}[c]{@{}l@{}} \textsc{Re-training} \\ (+ adapt diff) \end{tabular}}
                                                              & \textcolor{brown}{Acc} (\%)         & $73.42\rightarrow 73.42$       & $ 73.34 \rightarrow 73.40 $     \\ 
                                      &                       & \textcolor{cyan}{WN} (\textcolor{teal}{T})   & $24.3 \rightarrow 0$~($1,973$ s)   & $25.6 \rightarrow 0$~($1,973$ s) \\ \hline\hline \rule{0pt}{2ex} 
\multirow{12}{*}{\rotatebox[origin=c]{90}{\textbf{OGB-Products}}}    
                                      & \cellcolor[HTML]{FEDEDC} & \cellcolor[HTML]{FEDEDC}\textcolor{brown}{Acc} (\%)        & \cellcolor[HTML]{FEDEDC}$79.21\rightarrow 79.22$       & \cellcolor[HTML]{FEDEDC}$79.18\rightarrow 79.11$     \\ 
                                      & \cellcolor[HTML]{FEDEDC}\multirow{-2}{*}{\our}  & \cellcolor[HTML]{FEDEDC}\textcolor{cyan}{WN} (\textcolor{teal}{T}) & \cellcolor[HTML]{FEDEDC}$27.6 \rightarrow 0$~($0.06$ s)   & \cellcolor[HTML]{FEDEDC}$30.8 \rightarrow 0$~($0.06$ s) \\
                                      \cline{2-5} \rule{0pt}{2ex}   
                                      %%%%%%%%%%%%%%%%%%%%%%%%%%%%%%%%%%%%%%%%%%%%%%%%%%%%%%%%%%%%%%%%%%%%%%%%%%%%%%%%%%%%%%%%%%%%%%%%%%%%%%%%%%%%%%%%%%%%%%%%%%%%%%%%%%%%%%%%%%%%
                                      %%%%%%%%%%%%%%%%%%%%%%%%%%%%%%%%%%%%%%%%%%%%%%%%%%%%%%%%%%%%%%%%%%%%%%%%%%%%%%%%%%%%%%%%%%%%%%%%%%%%%%%%%%%%%%%%%%%%%%%%%%%%%%%%%%%%%%%%%%%%
                                      %%%%%%%%%%%%%%%%%%%%%%%%%%%%%%%%%%%%%%%%%%%%%%%%%%%%%%%%%%%%%%%%%%%%%%%%%%%%%%%%%%%%%%%%%%%%%%%%%%%%%%%%%%%%%%%%%%%%%%%%%%%%%%%%%%%%%%%%%%%%
                                      & \cellcolor[HTML]{C4F8F6} & \cellcolor[HTML]{C4F8F6}\textcolor{brown}{Acc} (\%)            & \cellcolor[HTML]{C4F8F6}$79.95\rightarrow 79.93$         & \cellcolor[HTML]{C4F8F6}$79.96\rightarrow 79.91$     \\ 
                                      & \cellcolor[HTML]{C4F8F6}\multirow{-2}{*}{\cellcolor[HTML]{C4F8F6}\begin{tabular}[c]{@{}l@{}} \our \\ (+ adapt diff) \end{tabular}}  & \cellcolor[HTML]{C4F8F6}\textcolor{cyan}{WN} (\textcolor{teal}{T}) & \cellcolor[HTML]{C4F8F6}$16.4 \rightarrow 0$~($0.06$ s)   & \cellcolor[HTML]{C4F8F6}$18.6\rightarrow 0$~($0.06$ s) \\ \cline{2-5} \rule{0pt}{2ex}  
                                      %%%%%%%%%%%%%%%%%%%%%%%%%%%%%%%%%%%%%%%%%%%%%%%%%%%%%%%%%%%%%%%%%%%%%%%%%%%%%%%%%%%%%%%%%%%%%%%%%%%%%%%%%%%%%%%%%%%%%%%%%%%%%%%%%%%%%%%%%%%%
                                      %%%%%%%%%%%%%%%%%%%%%%%%%%%%%%%%%%%%%%%%%%%%%%%%%%%%%%%%%%%%%%%%%%%%%%%%%%%%%%%%%%%%%%%%%%%%%%%%%%%%%%%%%%%%%%%%%%%%%%%%%%%%%%%%%%%%%%%%%%%%
                                      %%%%%%%%%%%%%%%%%%%%%%%%%%%%%%%%%%%%%%%%%%%%%%%%%%%%%%%%%%%%%%%%%%%%%%%%%%%%%%%%%%%%%%%%%%%%%%%%%%%%%%%%%%%%%%%%%%%%%%%%%%%%%%%%%%%%%%%%%%%%
                                      & \multirow{2}{*}{\begin{tabular}[c]{@{}l@{}} \textsc{GraphEraser} \\ ($\times 8$ subgraphs) \end{tabular}} 
                                                              & \textcolor{brown}{Acc} (\%)        & $70.80\rightarrow 70.78$       & $70.80\rightarrow 70.78$     \\ 
                                      &                       & \textcolor{cyan}{WN} (\textcolor{teal}{T}) & $25.4 \rightarrow 0$~($598$ s)   & $28.9 \rightarrow 0$~($598$ s) \\\cline{2-5} \rule{0pt}{2ex}    
                                      %%%%%%%%%%%%%%%%%%%%%%%%%%%%%%%%%%%%%%%%%%%%%%%%%%%%%%%%%%%%%%%%%%%%%%%%%%%%%%%%%%%%%%%%%%%%%%%%%%%%%%%%%%%%%%%%%%%%%%%%%%%%%%%%%%%%%%%%%%%%
                                      %%%%%%%%%%%%%%%%%%%%%%%%%%%%%%%%%%%%%%%%%%%%%%%%%%%%%%%%%%%%%%%%%%%%%%%%%%%%%%%%%%%%%%%%%%%%%%%%%%%%%%%%%%%%%%%%%%%%%%%%%%%%%%%%%%%%%%%%%%%%
                                      %%%%%%%%%%%%%%%%%%%%%%%%%%%%%%%%%%%%%%%%%%%%%%%%%%%%%%%%%%%%%%%%%%%%%%%%%%%%%%%%%%%%%%%%%%%%%%%%%%%%%%%%%%%%%%%%%%%%%%%%%%%%%%%%%%%%%%%%%%%%
                                      & \multirow{2}{*}{\textsc{Influence}+} 
                                                              & \textcolor{brown}{Acc} (\%)       & $72.23\rightarrow 72.78$       & $ 72.08 \rightarrow 72.51 $     \\ 
                                      &                       & \textcolor{cyan}{WN} (\textcolor{teal}{T}) & $8.9 \rightarrow 3.1$~($1.7$ s)   & $14.3 \rightarrow 4.2$~($1.9$ s) \\\cline{2-5} \rule{0pt}{2ex}    
                                      
                                      & \multirow{2}{*}{\textsc{Fisher}+} 
                                                              & \textcolor{brown}{Acc} (\%)      & $72.23\rightarrow 72.78$       & $ 72.08 \rightarrow 72.51 $     \\ 
                                      &                       & \textcolor{cyan}{WN} (\textcolor{teal}{T}) & $8.9 \rightarrow 3.1$~($1.3$ s)   & $14.3 \rightarrow 4.2$~($1.1$ s) \\ \cline{2-5} \rule{0pt}{2ex} 
                                      & \multirow{2}{*}{\begin{tabular}[c]{@{}l@{}} \textsc{Re-training} \\ (+ adapt diff) \end{tabular}}
                                                              & \textcolor{brown}{Acc} (\%)      & $79.95\rightarrow 79.74$       & $ 79.96\rightarrow 79.71 $     \\ 
                                      &                       & \textcolor{cyan}{WN} (\textcolor{teal}{T}) & $16.4 \rightarrow 0$~($661$ s)   & $18.6 \rightarrow 0$~($661$ s) \\ \hline\hline
\end{tabular}
}
\vspace{-3mm}
\end{table}

\noindent\textbf{Closeness to retraining from scratch.}
We compare the closeness of the unlearned solution  $\mathbf{w}_p$ to the retrained model $\mathbf{w}_u$ to verify our conclusion in Theorem~\ref{theorem:ell2_norm_of_weight_projections} and Proposition~\ref{proposition:closeness_of_projector_vs_inflence}.
We measure the difference between normalized weight parameters $\|\mathbf{w}_u - \mathbf{w}_p\|_2/\|\mathbf{w}\|_2$ and distance between the final activations $\mathbb{E}_{v_i \in \mathcal{B}} [ \| \sigma(\mathbf{w}_p^\top \mathbf{h}_i) - \sigma(\mathbf{w}_u^\top \mathbf{h}_i) \|_2 ]$ where $\mathcal{B} \in \{\mathcal{V}_\text{delete}, \mathcal{V}_\text{remain}, \mathcal{V}_\text{test}\}$.
Ideally, a powerful unlearning algorithm is expected to generate similar final weight parameters and activations to the retrained model.
We randomly select $1\%$ of the nodes from the training set as the deleted nodes $\mathcal{V}_\text{delete} \subset \mathcal{V}_\text{train}$ and the rest as remain nodes $\mathcal{V}_\text{remain} = \mathcal{V}_\text{train}\setminus \mathcal{V}_\text{delete}$.
As shown in Figure~\ref{fig:weight_diff_cora_citeseer_pubmed}, both the final activation (column 1, 2, 3) and the output parameters (column 4) of \our (blue curve) is closer to the weight obtained by retraining from scratch compared to baseline methods, which could reflect our result in  Proposition~\ref{proposition:closeness_of_projector_vs_inflence}.
Besides, we can observe that lower unlearning percentage leads to a smaller difference on the output weight parameters of \our (blue curve in column 4), which could reflect our theoretical result in  Theorem~\ref{theorem:ell2_norm_of_weight_projections}.

\noindent\textbf{Compare to non-linear models.}
We compare the performance of non-linear GNNs (introduced in Section~\ref{section:extension}) and linear GNNs, where the MLP extractor in non-linear \our is pre-trained by supervised learning on the features of all training set nodes but except the deleted ones. We have the following observations from Table~\ref{table:linear_gnn_with_its_non_linear_extension}: \circled{1} By comparing results in block 1, we know that using MLP as a feature extractor can improve the average F1-score accuracy, but it also increases the variance of the model performance; \circled{2} By comparing the results in block 1 and 2, we know that linear-GNN could achieve better performance than ordinary GNNs; \circled{3} By comparing results in block 2 and 3, we know that employing \textsc{GraphEraser} with non-linear GNNs will significantly hurt the performance of the original GNN models, which is due to the data heterogeneously and the lack of training data for each subgraph model.

\begin{table}[t]
\centering
\caption{
Comparison on the performance of linear GNN and its non-linear extension with ordinary GNNs.} \label{table:linear_gnn_with_its_non_linear_extension}
\vspace{-4mm}
\centering\setlength{\tabcolsep}{1mm}
\scalebox{0.88}{
\begin{tabular}{l l l l }
\hline\hline
                           & & Method           &  Accuracy  \\ \hline\hline \rule{0pt}{2ex} 
\multirow{6}{*}{OGB-Arxiv} & \multirow{2}{*}{\circled{1}} & Linear GNN + Adap diff      & $73.35 \pm 0.12$ \\ % \cline{2-4} 
                           & & Linear GNN + Adap diff + MLP & $73.41 \pm 0.31$ \\ \cline{2-4} \rule{0pt}{2ex} 
                           & \multirow{2}{*}{\circled{2}} & GCN              &  $71.74 \pm 0.29$ \\ % \cline{2-4} 
                           & & GraphSAGE        &  $71.49 \pm 0.27$ \\ \cline{2-4} \rule{0pt}{2ex} 
                           & \multirow{2}{*}{\circled{3}} & GCN + \textsc{GraphEraser}             & $66.52 \pm 0.31$ \\ % \cline{2-4} 
                           & & GraphSAGE + \textsc{GraphEraser}       & $62.96 \pm 0.26$
                           \\ \hline\hline \rule{0pt}{2ex} 
\multirow{8}{*}{OGB-Product} & \multirow{2}{*}{\circled{1}} & Linear GNN + Adap diff        & $80.25 \pm 0.09$ \\ % \cline{2-4} 
                             & & Linear GNN + Adap diff + MLP &  $80.30 \pm 0.40$ \\ \cline{2-4} \rule{0pt}{2ex}  
                             & \multirow{3}{*}{\circled{2}} & GAT              &  $79.45 \pm 0.59$ \\ % \cline{2-4} 
                             & & GraphSAGE        &  $78.70 \pm 0.36$ \\ % \cline{2-4}
                             & & GraphSaint       &  $79.08 \pm 0.24$ \\ \cline{2-4} \rule{0pt}{2ex} 
                             & \multirow{3}{*}{\circled{3}} & GAT + \textsc{GraphEraser}              & $60.23 \pm 0.71$ \\ % \cline{2-4} 10289.37s
                             & & GraphSAGE + \textsc{GraphEraser}        & $58.99 \pm 0.40$ \\ % \cline{2-4}
                             & & GraphSaint + \textsc{GraphEraser}       & $59.54 \pm 0.41$ \\ \hline\hline
                           
\end{tabular}}
% \vspace{-3mm}
\end{table}

\begin{figure}[t]
    \centering
    % \vspace{-4mm}
    \includegraphics[width=0.49\textwidth]{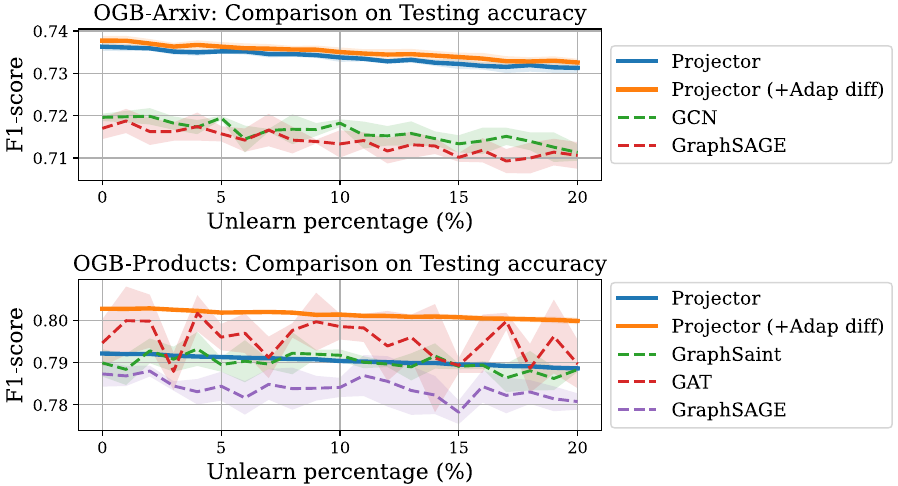}
    \vspace{-4mm}
    \caption{Comparison on the test performance with different number of node to unlearn. }
    \label{fig:test_perf}
    \vspace{-4mm}
\end{figure}

\noindent\textbf{Robustness of \oure.}
We study the change of testing accuracy as we progressively increase the unlearning ratio from $1\%$ to $20\%$, where a more stable model performance is preferred in real-world scenarios.
As shown in Figure~\ref{fig:test_perf}, the change of testing accuracy in \our is smaller (e.g., on the OGB-Arvix dataset the test accuracy of \our changes around $0.5\%$ while the GNNs change around $0.8\%\sim1\%$ ), more stable (i.e., the test accuracy fluctuate less when the fraction of unlearning nodes increases), and with accuracy even better than re-training ordinary GNNs.

\noindent\textbf{Evaluation on the $\delta$ term in Assumption~\ref{assumption:upper_bound_on_approximation_error}.} 

The performance of \oure's unlearned solution is highly dependent on the correlation between node features, which is captured by the $\delta$ term in Assumption~\ref{assumption:upper_bound_on_approximation_error}. Therefore, we report the $\delta$ by computing
\begin{equation}
    \delta = \max_{v_i \in \mathcal{V}} \left\| \mathbf{x}_i - \mathbf{X}_\text{remain}^\top \mathbf{X}_\text{remain} (\mathbf{X}_\text{remain}^\top \mathbf{X}_\text{remain} )^\dagger \mathbf{x}_i \right\|_2,
\end{equation}
where $\mathbf{X}_\text{remain} = [\mathbf{x}_1, \ldots, \mathbf{x}_{i-1}, \mathbf{x}_{i+1},\ldots, \mathbf{x}_n]$ is the stack of all remaining node features.
As shown in Table~\ref{table:delta_on_real_world}, the $\delta$ value is relatively small compared to the norm of average node features, which indicates the realism of our assumption and guarantees the performance of \oure's unlearned solution (even without finetuning).
Besides, we can observe that the $\delta$ value on the Cora dataset is larger than other datasets, this is because the feature of the Cora dataset is a binary-valued vector of size $1433$ which is very close to the total number of nodes in the graph ($2708$ nodes). When the node feature dimension is large and all values are either $0$ or $1$, representing any vectors with others becomes difficult, therefore resulting in a larger $\delta$.

\begin{table}[h]
\centering
\caption{Evaluation $\delta$ on real-world datasets.}
\vspace{-3mm}
\label{table:delta_on_real_world}
\centering\setlength{\tabcolsep}{1mm}
\scalebox{0.93}{
\begin{tabular}{cllll} 
\hline\hline
         & OGB-Arvix & OGB-Product & Cora & Pubmed \\ 
$\delta$ &  $0.3815$ &  $0.0915$  &   $0.2984$   &  $0.0049$      \\ 
$\| \frac{1}{n} \sum_{i=1}^n \mathbf{x}_i \|_2$ 
         &   $9.6369$  &  $161.4997$ &  $0.6923$    &   $0.5546$     \\ \hline\hline
\end{tabular}
}
\end{table}

\noindent\textbf{More experiment results.} More experiment results are deferred to the appendix. We compare \our with re-training non-linear GNNs under different node deletion schemes in Appendix~\ref{section:more_schemes}. We evaluate unlearning with membership inference attack in Appendix~\ref{section:membership_inference}. We ablation study the effectiveness of fine-tuning on \our in Appendix~\ref{section:finetuning_perf}. 
% We empirical evaluate our assumption on the closeness of projection-step $\delta$ in Appendix~\ref{section:eval_on_delta}.
% and evaluate the conditions on node features and graph propagation matrix of Proposition~\ref{prop:linear_as_expressive_as_nonlinear} in Appendix~\ref{}.

\section{Conclusion} \label{section:conclusion}
In this paper, we study graph representation unlearning by proposing a projection-based unlearning approach \oure. \our unlearns the deleted node features by projecting the weight parameters of a pre-trained model onto a subspace that is irrelevant to the deleted node features.
Empirical results on real-world dataset illustrate its effectiveness, efficiency, and robustness.

% \clearpage
\section*{Acknowledgements}
This work was supported in part by NSF grant 2008398.

\bibliography{reference}
\bibliographystyle{apalike}

\clearpage
\appendix\onecolumn

\renewcommand{\baselinestretch}{0.4}\normalsize
\tableofcontents

\renewcommand{\baselinestretch}{1.0}\normalsize

\clearpage
\section{More experiment results} \label{section:more_experiment_results}
In this section, we provide more empirical evaluation results. 
\begin{itemize} 
    \item Appendix~\ref{section:more_schemes}: We compare \our with re-training non-linear GNNs under different node deletion schemes.
    % \item Appendix~\ref{section:robustness}: We study the change of testing accuracy as we progressively increase the unlearning ratio. 
    \item Appendix~\ref{section:membership_inference}: We conduct experiments using membership inference attack framework.
    \item Appendix~\ref{section:finetuning_perf}: We ablation study the effectiveness of fine-tuning on \oure's unlearned model.
    % \item \textcolor{red}{Appendix~\ref{section:eval_on_delta}: We empirically evaluate our assumption on the closeness of projection-step in Assumption~\ref{assumption:upper_bound_on_approximation_error}.}
\end{itemize}
 
Code to reproduce the experiment results can be found at [\href{https://github.com/CongWeilin/Projector}{\textcolor{blue}{Repository}}].

\subsection{Linear vs non-linear GNN under different deleted nodes selection schemes}\label{section:more_schemes}

In this section, we compare the model performance of \our against re-training 2-layer GNNs from scratch under different unlearning settings.
We consider GCN~\cite{kipf2016semi}, GraphSAGE~\cite{hamilton2017inductive}, APPNP~\cite{klicpera2018predict}, and GAT~\cite{velivckovic2017graph} as the baseline 2-layer GNNs. Notice that using a 2-layer is a common choice in graph representation learning to balance between computation cost, training accuracy, and generalization.
To scale for large-graph training, we utilize the K-hop shadow sampler~\cite{shaDow} implemented in Pytorch Geometric~\cite{Fey/Lenssen/2019}.
We consider two different unlearning settings, i.e., unlearning $10\%$ training set nodes with the largest node degree (i.e., ``delete dense nodes'' in Table~\ref{table:compare_non_linear_diff_settings}) and unlearning $10\%$ training set nodes with the smallest node degree (i.e., ``delete sparse nodes'' in Table~\ref{table:compare_non_linear_diff_settings}), to simulate the potential real-world node deletion scenario.

\begin{table}[h]
\centering
\caption{Compare the model performance \our with re-training 2-layer GNNs from scratch under different deleted nodes selection schemes.}\label{table:compare_non_linear_diff_settings}
\vspace{-3mm}
\scalebox{0.85}{
\begin{tabular}{ll l lll}
\hline\hline
                      &            &            & Before node deletion & Delete dense nodes & Delete sparse nodes \\ \hline\hline
\multirow{6}{*}{Flickr (avg node degree 10)} & GCN & Re-train       & $50.01 \pm 0.13$ (241.08s) & $49.35 \pm 0.10$ (194.75s) & $49.79 \pm 0.18$ (234.93s) \\ 
                      & GraphSAGE & Re-train   & $51.34 \pm 0.14$ (243.00s) & $50.23 \pm 0.22$ (195.49s) & $50.93 \pm 0.20$ (236.77s) \\ 
                      & APPNP & Re-train       & $50.04 \pm 0.09$ (244.53s) & $49.03 \pm 0.17$ (200.34s) & $49.60 \pm 0.06$ (239.71s) \\ 
                      & GAT & Re-train         & $51.01 \pm 0.11$ (409.51s) & $49.78 \pm 0.23$ (313.75s) & $50.76 \pm 0.22$ (392.77s) \\  \cline{2-6} \rule{0pt}{2ex} 
                      & & Re-train  &  & $49.21 \pm 0.23$ (268.09s) & $51.84 \pm 0.13$ (316.76s) \\
                      & \multirow{-2}{*}{Linear-GNN} & Projector & \multirow{-2}{*}{$\mathbf{52.71} \pm 0.14$ (374.50s)} & $\mathbf{50.63} \pm 0.19$ (0.06s)   & $\mathbf{51.90} \pm 0.11$ (0.05s) \\  \hline
\multirow{6}{*}{Reddit (avg node degree 50)} 
                      & GCN & Re-train         & $93.04 \pm 0.09$ (1174.12s) & $92.95 \pm 0.06$ (910.51s)  & $91.47 \pm 0.09$ (1129.89s) \\ 
                      & GraphSAGE & Re-train   & $94.68 \pm 0.06$ (1005.13s) & $94.90 \pm 0.07$ (775.80s)  & $94.45 \pm 0.06$ (917.80s) \\ 
                      & APPNP & Re-train       & $93.73 \pm 0.06$ (1010.02s) & $94.08 \pm 0.10$ (780.10s)  & $93.09 \pm 0.08$ (925.56s) \\ 
                      & GAT & Re-train         & $92.82 \pm 0.10$ (1431.56s) & $93.12 \pm 0.05$ (1228.95s) & $92.41 \pm 0.10$ (1297.23s)\\  \cline{2-6} \rule{0pt}{2ex} 
                      &  & Re-train  & & $95.03 \pm 0.03$ (1290.16s) & $94.58 \pm 0.02$ (1532.51s) \\
                      & \multirow{-2}{*}{Linear-GNN} & Projector & \multirow{-2}{*}{$\mathbf{94.72} \pm 0.08$ (1630.53s)} & $\mathbf{95.09} \pm 0.03$ (0.24s)    & $\mathbf{94.66} \pm 0.07$ (0.32s) \\  \hline
\hline

\end{tabular}
}
\end{table}

We have the following observations From Table~\ref{table:compare_non_linear_diff_settings}:
\begin{itemize}
    \item Linear-GNNs could achieve even better performance than non-linear GNNs. For example, linear-GNN is around $0.1\sim 1\%$ better than 2-layer GNNs on Flickr and Reddit dataset. This also verifies the arguments in Proposition~\ref{prop:linear_as_expressive_as_nonlinear} that the expressive power of linear-GNNs mainly comes from its weight combination of multi-hop graph convolution operators. Besides, since linear-GNN has lower model complexity, it could generalize better than multi-layer GNNs.
    \item Interestingly, according to the second column of our results, we found that removing dense nodes on the sparser graph (e.g., Flickr) hurt the model performance to around $1\sim2\%$, however, removing dense nodes on the denser graph (e.g., Reddit) it will improves the model performance to around $1\%$. This is potentially because those dense nodes provide too much redundant information on the denser graph than on a sparser graph or due to the over-smoothing~\cite{xu2018powerful} issue caused by aggregating too many neighbors in the original graph.
    \item The performance of \our is around $0.06\sim 0.1\%$ better than re-training on the graph without the deleted nodes. This is potential because the output of \our could be interpreted as re-training on the remaining graph under some unknown importance sampling distribution, while this importance distribution help \our learn better from the remaining data.
\end{itemize}

% # python run_gnn.py --dataset reddit --batch_size 1024 --lr 0.001 --net GAT
% # original acc: $92.82 \pm 0.10$ (1431.56s)
% # large acc: $93.12 \pm 0.05$ (1228.95s)
% # small acc: $92.41 \pm 0.10$ (1297.23s)

% \clearpage
% \input{supplementary/nonlinear}
% \input{supplementary/robusness_projector}
% \clearpage

\subsection{Evaluation by Membership Inference Attack}\label{section:membership_inference}

In this section, we conduct membership inference attack~\cite{olatunji2021membership} to test whether GNN models could potentially leak information about the deleted nodes' membership information and whether \our could alleviate the information leakage issue. 
In the following, we will first give a brief introduction on the GNN membership inference attack settings used in~\cite{olatunji2021membership} then provide details on our experiment results.

\begin{figure}[h]
    \centering
    \includegraphics[width=0.8\textwidth]{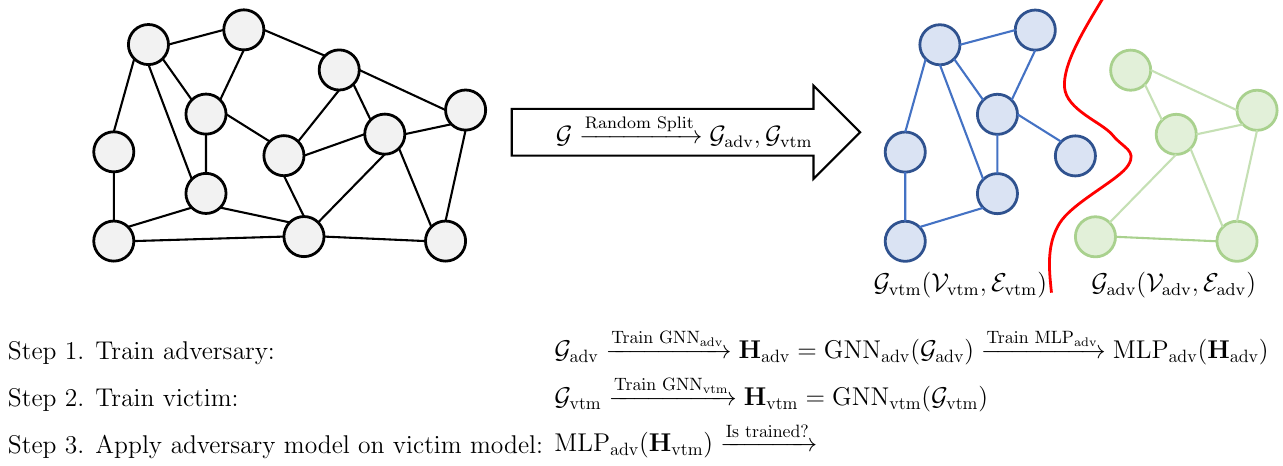}
    \caption{An overview on the workflow of membership inference attack.}
    \label{fig:membership_inference_attack}
\end{figure}

\noindent\textbf{Membership inference attack setting.}
An overview of the membership inference attack is introduced in Figure~\ref{fig:membership_inference_attack}, which follows the implementation of the graph membership inference attack that is proposed in~\cite{olatunji2021membership}.
Let us denote $\mathcal{G}(\mathcal{V},\mathcal{E})$ as the full graph.
In the membership inference attack, a necessary assumption is the graphs used by adversaries and victims are from the same distribution. To achieve this goal, we randomly split all nodes into adversary's node sets $\mathcal{V}_\text{adv} \subset \mathcal{V}$ and victim's node sets $\mathcal{V}_\text{vtm} = \mathcal{V} \setminus\mathcal{V}_\text{adv}$. 
Then, we define the subgraph induced by the two node sets as adversary subgraph $\mathcal{G}_\text{adv}$ and victim subgraph $\mathcal{G}_\text{vtm}$. This process is illustrated at the top of Figure~\ref{fig:membership_inference_attack}.
On the adversary side, the adversary first pre-trains a GNN model $\text{GNN}_\text{adv}$ on $\mathcal{G}_\text{adv}$ by only using a subset of nodes as training sets and extract the node representation as $\mathbf{H}_\text{adv}$. Then, a binary classifier $\text{MLP}_\text{adv}$ is trained on $\mathbf{H}_\text{adv}$ to classify whether a node has been used for training for not. 
On the victim side, the victim only need to train a GNN model $\text{GNN}_\text{vtm}$ on $\mathcal{G}_\text{vtm}$ by only use a subset of nodes as training sets and extract the node representation as $\mathbf{H}_\text{vtm}$. 
During the membership inference attack, the adversary applies $\text{MLP}_\text{adv}$ onto $\mathbf{H}_\text{vtm}$ to distinguish if a node is used for training.
Membership inference attack is more challenging on graph data because the adversary classifier $\text{MLP}_\text{adv}$ is applied to the node representation, and the node representation of some nodes might be very similar if they share the same neighborhood information.

\begin{table}[h]
\centering
\caption{Comparison on the membership inference attack accuracy before and after unlearning.}\label{table:membership_inference_attack}
\vspace{-3mm}
\scalebox{0.85}{
\begin{tabular}{llcccc}
\hline\hline
                          & Method~/~Phase         & $\mathcal{V}_\text{delete}$ as untrained & $\mathcal{V}_\text{train}^\text{after}$ as trained &  $\mathcal{V} \setminus \mathcal{V}_\text{train}^\text{before}$ as untrained & $\mathcal{V}_\text{train}^\text{before}$ \\ \hline\hline
\multirow{3}{*}{Cora} & \textsc{Before-unlearn} & \cellcolor[HTML]{C4F8F6}$ 46.35 \pm 16.99 $ & \cellcolor[HTML]{66DDAA}$ 52.31 \pm 16.31 $ & \cellcolor[HTML]{66DDAA}$ 63.17 \pm 12.12 $ & $ 57.79 \pm 2.84 $  \\ 
                          & \textsc{Re-training}    & \cellcolor[HTML]{FEDEDC}$59.37 \pm 11.82$ & $56.30 \pm 16.56$ & $62.98 \pm 14.01$ & $58.79 \pm 1.83$          \\ 
                          & \oure      & \cellcolor[HTML]{FEDEDC}$ 58.38 \pm 11.75 $ & \cellcolor[HTML]{FDD5B1}$ 51.64 \pm 16.23 $ & \cellcolor[HTML]{FDD5B1}$ 63.84 \pm 12.14 $ & $ 57.76 \pm 2.15 $  \\ \hline\hline
\multirow{3}{*}{Citeseer}     & \textsc{Before-unlearn} & \cellcolor[HTML]{C4F8F6}$47.33 \pm 31.21$ & \cellcolor[HTML]{66DDAA}$38.52 \pm 32.73$ & \cellcolor[HTML]{66DDAA}$68.30 \pm 29.97$ & $54.97 \pm 2.32$ \\ 
                          & \textsc{Re-training}    & \cellcolor[HTML]{FEDEDC}$66.33 \pm 25.60$ & $54.04 \pm 30.86$ & $55.53 \pm 28.94$ & $55.20 \pm 3.08$ \\ 
                          & \oure      & \cellcolor[HTML]{FEDEDC}$67.33 \pm 27.50$ & \cellcolor[HTML]{FDD5B1}$38.22 \pm 32.68$ & \cellcolor[HTML]{FDD5B1}$68.97 \pm 29.56$ & $55.05 \pm 2.17$ \\ \hline\hline
\end{tabular}
}
\end{table}

\noindent\textbf{Results.} 
We compare the accuracy of $\text{MLP}_\text{adv}$ classifies each node representation $\mathbf{H}_\text{vtm}$ as training or non-training set nodes before and after unlearning. Let us denote $\mathcal{V}_\text{train}^\text{before} \subset \mathcal{V}_\text{victm}$ as the subset of nodes used for training before node deletion, denote $\mathcal{V}_\text{train}^\text{after} \subset \mathcal{V}_\text{train}^\text{before}$ as the subset of nodes used for training after node deletion, and denote $\mathcal{V}_\text{delete} = \mathcal{V}_\text{train}^\text{before} \setminus \mathcal{V}_\text{train}^\text{after}$ as the nodes for deletion. 
Two victim models are trained on training set nodes $\mathcal{V}_\text{train}^\text{before}, \mathcal{V}_\text{train}^\text{after}$.
We report the accuracy on $\mathcal{V}_\text{train}^\text{after}, \mathcal{V}_\text{delete}, \mathcal{V}_\text{vtm} \setminus \mathcal{V}_\text{train}^\text{before}$ before and after unlearning. We are using the re-trained model as a baseline.
In terms of the size of node sets, we set $|\mathcal{V}_\text{train}^\text{before}| = 0.5 \times |\mathcal{V}_\text{victim}|$ and $| \mathcal{V}_\text{delete} | = 0.9 \times |\mathcal{V}_\text{train}^\text{before}|$. We have the following observations from Table~\ref{table:membership_inference_attack}:
\begin{itemize}
    \item By comparing the \colorbox[HTML]{C4F8F6}{\textsc{Before-unlearn}} with \colorbox[HTML]{FEDEDC}{\textsc{Re-training}} and \colorbox[HTML]{FEDEDC}{\oure}, we know that both re-training and our proposal could increase the probability that $\text{MLP}_\text{adv}$ classify $\mathcal{V}_\text{delete}$ at ``untrained''. More specifically, when applying $\text{MLP}_\text{adv}$ on the model before-unlearning, since $\mathcal{V}_\text{delete}$ are used before unlearning, the probability of classifying $\mathcal{V}_\text{delete}$ as ``untrained'' should be lower than $50\%$. However, after re-training or using our unlearning approach, the probability of classifying $\mathcal{V}_\text{delete}$ as ``untrained'' increases as the information on $\mathcal{V}_\text{delete}$ are removed during the unlearning process.
    \item By comparing the \colorbox[HTML]{66DDAA}{\textsc{Before-unlearn}} with \colorbox[HTML]{FDD5B1}{\textsc{Projector}} at each column, we can observe that the prediction of $\text{MLP}_\text{adv}$ on $\mathcal{V}_\text{train}^\text{after}$ and $\mathcal{V}\setminus \mathcal{V}_\text{train}^\text{before}$ are almost unchanged. This is also expected as our projection step only remove the trace of the deleted nodes and will preserve its model performance/behavior as much as possible according to our method design in Section~\ref{section:method}.
\end{itemize}

% \clearpage
\subsection{Performance Before and After Finetuning} \label{section:finetuning_perf}
% \noindent\textbf{Performance before and after finetuning.~}
The experiment results in Section~\ref{section:experiments} are reported  without the fine-tuning process as mentioned in Theorem~\ref{theorem:ell2_norm_of_weight_projections}. For the completeness of our discussion, we provide further results on the comparison of the training, validation, and testing accuracy of the unlearned model both with and without the fine-tuning process.

\noindent\textbf{Setup.}
In this experiment, we randomly select $1\%$ of the nodes from the training set to unlearn. During both training and fine-tunings, we early stop if the validation accuracy does not increase within $10$ iterations on the OGB-Arvix dataset and $1,000$ iterations on the OGB-Products dataset. We repeat the experiment $5$ times. Other setup remains the same as introduced in Section~\ref{section:experiment_details}.

\begin{figure}[h]
    \centering
    \includegraphics[width=\textwidth]{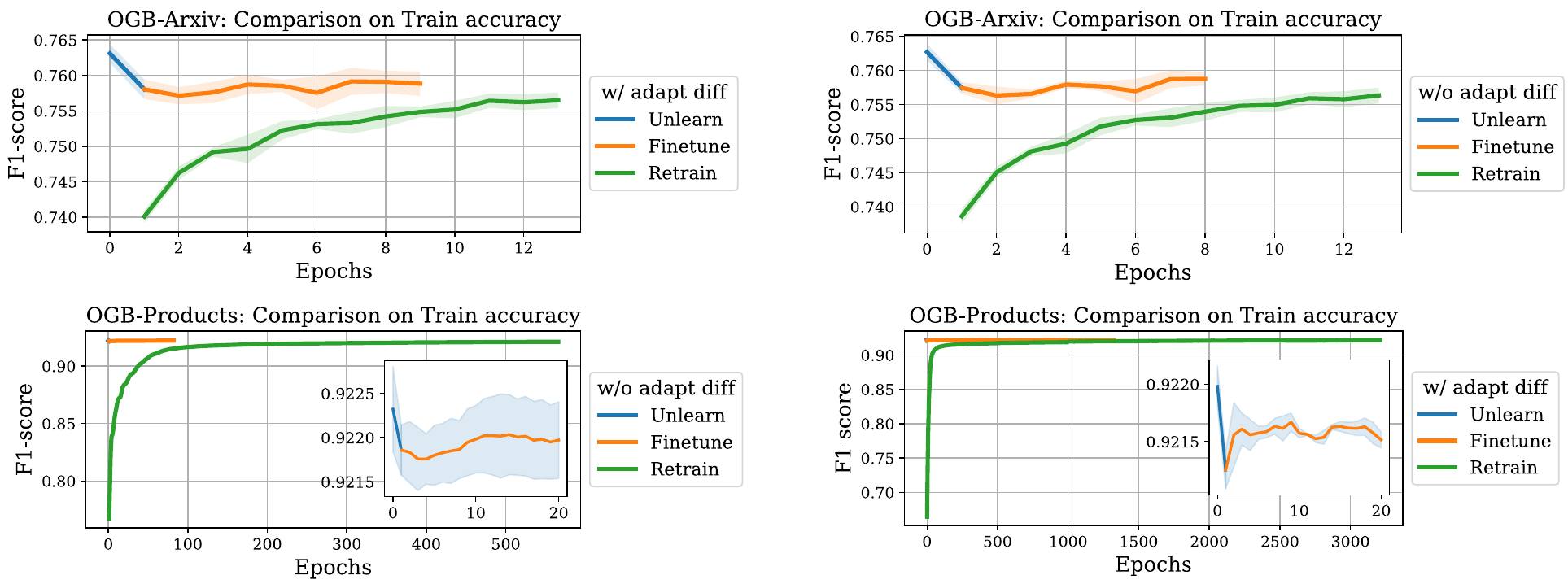}
    \vspace{-6mm}
    \caption{Evaluation on effect of finetuning on the training accuracy before and after $1\%$ of the node from training set deleted.}
    \label{fig:proj_retrain_arxiv_train_valid}
\end{figure}

\noindent\textbf{Results.}
According to our result in Figure~\ref{fig:proj_retrain_arxiv_train_valid}, we have the following observations: \circled{1} By looking at the \textcolor{blue}{blue curve}, we know that both the training and validation accuracy dropped after unlearning, which is expected as part of the information related to the deleted nodes are removed; \circled{2} By looking at the \textcolor{orange}{orange curve}, we can observe that the fine-tuning training accuracy indeed improves progressively but the improvement is relatively small, 
this is because the solution after \our unlearning is already close to the optimal solution, which could be partially explained by the hypothesis that \textit{small changes on the dataset will not results in massive changes on the optimal solution}. \circled{3}
By comparing the \textcolor{orange}{orange curve} and \textcolor{green}{green curve}, we know that fine-tuning on the unlearned solution (orange curve) could save a lot of time comparing to re-training from scratch (green curve).
Furthermore, we compare the F1-score on the test set in Table~\ref{table:finetune_results} and have the following observations: \circled{1} When without the adaptive diffusion operation, the generalization performance on the testing set between fine-tuning to re-training are relatively close; \circled{2} However, if using the adaptive diffusion operation, the unlearning solution (no matter with or without the fine-tuning step) always outperform re-training from stretch. This is potentially because more data are used to tune the scatter parameters in the adaptive diffusion, which leads to a better generalization ability; \circled{3} The performance before and after the fine-tuning is relatively close, which indicates the impressive generalizability of our unlearning solution.

%%%%%%%%%%%%%%%%%%%%%%%%%%%%%%%%%%%%%%%%%%%%%%%%%%%%%%%%%
\begin{table}[h]
\centering
\caption{Comparison of F1-score on the testing set before unleanring, after unlearning, after fine-tuning, and re-training with $1\%$ of the node from training set deleted on OGB datasets.} \label{table:finetune_results}
\scalebox{0.86}{
\begin{tabular}{l l c | l l c }
\hline\hline
\textbf{Arxiv}             & Status            & Test F1-score (\%) & \textbf{Products}             & Status            & Test F1-score (\%) \\ \hline\hline
\multirow{4}{*}{\our}          & Before unlearning & $73.25 \pm 00.23$  & \multirow{4}{*}{\our}          & Before unlearning & $79.11 \pm 00.08$  \\ 
                               & After unlearning  & $72.97 \pm 00.24$  &                                & After unlearning  & $78.96 \pm 00.06$  \\ 
                               & After fine-tuning & $73.03 \pm 00.11$  &                                & After fine-tuning & $79.06 \pm 00.06$  \\ \cline{2-3} \cline{5-6}
                               & Re-training       & $73.02 \pm 00.11$  &                                & Re-training       & $78.78 \pm 00.14$  \\ \hline\hline
\multirow{4}{*}{\begin{tabular}[c]{@{}l@{}} \our \\ (+ adapt diff) \end{tabular}}  & Before unlearning & $73.35 \pm 00.12$ & \multirow{4}{*}{\begin{tabular}[c]{@{}l@{}} \our \\ (+ adapt diff) \end{tabular}}  & Before unlearning & $80.25 \pm 00.09$ \\ 
                                                                                   & After unlearning  & $73.09 \pm 00.12$ &                                                                                    & After unlearning  & $79.95 \pm 00.12$ \\ 
                                                                                   & After fine-tuning & $73.13 \pm 00.12$ &                                                                                    & After fine-tuning & $80.02 \pm 00.37$ \\ \cline{2-3} \cline{5-6}
                                                                                   & Re-training       & $73.00 \pm 00.12$ &                                                                                    & Re-training       & $79.87 \pm 00.47$ \\ \hline\hline
\end{tabular}
}
\end{table}
\section{Missing details from Section~\ref{section:experiments} (experiment section)}

\subsection{Details on baseline methods} \label{section:baseline_details}

In this paper, we consider exact unlearning method~\textsc{GraphEraser}~\cite{chen2021graph}, approximate unlearning method \textsc{Influence+}~\cite{guo2020certified} and \textsc{Fisher+}~\cite{golatkar2020eternal} as baseline methods.

\noindent\textbf{Details on \textsc{GraphEraser}.}
\textsc{GraphEraser} is an exact unlearning method.
\textsc{GraphEraser} proposes to split the original graph into multiple shards (i.e., subgraphs) and train an independent model on each data shard. During inference, \textsc{GraphEraser} averages the prediction of each shard model as the final prediction. Upon receiving unlearning requests, \textsc{GraphEraser} only needs to re-train the specific shard model where the deleted data belongs to. 
In the experiment, we split all nodes into $8$ shards using graph partition algorithm~\href{http://glaros.dtc.umn.edu/gkhome/metis/metis/overview}{\texttt{METIS}} and use mean average for model aggregation. Each shard model is trained with enough epochs and we return the epoch model with the highest validation score. 
Our implementation is based on their official implementation\footnote{\url{https://github.com/MinChen00/Graph-Unlearning}}~and is general enough to captured the main spirit of \textsc{GraphEraser}, i.e., split data into multiple shards and train a shard model on each shard. METIS allows us to split the original graph into multiple subgraphs while preserving the original graph structure as much as possible.
We also test their official implementation with different model aggregation and graph partition strategies, but their performance is not as good as the \texttt{METIS} partitioning and mean aggregation on the more challenging OGB datasets. 

\noindent\textbf{Details on \textsc{Influence+}.}
\textsc{Influence+} is approximate unlearning method and is implemented based on its official code\footnote{\url{https://github.com/facebookresearch/certified-removal}}.
\textsc{Influence+} proposes to unlearn by removing the influence of the deleted data on the model parameters.
Formally, let $\mathcal{D}_d \subset \mathcal{D}$ denote the deleted subset of training data, $\mathcal{D}_r = \mathcal{D}\setminus \mathcal{D}_d$ denote the remaining data, $\mathcal{L}(\mathbf{w})$ is the objective function, and  $\mathbf{w}$ is the model parameters before unlearning.
Then, \textsc{Influence+}~unlearn by  $\mathbf{w}^u = \mathbf{w} + \mathbf{H}_r^{-1} \mathbf{g}_d$, which is derived from the first-order Taylor approximation on gradient, where $\mathbf{w}^u$ is the parameters after unlearning, $\mathbf{H}_r=\nabla^2 \mathcal{L}(\mathbf{w}, \mathcal{D}_r )$ is the Hessian computed on the remaining data, and $\mathbf{g}_d = \nabla \mathcal{L}(\mathbf{w}, \mathcal{D}_d)$ is the gradient computed on the deleted data.
To mitigate the potential information leakage,
\textsc{Influence+} utilizes a perturbed objective function $ \mathbf{L}(\mathbf{w}) + \mathbf{b}^\top \mathbf{w}$, where $\mathbf{b}$ is the random noise.
\textsc{Influence+} requires the loss function as logistic regression, we use the one-vs-rest strategy splits the multi-class classification into one binary classification problem per class and train with logistic regression.
Besides, \textsc{Influence+} requires the \textit{i.i.d.} data and cannot handle graph structured data, we opt to update both the deleted and affected nodes in parallel. A reader who is interesting the mathematically details could refer to Section~\ref{section:dependency_issue_in_finite_sum}.

\noindent\textbf{Details on \textsc{Fisher+}.}
\textsc{Fisher+} is approximate unlearning method and is edited based on their official code\footnote{\url{https://github.com/AdityaGolatkar/SelectiveForgetting}}.
\textsc{Fisher+} performs Fisher forgetting by taking a single step of Newton's method on the remaining training data, then performing noise injection to model parameters to mitigate the potential information leaking.
The model parameters after unlearning is given by 
$\smash{\mathbf{w}^u = \mathbf{w} - \mathbf{H}_r^{-1} \mathbf{g}_r + \mathbf{H}_r^{-1/4} \mathbf{b}},$
where $\mathbf{H}_r=\nabla^2 \mathcal{L}(\mathbf{w}, \mathcal{D}_r )$ is Hessian and $\mathbf{g}_r=\nabla \mathcal{L}(\mathbf{w}, \mathcal{D}_r )$ is gradient computed on the remaining data $\mathcal{D}_r$, and $\mathbf{b}$ is the random noise.
% Besides, \textsc{Fisher+}-based unlearning requires the \textit{i.i.d.} data and cannot handle graph structured data, we opt to remove both the deleted and affected nodes.

\noindent\textbf{Details on multi-layer GNNs.}
For experiments on OGB datasets, we take their code from the Open Graph Benchmark's online public implementation and use the same hyper-parameters as originally provided.
For example, the implementations on OGB-Arxiv is based on the code at \href{https://github.com/snap-stanford/ogb/tree/master/examples/nodeproppred/arxiv}{\textcolor{blue}{here}} and the implementation on OGB-Products is based on code at \href{https://github.com/snap-stanford/ogb/tree/master/examples/nodeproppred/products}{\textcolor{blue}{here}}.
For experiments on other datasets, we take the example code from PyTorch Geometric at \href{https://github.com/pyg-team/pytorch_geometric/tree/master/examples}{\textcolor{blue}{here}} and use the same hyper-parameters as originally provided.

\subsection{Details on experiment setups} \label{section:experiment_details}

\noindent\textbf{Experiment environment.}
We conduct experiments on a single machine with Intel i$9$ CPU, Nvidia RTX $3090$ GPU, and $64$GB RAM memory. 
The code is written in Python $3.7$ and we use PyTorch $1.4$ on CUDA $10.1$ for model training. 
We repeat the experiment $5$ times and report the average results (for all experiments) and its standard deviation (for all experiment results except the Table~\ref{table:feature_injection_test} due to space limit).
 
 \noindent\textbf{Model configuration.}
% For \textit{``feature-label injection test''}, 
For fair comparision, the same linear-GNN is used for \our and baseline methods is used:
we use $3$-layer linear-GNN with shallow-subgraph sampler~\cite{zeng2020deep} for OGB-Arxiv and OGB-Products dataset, use $2$-layer linear-GNN with full-batch training for Cora and Pubmed dataset.
During training, label reuse tricks in~\cite{wang2021bag} are used that leverage the training set node label information for inference.
In terms of the linear GNN model we used in \oure and all other baselines, we train the linear-GNN using SGD with momentum with learning rate selected from $\{0.1, 1.0\}$, momentum as $0.9$, adaptive aggregation step size $\gamma=1$, and regularization as $10^{-6}$. Besides, we choose the regularization term $\lambda$ in \textsc{Influence+} and \textsc{Fisher+} to balance the performance before and after unlearning: when $\lambda$ is small, we are facing the gradient exploding issue where the gradient norm is an order of magnitude larger than the weight norm, such that the unlearned model cannot generate meaningful predictions. However, a larger $\lambda$ will hurt model's learning ability and results in a poor performance before unlearning.

\noindent\textbf{Details on dataset.~} We summarize the datasets that are used for experiments in Table~\ref{table:dataset_stat}.
\begin{table}[h]
\centering
\caption{Statistics of the datasets used in our experiments.} \label{table:dataset_stat}
\scalebox{0.99}{
\begin{tabular}{ l l l l l }
\hline\hline
             & \# Nodes  & \# Edges                                                                        & \# Features                                        & \# Classes \\ \hline\hline
OGB-Arxiv    & 169,343   & 1,166,243                                                                       & 128                                                & 40         \\ 
OGB-Products & 2,449,029 & 61,859,140                                                                      & 100                                                & 47         \\ 
Cora         & 2,708     & 10,556                                                                          & 1,433                                              & 7          \\ 
Pubmed       & 19,717    & 88,648                                                                          & 500                                                & 3          \\ 
Flickr       & 89,250    & 899,756                                                                         & 500                                                & 7          \\ 
Reddit       & 232,965   & 114,615,892 & 602 & 41         \\ \hline\hline
\end{tabular}}
% \vspace{-3mm}
\end{table}

% \noindent\textbf{Baseline implementation details.}~
% The code for \textsc{Influence+} and \textsc{Fisher+} are modified from \href{https://github.com/facebookresearch/certified-removal}{
% certified-removal}. Ordinary GNNs' implementations on OGB-Arxiv are modified from~\href{https://github.com/snap-stanford/ogb/tree/master/examples/nodeproppred/arxiv}{ogb-arxiv} and on OGB-Productsis are modified from~\href{https://github.com/snap-stanford/ogb/tree/master/examples/nodeproppred/products}{ogb-products}.

% \textcolor{red}{Add 1 more experiment on training + testing loss/ accuracy + that $\delta$}
% \input{tables/delete_data_reply_test_cora}

% \clearpage
% \clearpage
\section{Dependency issue in applying existing unlearning approaches} \label{section:dependency_issue_in_finite_sum}

Most unlearning approaches~\cite{wu2020deltagrad,guo2020certified,izzo2021approximate} are designed for the settings where the loss function can be decomposed over individual training samples. Directly generalizing the aforementioned general machine unlearning methods to graph structured data is infeasible due to the node dependency. In other word, one cannot directly unlearn a specific node $v_i$, but have to remove the effect of all its multi-hop neighbors in parallel if using these methods.

In the following, we use~\cite{guo2020certified} as an example to illustrate the key issue. The discussion also applied to other machine unlearning methods that require the loss function to be decomposed over individual training samples.
In the following, we first recall how the influence function is used to update the weight parameters in~\cite{guo2020certified}, then highlight why node dependency makes applying~\cite{guo2020certified} to graph-structured data challenging and introduce a solution to alleviate this issue.

\noindent\textbf{Influence function in~\cite{guo2020certified}.}
The influence function used in~\cite{guo2020certified} captures the change in model parameters due to removing a data point from the training set. Let $L(\mathbf{w})$ denote the finite-sum objective function computed on the full training set $\{\mathbf{x}_i\}_{i=1}^n$ with optimal solution
\begin{equation} \label{eq:w_star_optimal}
    \mathbf{w}_\star = \arg\min_\mathbf{w} L(\mathbf{w}), \text{where}~L(\mathbf{w}) = \sum_{i=1}^n \ell(\mathbf{w}^\top \mathbf{x}_i, y_i)
\end{equation}
and $L_{\setminus n}(\mathbf{w})$ denote the objective function without data point $(\mathbf{x}_n, y_n)$ with optimal solution
\begin{equation} \label{eq:w_setminus_n_optimal}
    \mathbf{w}_{\setminus n} = \arg\min_\mathbf{w} L_{\setminus_n}(\mathbf{w}), \text{where}~L_{\setminus_n}(\mathbf{w}) =  \sum_{i=1}^{n-1} \ell(\mathbf{w}^\top \mathbf{x}_i, y_i) = L(\mathbf{w}) - \ell(\mathbf{w}^\top \mathbf{x}_n, y_n).
\end{equation}
From $\mathbf{w}_{\setminus n} = \arg\min_\mathbf{w} L_{\setminus_n}(\mathbf{w})$ and the convexity of the objective function $L_{\setminus_n}$, we know that $\nabla L_{\setminus n}(\mathbf{w}_{\setminus n}) = \mathbf{0}$. Therefore, we have
\begin{equation} \label{eq:leave_one_out_formulation}
    \begin{aligned}
    0 &= \nabla L(\mathbf{w}_{\setminus n}) - \nabla \ell(\mathbf{w}_{\setminus n}^\top \mathbf{x}_n, y_n) \\
    &\underset{(a)}{\approx} \left[\nabla L(\mathbf{w}_\star) + \nabla^2 L(\mathbf{w}_\star) (\mathbf{w}_{\setminus n}-\mathbf{w}_\star)\right] - \left[ \nabla \ell(\mathbf{w}_\star^\top \mathbf{x}_n, y_n) + \nabla^2 \ell(\mathbf{w}_\star^\top \mathbf{x}_n, y_n) (\mathbf{w}_{\setminus n}-\mathbf{w}_\star) \right]\\
    &= \left[\nabla L(\mathbf{w}_\star) - \nabla \ell(\mathbf{w}_\star^\top \mathbf{x}_n, y_n) \right] + \left[\nabla^2 L(\mathbf{w}_\star) - \nabla^2 \ell(\mathbf{w}_\star^\top \mathbf{x}_n, y_n) \right] (\mathbf{w}_{\setminus n}-\mathbf{w}_\star) \\
    &\underset{(b)}{=} \left[ - \nabla \ell(\mathbf{w}_\star^\top \mathbf{x}_n, y_n) \right] + \left[\nabla^2 L(\mathbf{w}_\star) - \nabla^2 \ell(\mathbf{w}_\star^\top \mathbf{x}_n, y_n) \right] (\mathbf{w}_{\setminus n}-\mathbf{w}_\star),
    \end{aligned}
\end{equation}
where $(a)$ is the first-order Taylor expansion and $(b)$ due to $\nabla L(\mathbf{w}_\star) = \mathbf{0}$ for $\mathbf{w}_\star = \arg\min_\mathbf{w} L(\mathbf{w})$. Re-arranging the above equation we have
\begin{equation}
\begin{aligned}
\mathbf{w}_{\setminus n}\approx \mathbf{w}_\star  + \underbrace{\left[\nabla^2 L(\mathbf{w}_\star) - \nabla^2 \ell(\mathbf{w}_\star^\top \mathbf{x}_n, y_n) \right]  \nabla \ell(\mathbf{w}_\star^\top \mathbf{x}_n, y_n) }_{\text{influence function}},
\end{aligned}
\end{equation}
where the second term on the right hand side is the so called influence function.

\noindent\textbf{Challenges due to dependency in graph.}
Please notice that the objective function in Eq.~\ref{eq:w_star_optimal} and Eq.~\ref{eq:w_setminus_n_optimal} are finite-sum formulation. In the following, we will show that directly using the second-order method in~\cite{guo2020certified} is not allowed due to the node dependency in graph.
Before getting started, let me first introduce some notations: 
\begin{itemize}
    \item Let us denote the graph before node deletion as $\mathcal{G}$, where the graph structure is captured by adjacency matrix $\mathbf{A}\in\{0,1\}^{n\times n}$ and node feature matrix is $\mathbf{X}$. The row normalized propagation matrix us computed as $\mathbf{P} = \mathbf{D}^{-1} \mathbf{A}$.
    \item Let us denote the graph after node deletion as $\mathcal{G}_{\setminus n}$, where the graph structure is captured by adjacency matrix $\mathbf{A}_{\setminus n}\in\{0,1\}^{(n-1)\times (n-1)}$ and node feature matrix is $\mathbf{X}_{\setminus n} \in \mathbb{R}^{(n-1)\times d}$. The row normalized propagation matrix us computed as $\mathbf{P}_{\setminus n} = \mathbf{D}_{\setminus n}^{-1} \mathbf{A}_{\setminus n}$.
\end{itemize}

For simplicity, let us only consider $1$-hop SGC, which is already enough to illustrate why node dependency makes applying machine unlearning methods to graph structured data challenging.
In graph structured data, let $F(\mathbf{w})$ denote the objective function computed on the full training graph $\mathcal{G}$ with optimal solution
\begin{equation} 
    \mathbf{w}_\star = \arg\min_\mathbf{w} L(\mathbf{w}), \text{where}~L(\mathbf{w}) = \sum_{i=1}^n \ell(\mathbf{w}^\top [\mathbf{P}\mathbf{X}]_i, y_i)
\end{equation}
and $L_{\setminus n}(\mathbf{w})$ denote the objective function computed on graph $\mathcal{G}_{\setminus n}$ without node $n$ , with optimal solution
\begin{equation} 
    \begin{aligned}
    \mathbf{w}_{\setminus n} = \arg\min_\mathbf{w} L_{\setminus_n}(\mathbf{w}), \text{where}~L_{\setminus_n}(\mathbf{w}) &=  \sum_{i=1}^{n-1} \ell(\mathbf{w}^\top [\mathbf{P}_{\setminus n} \mathbf{X}]_i, y_i) \\
    &\underset{(a)}{\neq} L(\mathbf{w}) - \ell(\mathbf{w}^\top [\mathbf{P}\mathbf{X}]_n, y_n).
    \end{aligned}
\end{equation}
Due to the inequality of $(a)$,
we cannot directly use the second-order method in~\cite{guo2020certified} to approximate $\mathbf{w}_{\setminus n}$ from $\mathbf{w}_\star$.
Please notice that this equality is important in Eq.~\ref{eq:leave_one_out_formulation} before using first-order Taylor expansion.

\noindent\textbf{Get around this issue by deleting more nodes.}
One way to alleviate this issue is to update all the affected nodes $\mathcal{V}_\text{affect} = \{ v_n \} \cup \mathcal{N}(v_n)$ in parallel.
To see this, according to the definition of $\mathcal{V}_\text{affect}$, we know $[ \mathbf{P} \mathbf{X} ]_i = [\mathbf{P}_{\setminus n} \mathbf{X}_{\setminus n}]_i,~\forall v_i \in \mathcal{V}\setminus\mathcal{V}_\text{affect}$
because all the final-layer output of any node in $\mathcal{V}\setminus \mathcal{V}_\text{affect}$ are remaining the same after node deletion. 
Then, we can define the new objective function $L_{\setminus \mathcal{V}_\text{affect}}(\mathbf{w})$ on node set $\mathcal{V}\setminus \mathcal{V}_\text{affect}$
\begin{equation}
    \begin{aligned}
    L_{\setminus \mathcal{V}_\text{affect}}(\mathbf{w}) 
    &= \sum_{i\in\mathcal{V}\setminus \mathcal{V}_\text{affect}} \ell(\mathbf{w}^\top [\mathbf{P}_{\setminus n} \mathbf{X}_{\setminus n}]_i, y_i) \\
    &= \sum_{i\in\mathcal{V}\setminus \mathcal{V}_\text{affect}} \ell(\mathbf{w}^\top [\mathbf{P} \mathbf{X}]_i, y_i) \\
    &\underset{(a)}{=} L(\mathbf{w}) - \sum_{i\in\mathcal{V}_\text{affect}} \ell(\mathbf{w}^\top [\mathbf{P} \mathbf{X}]_i, y_i) + \sum_{i\in\mathcal{V}_\text{affect}\setminus\{n\}}\ell(\mathbf{w}^\top [\mathbf{P}_{\setminus n} \mathbf{X}_{\setminus n}]_i, y_i),
    \end{aligned}
\end{equation}
where the equality in $(a)$ is what we are looking for and is similar to the last term in Eq.~\ref{eq:w_setminus_n_optimal}.
To this end, let us define $\mathbf{w}_{\setminus \mathcal{V}_\text{affect}} = \arg\min_\mathbf{w} L_{\setminus \mathcal{V}_\text{affect}}(\mathbf{w})$, then we have
\begin{equation} \label{eq:dependency_influence_main}
    \begin{aligned}
    0 &= \nabla L(\mathbf{w}_{\setminus \mathcal{V}_\text{affect}}) - \sum_{i\in\mathcal{V}_\text{affect}} \nabla \ell(\mathbf{w}_{\setminus \mathcal{V}_\text{affect}}^\top [\mathbf{P}\mathbf{X}]_i, y_i) + \sum_{i\in\mathcal{V}_\text{affect}\setminus\{n\}}\nabla \ell(\mathbf{w}^\top [\mathbf{P}_{\setminus n} \mathbf{X}_{\setminus n}]_i, y_i) \\
    &\approx \left[\nabla L(\mathbf{w}_\star) - \sum_{i\in\mathcal{V}_\text{affect}} \nabla \ell(\mathbf{w}_\star^\top [\mathbf{P}\mathbf{X}]_i, y_i) + \sum_{i\in\mathcal{V}_\text{affect}\setminus\{n\}} \nabla \ell(\mathbf{w}_\star^\top [\mathbf{P}_{\setminus n} \mathbf{X}_{\setminus n}]_i, y_i)\right] \\
    &\quad + \left[\nabla^2 L(\mathbf{w}_\star) - \sum_{i\in\mathcal{V}_\text{affect}} \nabla^2 \ell(\mathbf{w}_\star^\top [\mathbf{P}\mathbf{X}]_i, y_i) + \sum_{i\in\mathcal{V}_\text{affect}\setminus\{n\}}\nabla^2 \ell(\mathbf{w}_\star^\top [\mathbf{P}_{\setminus n} \mathbf{X}_{\setminus n}]_i, y_i)\right] (\mathbf{w}_{\setminus \mathcal{V}_\text{affect}}-\mathbf{w}_\star) \\
    &\underset{(a)}{=} \underbrace{\left[ - \sum_{i\in\mathcal{V}_\text{affect}} \nabla \ell(\mathbf{w}_\star^\top [\mathbf{P}\mathbf{X}]_i, y_i) + \sum_{i\in\mathcal{V}_\text{affect}\setminus\{n\}} \nabla \ell(\mathbf{w}_\star^\top [\mathbf{P}_{\setminus n} \mathbf{X}_{\setminus n}]_i, y_i) \right]}_{\mathbf{v}} \\
    &\quad + \underbrace{\left[\nabla^2 L(\mathbf{w}_\star) - \sum_{i\in\mathcal{V}_\text{affect}} \nabla^2 \ell(\mathbf{w}_\star^\top [\mathbf{P}\mathbf{X}]_i, y_i) + \sum_{i\in\mathcal{V}_\text{affect}\setminus\{n\}}\nabla^2 \ell(\mathbf{w}_\star^\top [\mathbf{P}_{\setminus n} \mathbf{X}_{\setminus n}]_i, y_i) \right]}_{\mathbf{H}} (\mathbf{w}_{\setminus \mathcal{V}_\text{affect}}-\mathbf{w}_\star).
    \end{aligned}
\end{equation}
As a result, we can approximate $\mathbf{w}_{\setminus \mathcal{V}_\text{affect}}$ by
\begin{equation}
    \mathbf{w}_{\setminus \mathcal{V}_\text{affect}} = \mathbf{w}_\star + \mathbf{H}^{-1} \mathbf{v},
\end{equation}
where both Hessian $\mathbf{H}$ and gradient $\mathbf{v}$ are defined in Eq.~\ref{eq:dependency_influence_main}, which might induced massive computation cost as the number of affected nodes $|\mathcal{V}_\text{affect}|$ goes exponentially with respect to the number of layers.

% \clearpage

%%%%%%%%%%%%%%%%%%%%%%%%%%%%%%%%%%%%%%%%%%%%%%%%%%%%%%%%%
%%%%%%%%%%%%%%%%%%%%%%%%%%%%%%%%%%%%%%%%%%%%%%%%%%%%%%%%%
%%%%%%%%%%%%%%%%%%%%%%%%%%%%%%%%%%%%%%%%%%%%%%%%%%%%%%%%%

\section{Connections between differential privacy and machine unlearning}\label{section:differental_privacy_vs_unlearning}

% https://mukulrathi.com/privacy-preserving-machine-learning/deep-learning-differential-privacy/

The biggest difference between differential privacy and unlearning is whether the effect of data is removed from the model parameters. 
Please notice that the two methods can be used in parallel.
For example, when using approximate unlearning~\cite{guo2020certified,golatkar2020eternal,golatkar2021mixed}, since these methods could not guarantee a perfect data removal but just approximately removed, they propose to use differential privacy with their approximation unlearning method to further protect information leakage.
More specifically, 

\begin{itemize}
    \item \textit{Differential privacy} is designed to protect against the privacy leakage issue. In particular, they want to make a model trained on two different datasets behave similarly. The most widely accepted method is to control how much a model learned from each training example by adding random noise. 
    \item \textit{Machine unlearning}  is designed to remove the effect of a data point on the pre-trained. For example in  approximate unlearning~\cite{guo2020certified,golatkar2020eternal,wu2020deltagrad}, we want to make the re-training from scratch model $\mathbf{w}_u$ behaves similar to the model after unlearning $\mathbf{w}_p$, but whether data are perfectly removed are not guaranteed; in exact unlearning~\cite{chen2021graph}, we  want to make the re-training from scratch model $\mathbf{w}_u$ behaves similar to the model after unlearning $\mathbf{w}_p$, but guarantee the data are perfectly removed.
\end{itemize}

In the following, we will answer two questions related to differential privacy and unlearning:

\underline{\noindent\textit{Q1: Do we need to unlearning if a model is $(\epsilon,\delta)$-differential privacy?}}

If we looking for approximately remove the trace of private data, then a differential privacy method is enough (but may not be as efficient as approximate unlearning methods~\cite{guo2020certified}). However, if we are looking for perfectly remove the trace of private data, then differential privacy is not enough.
To see this, let us first recall the definition of $(\epsilon,\delta)$-differential privacy.

\begin{definition} [$(\epsilon,\delta)$-differential privacy] \label{def:differential_privacy}
Let $\epsilon>0$ be a positive real number, $\mathcal{D}, \mathcal{D}^\prime $ denote any two datasets that differ one a single element, $\mathcal{A}$ be a randomized algorithm that takes a dataset $\mathcal{D}, \mathcal{D}^\prime$ as input, $\mathcal{S}$ denote any subset of the image of $\mathcal{A}$. Then, we say $\mathcal{A}$ is $(\epsilon,\delta)$-differential privacy if $$P[\mathcal{A}(\mathcal{D}) \in \mathcal{S}] \leq \exp(\epsilon) \cdot P[\mathcal{A}(\mathcal{D}^\prime) \in \mathcal{S}] + \delta.$$
\end{definition}

To generalize the differential privacy definition to machine unlearning, let us think of $\mathcal{D}$ be the original dataset and $\mathcal{D}^\prime=\mathcal{D}\setminus \{(\mathbf{x}_i, y_i)\}$ be the remaining dataset after delete data $(\mathbf{x}_i, y_i)$.
Then, we can think of $\mathcal{A}$ as a composition of learning and unlearning algorithm:
\begin{equation}
    \mathcal{A}(\mathcal{D}) = \text{Unlearn}\Big(\text{Learn}(\mathcal{D}), (\mathbf{x}_i, y_i) \Big), \mathcal{A}(\mathcal{D}^\prime) = \text{Unlearn}\Big(\text{Learn}(\mathcal{D}^\prime), (\mathbf{x}_i, y_i) \Big)
\end{equation}
More specifically, $\mathcal{A}(\mathcal{D})$ can be think of as first training a model on $\mathcal{D}$ then unlearn to produce the unlearned solution $\mathbf{w}_p$;
$\mathcal{A}(\mathcal{D}^\prime)$ can be think of as re-training from scratch on $\mathcal{D}^\prime$ with $\mathbf{w}_u = \mathcal{A}(\mathcal{D}^\prime)$ since $\mathcal{D}^\prime$ does not contain $(\mathbf{x}_i, y_i)$.
In other word, differential privacy algorithm can also guarantee the behavior of the the unlearning solution $\mathbf{w}_p$ similar to re-training from scratch solution $\mathbf{w}_u$, which leads to our approximate unlearning goal.
However, $\mathbf{w}_p \approx \mathbf{w}_u$ not necessarily means $\mathbf{w}_p$ perfectly unlearn all the information related to the deleted data $(\mathbf{x}_i, y_i)$. Therefore, we cannot say a model is perfectly unlearned by using differential privacy method.

\underline{\noindent\textit{Q2: Can we have a differential privacy-like bounds for \oure?}}
A reader familiar with differential privacy might expect a differential privacy-like bound similar to Definition~\ref{def:differential_privacy}.  To understand why it is infeasible, let us first recall that $\epsilon$ term is the private budget that is related to the random noise distribution.
Since our method is exact unlearning which can be shown from the algorithm level (i.e., "the support of the unlearned weight does not include the deleted node feature" could be guaranteed by the algorithm design), we don't need random noise and therefore we cannot show this kind of inequality.
The reason why the approximate unlearning method~\cite{guo2020certified,chien2022certified} has such a similar guarantee to differential privacy is because they are approximate unlearning method that use differential privacy type of random noise to unlearn.

%%%%%%%%%%%%%%%%%%%%%%%%%%%%%%%%%%%%%%%%%%%%%%%%%%%%%%%%%
%%%%%%%%%%%%%%%%%%%%%%%%%%%%%%%%%%%%%%%%%%%%%%%%%%%%%%%%%
%%%%%%%%%%%%%%%%%%%%%%%%%%%%%%%%%%%%%%%%%%%%%%%%%%%%%%%%%

% \clearpage

\section{Why checking the closeness to re-trained solution along is not enough for unlearning?} \label{section:proof_unlearn_without_change}

Most approximate unlearning algorithms aim to generate the approximate unlearned model that is close to an exactly retrained model~\cite{wu2020deltagrad,aldaghri2021coded,izzo2021approximate}. However, 
as pointed out by~\cite{thudi2021necessity,guo2020certified}, one cannot infer \textit{``whether the data have been deleted''} solely from \textit{``the closeness of the approximately unlearned and exactly retrained model''}. In fact, 
\cite{thudi2021necessity} empirically shows that one can even unlearn the data without modifying the parameters, which highlights the importance of showing the data removal guarantee from the algorithmic-level.
In Theorem~\ref{thm:unlearn_without_change} below, we provide theoretical justification for the empirical observation in~\cite{thudi2021necessity} under the binary classification setting and the proof  is deferred to Appendix~\ref{section:proof_unlearn_without_change}. 
\begin{theorem} \label{thm:unlearn_without_change}
Consider a general binary classification problem using logistic regression 
$$\smash{\min_{\mathbf{w} \in \mathbb{R}^d}~f_{\text{LR}}(\mathbf{w}) = \frac{\lambda}{2} \| \mathbf{w} \|_2^2 + \sum_{i=1}^N \log\big( 1 + \exp(-y_i \mathbf{w}^\top \mathbf{x}_i) \big)}.$$
Upon receiving any request to unlearn $\mathbf{x}_i$, if it is misclassified by the optimal weight $\mathbf{w}_\star = \arg\min_\mathbf{w} f_{\text{LR}}(\mathbf{w})$, i.e., $y_i \mathbf{w}_\star^\top \mathbf{x}_i < 0$, we can unlearn without modifying the optimal weight $\mathbf{w}_\star$ as
the optimally conditions are still satisfied.
% it does not carry any information on $\mathbf{x}_i$.
\end{theorem}
An immediate implication of above theorem is that one could  exactly unlearn a misclassified data point even without  changing the model parameters.
% the solution before and after exactly unlearning a misclassified data point could be equivalent.
However, approximate unlearning methods~\cite{guo2020certified,golatkar2020eternal} cannot realize this from their algorithmic-level, which will output an estimated solution that is not only different from the exact unlearning solution but also could not fully unlearn the sensitive information. Although these methods borrow ideas from differential privacy (unlearning is different from the differential privacy, please refer to Appendix~\ref{section:differental_privacy_vs_unlearning} for details) to add a noise to unlearned model to avoid information leakage, this could also potentially deteriorate the accuracy of the unlearned model. 
% In fact, due to the nature of the approximate unlearning approaches, these approximate unlearning approaches cannot guarantee the perfect removal of all information related to the deleted data.
% Intuitively, such observation makes sense because the output of approximate unlearning is non-equivalent to the result of exact unlearning, 
Such observations highlights the importance of an algorithmic-level data removal guarantee over approximate unlearning.

The proof follows from standard optimality conditions. Let us consider a binary classification problem with $N$ training samples using regularized logistic regression with the following empirical risk:
\begin{equation}~\label{eq:primal_problem}
    \min_{\mathbf{w} \in \mathbb{R}^d}~f(\mathbf{w}) = \frac{\lambda}{2} \| \mathbf{w} \|_2^2 + \sum_{i=1}^N \log\Big( 1 + \exp(-y_i \mathbf{w}^\top \mathbf{x}_i) \Big).
\end{equation}

By utilizing the following inequality
\begin{equation}~\label{eq:primal_dual_inequality}
     \log(1+\exp(-z)) \geq \big( - \alpha \log \alpha - (1-\alpha) \log(1-\alpha) \big) - \alpha z,~
     \forall \alpha\in[0,1],
\end{equation}
we can lower bound the objective in Eq.~\ref{eq:primal_problem} by
\begin{equation}
    \begin{aligned}
    f(\mathbf{w}) 
    & = \frac{\lambda}{2} \| \mathbf{w} \|_2^2 + \sum_{i=1}^N \log\Big( 1 + \exp(-y_i \mathbf{w}^\top \mathbf{x}_i) \Big) \\
    & \geq \frac{\lambda}{2} \| \mathbf{w} \|_2^2 + \sum_{i=1}^N \left(  - \alpha_i \log \alpha_i - (1-\alpha_i) \log(1-\alpha_i)  - \alpha_i y_i\mathbf{w}^\top \mathbf{x}_i \right) \\
    & = L(\mathbf{w}, \bm{\alpha}),
    \end{aligned}
\end{equation}
where the right hand size of inequality can be think of as a function of parameters $\mathbf{w}$ and $\bm{\alpha}$.

\noindent\textbf{Optimal solution lies in linear span of input features.}~
From the stationarity condition of the KKT conditions~\cite{boyd2004convex}, we know that the gradient of $L(\mathbf{w}, \bm{\alpha})$ with respect to $\mathbf{w}$ at the optimal point $\mathbf{w}^\star$ equals to zero, i.e.,
\begin{equation}
    \frac{\partial L(\mathbf{w}_\star, \bm{\alpha})}{\partial \mathbf{w}} = \lambda \mathbf{w} - \sum_{i=1}^N \alpha_i y_i \mathbf{x}_i = 0~\Rightarrow \mathbf{w}_\star = \frac{1}{\lambda} \sum_{i=1}^N \alpha_i y_i \mathbf{x}_i,
\end{equation}
which implies that the optimal solution, regardless of the initialization, is a linear combination of all training samples $\{ \mathbf{x}_1,\ldots,\mathbf{x}_N\} $.

\noindent\textbf{Unlearning without modifying the parameters.}~
By using the complementary slackness conditions of the KKT conditions, we know that for any $i\in\{1,\ldots,N\}$, we have
\begin{equation}\label{eq:complementary_slackness}
    \alpha_i = 0 ~\vee~ y_i \mathbf{w}_\star^\top \mathbf{x}_i = -\log \alpha_i - \left(-1+\frac{1}{\alpha_i} \right) \log(1-\alpha_i) > 0.
\end{equation}
From Eq.~\ref{eq:complementary_slackness}, we know that $\alpha_i \neq 0$ if and only if $\mathbf{x}_i$ can be correctly classified by optimal weight $\mathbf{w}_\star$, i.e., $y_i \mathbf{w}_\star^\top \mathbf{x}_i > 0$; otherwise, we have $\alpha_i=0$ if $\mathbf{x}_i$ cannot be correctly classified by optimal weight $\mathbf{w}_\star$, which can be directly unlearned without requiring to modify the weight parameter $\mathbf{w}_\star$.
When removing data $\mathbf{x}_i$ with dual variable as $\alpha_i=0$, the KKT optimality condition still hold, similar to the main idea of SVM unlearning~\cite{cauwenberghs2000incremental}, therefore further updating the weight parameters is not required.

% \clearpage
\section{Proof of Proposition~\ref{proposition:projection_step}}
\label{section:proof_alternative_orthogonal_projection}

To find the coefficients of the orthogonal projection, let us write down the following $r = |\mathcal{V}_\text{remain}|$ simultaneous conditions on linear equations: 
\begin{equation}\label{eq:simultaneous_orthogonal_condition}
    \left\langle \mathbf{x}_i, \mathbf{w} - \Pi_\mathcal{U}(\mathbf{w}) \right\rangle=0,~
    \forall i \in \mathcal{V}_\text{remain}.
    % = \langle \mathbf{x}_i, \mathbf{w} - \sum_{j=r+1}^N \alpha_j \mathbf{x}_j \rangle
\end{equation}

For notation simplicity, let $\mathbf{X}_\text{remain} = [\mathbf{x}_{r+1}, \ldots, \mathbf{x}_N] \in \mathbb{R}^{r\times d}$ denote the remaining node features and $\bm{\alpha} = [\alpha_{r+1},\ldots, \alpha_N] \in\mathbb{R}^{R}$ denote the vectorized coordinates, then the Eq.~\ref{eq:simultaneous_orthogonal_condition} can be formulated as $\mathbf{X}_\text{remain} (\mathbf{w} - \mathbf{X}_\text{remain}^\top \bm{\alpha} ) = \mathbf{0}$. 
As a result, we have 
\begin{equation}\label{eq:original_orthogonal_projection}
    \mathbf{X}_\text{remain} (\mathbf{w} - \mathbf{X}_\text{remain}^\top \bm{\alpha} ) = \mathbf{0}
    \Rightarrow    
    \bm{\alpha} = (\mathbf{X}_\text{remain} \mathbf{X}_\text{remain}^\top )^{\dagger} \mathbf{X}_\text{remain} \mathbf{w}.
\end{equation}
However, notice that $\mathbf{X}_\text{remain} \mathbf{X}_\text{remain}^\top$ is an $r \times r$ matrix and the computation of its inverse requires $\mathcal{O}(r^3)$ computation cost, which is computational prohibitive.
As an alternative, we reconsider Eq.~\ref{eq:original_orthogonal_projection} from a different perspective by viewing it as the limit of the Ridge estimator with the Ridge parameter going to zero, i.e.,
\begin{equation}
    (\mathbf{X}_\text{remain} \mathbf{X}_\text{remain}^\top )^{\dagger} \mathbf{X}_\text{remain} \mathbf{w}  = \lim_{\epsilon\rightarrow 0} (\epsilon\mathbf{I}_N + \mathbf{X}_\text{remain} \mathbf{X}_\text{remain}^\top )^{\dagger} \mathbf{X}_\text{remain} \mathbf{w}.
\end{equation}

Then, by using the Woodbury identity~\cite{golub2013matrix} we have
\begin{equation}
    \begin{aligned}
    & (\epsilon\mathbf{I}_N + \mathbf{X}_\text{remain} \mathbf{X}_\text{remain}^\top )^{\dagger} \mathbf{X}_\text{remain} \mathbf{w} \\
    &= \Big( \frac{1}{\epsilon} \mathbf{I}_N - \mathbf{X}_\text{remain} (\mathbf{I}_N + \epsilon \mathbf{X}_\text{remain}^\top \mathbf{X}_\text{remain})^{\dagger} \mathbf{X}_\text{remain}^\top \Big) \mathbf{X}_\text{remain} \mathbf{w} \\
    &= \frac{1}{\epsilon} \mathbf{X}_\text{remain} \mathbf{w} - \mathbf{X}_\text{remain} (\mathbf{I}_N + \epsilon \mathbf{X}_\text{remain}^\top \mathbf{X}_\text{remain})^{\dagger} \mathbf{X}_\text{remain}^\top \mathbf{X}_\text{remain} \mathbf{w} \\
    &= \frac{1}{\epsilon} \mathbf{X}_\text{remain} \mathbf{w} - \frac{1}{\epsilon} \mathbf{X}_\text{remain} (\mathbf{I}_N + \epsilon \mathbf{X}_\text{remain}^\top \mathbf{X}_\text{remain})^{\dagger} (\epsilon \mathbf{X}_\text{remain}^\top \mathbf{X}_\text{remain}) \mathbf{w} \\
    &= \frac{1}{\epsilon} \mathbf{X}_\text{remain} \mathbf{w} - \frac{1}{\epsilon}\mathbf{X}_\text{remain} (\mathbf{I}_N + \epsilon \mathbf{X}_\text{remain}^\top \mathbf{X}_\text{remain})^{\dagger} (\mathbf{I}_N + \epsilon \mathbf{X}_\text{remain}^\top \mathbf{X}_\text{remain} - \mathbf{I}_N ) \mathbf{w} \\
    &= \frac{1}{\epsilon} \mathbf{X}_\text{remain} \mathbf{w} - \frac{1}{\epsilon}\mathbf{X}_\text{remain} \mathbf{w} + \frac{1}{\epsilon} \mathbf{X}_\text{remain} (\mathbf{I}_N + \epsilon \mathbf{X}_\text{remain}^\top \mathbf{X}_\text{remain})^{\dagger} \mathbf{w} \\
    &= \mathbf{X}_\text{remain} (\epsilon\mathbf{I}_d + \mathbf{X}_\text{remain}^\top \mathbf{X}_\text{remain})^{\dagger} \mathbf{w}.
    \end{aligned}
\end{equation}
By taking the limit on both side, we have 
\begin{equation} \label{eq:alternative_orthogonal_projection}
    \bm{\alpha} = \mathbf{X}_\text{remain} (\mathbf{X}_\text{remain}^\top \mathbf{X}_\text{remain} )^{\dagger}  \mathbf{w},
\end{equation}
where $\mathbf{X}_\text{remain}^\top \mathbf{X}_\text{remain}$ is an $d\times d$ matrix and its inverse requires $\mathcal{O}(d^3)$, which is much cheaper.

% \clearpage

\clearpage
\section{Proof of Theorem~\ref{theorem:ell2_norm_of_weight_projections}}
\label{section:proof_of_ell2_norm_of_weight_projections}

\begin{figure}[h]
    \centering
    \includegraphics[width=0.95\textwidth]{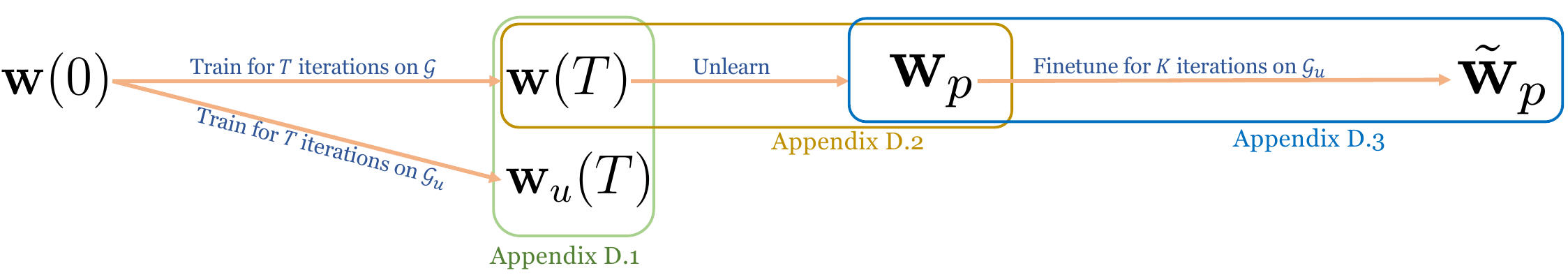}
    \vspace{-3mm}
    \caption{An illustration on the proof strategy of Theorem~\ref{theorem:ell2_norm_of_weight_projections}, where $\mathcal{G}$ stands for the graph used before node deletion and $\mathcal{G}_u$ stands for the graph after the node deletion.}
    \label{fig:main_idea_on_proof}
\end{figure}

Let us define $\mathbf{w}(t),~\mathbf{w}_u(t)$ as the weight parameters obtained by using full-batch GD training from scratch on $\mathcal{V}_\text{train},~\mathcal{V}_\text{remain}$ for $t$ epochs with the same initialization $\mathbf{w}(0) = \mathbf{w}_u(0)$, and $\mathbf{w}_p(t)$ denote the weight parameters obtained by applying \our on $\mathbf{w}(t)$.
To help reader better understand the proof strategy used in this section, we provide an overview on our proof strategy of Theorem~\ref{theorem:ell2_norm_of_weight_projections}.
As shown in Figure~\ref{fig:main_idea_on_proof}, we derive the upper bound of $\| \mathbf{w}_u(T) - \mathbf{w}_p \|_2$ by first expanding the formula into two terms
\begin{equation}
    \| \mathbf{w}_u(T) - \mathbf{w}_p \|_2 \leq \| \mathbf{w}_u(T) - \mathbf{w}(T) \|_2 + \| \mathbf{w}(T) - \mathbf{w}_p \|_2.
\end{equation}
Then, we derive the upper bound of the first term in Appendix~\ref{section:W_u_diff_W_at_T} and the second term in Appendix~\ref{section:W_p_diff_W_at_T}, and obtain the upper bound of $\| \mathbf{w}_u(T) - \mathbf{w}_p \|_2 $.
Then, by following the standard convergence analysis of smooth convex function, we obtain the upper bound training error $F^u(\tilde{\mathbf{w}}_p) - \min_\mathbf{w} F^u(\mathbf{w})$ 
% in terms of $\| \mathbf{w}_\star - \mathbf{w}_p \|_2,~\mathbf{w}_\star = \arg\min_\mathbf{w} F^u(\mathbf{w}) $ and number of fune-tuning steps $K$ 
in Appendix~\ref{section:fine-tune-rate}.

\subsection{Upper bound on $\| \mathbf{w}_u(T) - \mathbf{w}(T)\|_2$} \label{section:W_u_diff_W_at_T}

Let first recall that the gradient of Eq.~\ref{eq:ovr_logistic_regression} and Eq.~\ref{eq:closeness_after_delete_objective} are computed as
\begin{equation} \label{eq:full_gradient_before_after_unlearning}
    \begin{aligned}
    \nabla F(\mathbf{w}) &= \frac{1}{|\mathcal{V}_\text{train}|}\sum_{i\in\mathcal{V}_\text{train}} \nabla f_i(\mathbf{w}),~
    \nabla f_i(\mathbf{w}) = -y_i \sigma(- y_i \mathbf{w}^\top \mathbf{h}_i) \mathbf{h}_i + \lambda \mathbf{w},\\
    \nabla F^u(\mathbf{w}) &= \frac{1}{|\mathcal{V}_\text{remain}|}\sum_{i\in\mathcal{V}_\text{remain}} \nabla f^u_i(\mathbf{w}),~
    \nabla f^u_i(\mathbf{w}) = -y_i \sigma(- y_i \mathbf{w}^\top \mathbf{h}^u_i) \mathbf{h}_i^u + \lambda \mathbf{w},
    \end{aligned}
\end{equation}
where $\sigma(x) = \frac{1}{1+\exp(-x)}$ is the Sigmoid function.
From Eq.~\ref{eq:full_gradient_before_after_unlearning}, we know that $f_i(\mathbf{w}), f_i^u(\mathbf{w})$ is $(\lambda+P_s^2 B_x^2)$-smoothness, which is shown as follows
\begin{equation} \label{eq:smoothness}
    \begin{aligned}
    \| \nabla f_i(\mathbf{w}_1) - \nabla f_i(\mathbf{w}_2) \|_2
    &= \left\| -y_i \sigma(- y_i \mathbf{w}_1^\top \mathbf{h}_i) \mathbf{h}_i + y_i \sigma(- y_i \mathbf{w}_2^\top \mathbf{h}_i) \mathbf{h}_i + \lambda(\mathbf{w}_1 - \mathbf{w}_2)\right\|_2 \\
    &\leq \| y_i \mathbf{h}_i \|_2 \cdot |\sigma(- y_i \mathbf{w}_1^\top \mathbf{h}_i) + \sigma(- y_i \mathbf{w}_2^\top \mathbf{h}_i)| + \lambda \| \mathbf{w}_1 - \mathbf{w}_2 \|_2 \\
    &\leq \| y_i \mathbf{h}_i \|_2 \cdot \|y_i \mathbf{w}_1^\top \mathbf{h}_i - y_i \mathbf{w}_2^\top \mathbf{h}_i \|_2 + \lambda \| \mathbf{w}_1 - \mathbf{w}_2 \|_2\\
    &\leq \| y_i \mathbf{h}_i \|_2^2 \cdot \| \mathbf{w}_1 - \mathbf{w}_2\|_2 + \lambda \| \mathbf{w}_1 - \mathbf{w}_2 \|_2\\
    &\leq \left( \max_i \left\| [\mathbf{P}^L \mathbf{X}]_i \right\|_2^2 \right) \| \mathbf{w}_1 - \mathbf{w}_2\|_2 + \lambda \| \mathbf{w}_1 - \mathbf{w}_2 \|_2\\
    &\leq (\lambda+P_s^2 B_x^2) \| \mathbf{w}_1 - \mathbf{w}_2\|_2, \\
    \| \nabla f_i^u(\mathbf{w}_1) - \nabla f_i^u(\mathbf{w}_2) \|_2 &\leq \left( \max_i \left\| [\mathbf{P}_u^L \mathbf{X}]_i \right\|_2^2 \right) \| \mathbf{w}_1 - \mathbf{w}_2\|_2 + \lambda \| \mathbf{w}_1 - \mathbf{w}_2 \|_2 \\
    &\leq (\lambda + P_s^2 B_x^2) \| \mathbf{w}_1 - \mathbf{w}_2\|_2. 
    \end{aligned}
\end{equation}

We can upper bound the difference of the gradient computed before and after data deletion by
\begin{equation}
    \begin{aligned}
    &\mathbb{E} \left[ \| \nabla F(\mathbf{w}) -  \nabla F^u(\mathbf{w}) \|_2 \right]\\
    &\quad= \mathbb{E} \left[\left\|  \frac{1}{|\mathcal{V}_\text{train}|}\sum_{i\in\mathcal{V}_\text{train}} \nabla f_i(\mathbf{w}) - \frac{1}{|\mathcal{V}_\text{remain}|}\sum_{i\in\mathcal{V}_\text{remain}} \nabla f_i^u(\mathbf{w}_u) \right\|_2 \right]\\
    &\quad\underset{(a)}{\leq} \left\|  \frac{1}{|\mathcal{V}_\text{train}|}\sum_{i\in\mathcal{V}_\text{train}} \nabla f_i(\mathbf{w}) - \frac{1}{|\mathcal{V}_\text{train}|}\sum_{i\in\mathcal{V}_\text{train}} \nabla f_i^u(\mathbf{w}) \right\|_2 \\
    &\quad\quad + \mathbb{E}\left[ \left\|  \frac{1}{|\mathcal{V}_\text{train}|}\sum_{i\in\mathcal{V}_\text{train}} \nabla f_i^u(\mathbf{w}) - \frac{1}{|\mathcal{V}_\text{remain}|}\sum_{i\in\mathcal{V}_\text{remain}} \nabla f_i^u(\mathbf{w}_u)  \right\|_2 \right],
    \end{aligned}
\end{equation}
% \begin{equation}
%     \begin{aligned}
%     &\mathbb{E} \left[ \| \nabla F(\mathbf{w}) -  \nabla F^u(\mathbf{w}) \|_2 \right]\\
%     &\quad= \mathbb{E} \left[\left\|  \frac{1}{|\mathcal{V}_\text{train}|}\sum_{i\in\mathcal{V}_\text{train}} y_i \sigma(- y_i \mathbf{w}^\top \mathbf{h}_i) \mathbf{h}_i - \frac{1}{|\mathcal{V}_\text{remain}|}\sum_{i\in\mathcal{V}_\text{remain}} y_i \sigma(- y_i \mathbf{w}^\top \mathbf{h}^u_i) \mathbf{h}_i^u \right\|_2 \right]\\
%     &\quad\underset{(a)}{\leq} \left\|  \frac{1}{|\mathcal{V}_\text{train}|}\sum_{i\in\mathcal{V}_\text{train}} y_i \sigma(- y_i \mathbf{w}^\top \mathbf{h}_i) \mathbf{h}_i - \frac{1}{|\mathcal{V}_\text{train}|}\sum_{i\in\mathcal{V}_\text{train}} y_i \sigma(- y_i \mathbf{w}^\top \mathbf{h}^u_i) \mathbf{h}_i^u \right\|_2 \\
%     &\quad\quad + \mathbb{E}\left[ \left\|  \frac{1}{|\mathcal{V}_\text{train}|}\sum_{i\in\mathcal{V}_\text{train}} y_i \sigma(- y_i \mathbf{w}^\top \mathbf{h}^u_i) \mathbf{h}_i^u - \frac{1}{|\mathcal{V}_\text{remain}|}\sum_{i\in\mathcal{V}_\text{remain}} y_i \sigma(- y_i \mathbf{w}^\top \mathbf{h}^u_i) \mathbf{h}_i^u \right\|_2 \right]\\
%     \end{aligned}
% \end{equation}
where $(a)$ is achieved by adding and subtracting the same term and $\| \mathbf{a} + \mathbf{b} \|_2 \leq \| \mathbf{a} \|_2 + \| \mathbf{b}\|_2$ and the expectation on the randomness the deleted node selection.

The first term on the right hand side of inequality $(a)$ can be further upper bounded by
\begin{equation}
    \begin{aligned}
    & \left\|  \frac{1}{|\mathcal{V}_\text{train}|}\sum_{i\in\mathcal{V}_\text{train}} y_i \sigma(- y_i \mathbf{w}^\top \mathbf{h}_i) \mathbf{h}_i - \frac{1}{|\mathcal{V}_\text{train}|}\sum_{i\in\mathcal{V}_\text{train}} y_i \sigma(- y_i \mathbf{w}^\top \mathbf{h}^u_i) \mathbf{h}_i^u \right\|_2 \\
    &\quad \underset{(a)}{\leq} \frac{1}{|\mathcal{V}_\text{train}|}\sum_{i\in\mathcal{V}_\text{train}} \left\|   y_i \sigma(- y_i \mathbf{w}^\top \mathbf{h}_i) \mathbf{h}_i - y_i \sigma(- y_i \mathbf{w}^\top \mathbf{h}_i) \mathbf{h}_i^u \right\|_2 + \\
    &\quad\quad + \frac{1}{|\mathcal{V}_\text{train}|}\sum_{i\in\mathcal{V}_\text{train}} \left\|   y_i \sigma(- y_i \mathbf{w}^\top \mathbf{h}_i) \mathbf{h}_i^u -  y_i \sigma(- y_i \mathbf{w}^\top \mathbf{h}^u_i) \mathbf{h}_i^u \right\|_2  \\
    &\quad \leq \frac{1}{|\mathcal{V}_\text{train}|}\sum_{i\in\mathcal{V}_\text{train}} |   y_i \sigma(- y_i \mathbf{w}^\top \mathbf{h}_i) | \left\| \mathbf{h}_i - \mathbf{h}_i^u \right\|_2 + \\
    &\quad\quad + \frac{1}{|\mathcal{V}_\text{train}|}\sum_{i\in\mathcal{V}_\text{train}} \left\|   y_i  \mathbf{h}_i^u \right\|_2 | \sigma(- y_i \mathbf{w}^\top \mathbf{h}_i) -  \sigma(- y_i \mathbf{w}^\top \mathbf{h}^u_i)  |  \\
    &\quad \underset{(b)}{\leq} \frac{1}{|\mathcal{V}_\text{train}|}\sum_{i\in\mathcal{V}_\text{train}} \left( \| \mathbf{h}_i - \mathbf{h}_i^u \|_2 + \|   \mathbf{h}_i^u \|_2 \| \mathbf{w} \|_2 \| \mathbf{h}_i - \mathbf{h}_i^u\|_2 \right) \\
    &\quad = \frac{1}{|\mathcal{V}_\text{train}|}\sum_{i\in\mathcal{V}_\text{train}} \left(1+\|   \mathbf{h}_i^u \|_2 \| \mathbf{w} \|_2 \right) \| \mathbf{h}_i - \mathbf{h}_i^u\|_2,
    \end{aligned}
\end{equation}
where $(a)$ is due to $\| \mathbf{a} + \mathbf{b} \|_2 \leq \| \mathbf{a} \|_2 + \| \mathbf{b}\|_2$ and $(b)$ is due to $| y_i | = 1$ and $| \sigma(x) - \sigma(y) | \leq | x - y |$.
By using the definition of $\mathbf{h}_i^u$ and $\mathbf{h}_i$ and Assumption~\ref{assumption:upper_bound_on_node_feats}, we have for any $i$
\begin{equation}
    \begin{aligned}
    \| \mathbf{h}_i - \mathbf{h}_i^u \|_2 
    &= \max_j \| [ \mathbf{P}^L \mathbf{X} - \mathbf{P}^L_u \mathbf{X} ]_j \|_2 \\
    &\leq \max_j \| \mathbf{x}_j \|_2 \cdot \max_j \left\| [ \mathbf{P}^L - \mathbf{P}^L_u ]_j \right\|_2 \\
    & \leq B_x P_d, \\
    \| \mathbf{h}_i^u \|_2 &= \max_j \| [ \mathbf{P}^L_u \mathbf{X} ]_j \|_2 \\
    &\leq  \max_j \| \mathbf{x}_j \|_2  \cdot \max_j \left\| [ \mathbf{P}^L_u ]_j \right\|_2 \\
    &\leq B_x P_s.
    \end{aligned}
\end{equation}
Besides, the upper bound of the second term on the right hand side of inequality $(a)$ is from Assumption~\ref{assumption:expectation_random_delete_nodes}.
By plugging the results back to the right hand side of inequality $(a)$, we have
\begin{equation} \label{eq:grad_diff_before_after_unlearn}
    \| \nabla F(\mathbf{w}) -  \nabla F^u(\mathbf{w}) \|_2 
    \leq (1+B_x B_w P_s) B_x P_d + G.
    % G + 2BP
\end{equation}

 Therefore, according to the gradinet update rule in gradient descent, we can bound 
the change of model parameters $\mathbf{w}_u(t)$ and $\mathbf{w}(t)$ by
\begin{equation}
    \begin{aligned}
    &\| \mathbf{w}_u(t+1) - \mathbf{w}(t+1) \|_2 \\
    &\quad = \left\| \mathbf{w}_u(t) - \eta \nabla F^u(\mathbf{w}_u(t)) -  \mathbf{w}(t) + \eta \nabla F(\mathbf{w}(t))  \right\|_2 \\
    % &\quad = \left\| \left( \mathbf{w}_u^{t} - \mathbf{w}^{t}\right) - \eta \left( \nabla F^u(\mathbf{w}_u(t))  - \nabla F(\mathbf{w}(t)) \right) \right\|_2 \\
    &\quad\leq \| \mathbf{w}_u^{t} - \mathbf{w}^{t} \|_2 + \eta \| \nabla F^u(\mathbf{w}_u(t))  - \nabla F(\mathbf{w}_u(t)) + \nabla F(\mathbf{w}_u(t)) - \nabla F(\mathbf{w}(t)) \|_2 \\
    &\quad \leq (1+\eta (\lambda+B_x^2B_s^2)) \| \mathbf{w}_u^{t} - \mathbf{w}^{t} \|_2 + \eta \| \nabla F^u(\mathbf{w}_u(t))  - \nabla F(\mathbf{w}_u(t))  \|_2 \\
    &\quad\underset{(a)}{\leq}  (1+\eta (\lambda+B_x^2B_s^2)) \| \mathbf{w}_u^{t} - \mathbf{w}^{t} \|_2 + \eta \Big( (1+B_x B_w P_s) B_x P_d + G  \Big), \\
    \end{aligned}
\end{equation}
where $(a)$ is due to Eq.~\ref{eq:smoothness} and Eq.~\ref{eq:grad_diff_before_after_unlearn}.
Then after $T$ iterations, we can bound the different between two parameters as
\begin{equation}
    \| \mathbf{w}_u(T) - \mathbf{w}(T) \|_2 \leq \eta \Big( (1+B_x B_w P_s) B_x P_d + G  \Big) \sum_{t=1}^T (1+\eta (\lambda+B_x^2B_s^2))^{t-1}.
\end{equation}

\subsection{Upper bound on $\| \mathbf{w}_p - \mathbf{w} (T)\|_2$ for \our} \label{section:W_p_diff_W_at_T}

From Proposition~\ref{proposition:projection_step} we know that there exist a set of coordinates $\{\beta_i~|~i\in\mathcal{V}\}$ such that $\mathbf{w}(T) = \sum_{j\in\mathcal{V}} \beta_j \mathbf{x}_j$ holds.
Besides, by using Eq.~\ref{eq:full_gradient_before_after_unlearning}, we have 
\begin{equation}
    \begin{aligned}
    \nabla F(\mathbf{w}(t)) &= \frac{1}{|\mathcal{V}_\text{train}|}\sum_{i\in\mathcal{V}_\text{train}} - y_i \sigma(- y_i \mathbf{w}^\top(t) \mathbf{h}_i) \mathbf{h}_i + \lambda \mathbf{w}(t)\\
    &= \frac{1}{|\mathcal{V}_\text{train}|}\sum_{i\in\mathcal{V}_\text{train}} - y_i \sigma(- y_i \mathbf{w}^\top(t) \mathbf{h}_i) [\mathbf{P}^L \mathbf{X}]_i + \lambda \mathbf{w}(t)\\
    &= \frac{1}{|\mathcal{V}_\text{train}|}\sum_{i\in\mathcal{V}_\text{train}} - y_i \sigma(- y_i \mathbf{w}^\top(t) \mathbf{h}_i) \left( \sum_{j\in\mathcal{V}} [\mathbf{P}^L]_{ij}\mathbf{x}_j \right) + \lambda \mathbf{w}(t) \\
    &= \sum_{j\in\mathcal{V}} \left( \frac{1}{|\mathcal{V}_\text{train}|}\sum_{i\in\mathcal{V}_\text{train}} - y_i \sigma(- y_i \mathbf{w}^\top(t) \mathbf{h}_i) [\mathbf{P}^L]_{ij} \right) \mathbf{x}_j + \lambda \mathbf{w}(t). 
    \end{aligned}
\end{equation}

Then, according to the gradient descent update rule $\mathbf{w}(t+1) = \mathbf{w}(t) - \eta \nabla F(\mathbf{w}(t))$, we have
\begin{equation}\label{eq:recursive_weight}
    \begin{aligned}
     \mathbf{w}(t+1)  
    &=  (1-\eta \lambda) \times \mathbf{w}(t) - \eta \sum_{j\in\mathcal{V}} \left( \frac{1}{|\mathcal{V}_\text{train}|}\sum_{i\in\mathcal{V}_\text{train}} - y_i \sigma(- y_i \mathbf{w}^\top(t) \mathbf{h}_i) [\mathbf{P}^L]_{ij} \right) \mathbf{x}_j  \\
    &= \sum_{k=0}^t \eta (1-\eta\lambda)^{t-k} \left[ \sum_{j\in\mathcal{V}} \left( \frac{1}{|\mathcal{V}_\text{train}|}\sum_{i\in\mathcal{V}_\text{train}} - y_i \sigma(- y_i \mathbf{w}^\top(k) \mathbf{h}_i) [\mathbf{P}^L]_{ij} \right) \mathbf{x}_j \right]
    % &\underset{(a)}{\leq} \| \mathbf{w}(t) \|_2 + \eta B_x \cdot \max_{i\in \mathcal{V}_\text{train}} [\mathbf{P}^L]_{ij} \\
    % &\underset{(b)}{\leq} \| \mathbf{w}(t) \|_2 + \eta B_x
    \end{aligned}
\end{equation}
Then, we know that after $T$ iterations of gradient updates, for any $j \in\mathcal{V}$ we have
\begin{equation}
    \mathbf{w}(T) = \sum_{j\in\mathcal{V}} \beta_j \mathbf{x}_j,~\text{where}~
    \beta_j \leq \eta T \cdot \max_{i\in \mathcal{V}_\text{train}} [\mathbf{P}^L]_{ij} \leq \eta T,
\end{equation}
the first inequality is due to $\lambda\eta < 1, |y_i|=1, 0 <\sigma(x) < 1$ and the second inequality hold because any element $0 \leq [\mathbf{P}^L]_{ij} \leq 1$.

After projection, we find another set of $\{\alpha_i~|~i\in\mathcal{V}_\text{remain}\} $ and construct the unlearned weight parameter by $\mathbf{w}_p = \sum_{i\in\mathcal{V}_\text{remain}} \alpha_i \mathbf{x}_i$.
Then, the upper bound on $\| \mathbf{w}_p - \mathbf{w}(T)\|_2$ can be written as
\begin{equation}
    \begin{aligned}
    \| \mathbf{w}_p - \mathbf{w}(T)\|_2 
    &= \min_{\bm{\alpha} } \left\| \sum_{i\in\mathcal{V}_\text{remain}} \alpha_i \mathbf{x}_i - \sum_{i\in\mathcal{V}} \beta_i \mathbf{x}_i \right\|_2 \\
    &= \min_{\bm{\alpha} } \left\| \sum_{i\in\mathcal{V}_\text{remain}} (\alpha_i-\beta_i) \mathbf{x}_i - \sum_{i\in\mathcal{V}_\text{delete}} \beta_i \mathbf{x}_i \right\|_2 \\
    &\leq \sum_{i\in\mathcal{V}_\text{delete}} \beta_j \cdot \min_{\bm{\alpha} } \left\| \sum_{i\in\mathcal{V}_\text{remain}} \frac{\alpha_i-\beta_i}{\beta_j} \mathbf{x}_i -  \mathbf{x}_j \right\|_2 \\
    &\leq |\mathcal{V}_\text{delete}| \cdot \max_{j\in\mathcal{V}_\text{delete}} \beta_j \cdot \min_{\bm{\alpha} } \left\| \sum_{i\in\mathcal{V}_\text{remain}} \frac{\alpha_i-\beta_i}{\beta_j} \mathbf{x}_i -  \mathbf{x}_j \right\|_2 \\
    &\leq \eta T |\mathcal{V}_\text{delete}| \cdot \underbrace{\max_{j\in\mathcal{V}_\text{delete}}   \min_{\bm{\alpha} } \left\| \sum_{i\in\mathcal{V}_\text{remain}} \frac{\alpha_i-\beta_i}{\beta_j} \mathbf{x}_i -  \mathbf{x}_j \right\|_2}_{(a)}.
    % &\leq |\mathcal{V}_\text{delete}| \cdot \eta T \cdot \max_{i\in \mathcal{V}_\text{train}} [\mathbf{P}^L]_{ij} \cdot \min_{\bm{\gamma}} 
    \end{aligned}
\end{equation}
Notice that $(a)$ is equivalent to finding another set of coefficient $\bm{\gamma}$ that
\begin{equation}
    \max_{j\in\mathcal{V}_\text{delete}} \min_{\bm{\gamma}} \left\| \sum_{i\in\mathcal{V}_\text{remain}} \gamma_i \mathbf{x}_i - \mathbf{x}_j \right\|_2 \leq \delta,
\end{equation}
where the upper bound is due to Assumption~\ref{assumption:upper_bound_on_approximation_error}.
Therefore, we have
\begin{equation} ~\label{eq:projector_diff}
    \| \mathbf{w}_p - \mathbf{w}(T)\|_2 \leq \delta \eta T |\mathcal{V}_\text{delete}|.
\end{equation}

\subsection{Convergence rate for fune-tuning}\label{section:fine-tune-rate}
Let $\mathbf{w}_p(k)$ denote fune-tuning on $\mathbf{w}_p$ for $k$ iterations, where $\mathbf{w}_p(0) = \mathbf{w}_p$ and $\mathbf{w}_p(K) = \tilde{\mathbf{w}}_p$ as we used in Theorem~\ref{theorem:ell2_norm_of_weight_projections}..
By knowing $F^u$ is $(\lambda + B_x^2 P_s^2)$-smoothness, we have
\begin{equation}
    \begin{aligned}
    & F^u(\mathbf{w}_p(k+1)) \\
    &\underset{(a)}{\leq} F^u(\mathbf{w}_p(k)) + \langle \nabla F^u(\mathbf{w}_p(k)), \mathbf{w}_p(k+1) - \mathbf{w}_p(k)\rangle + \frac{(\lambda + B_x^2 P_s^2)}{2} \| \mathbf{w}_p(k+1) - \mathbf{w}_p(k) \|_2^2 \\
    &= F^u(\mathbf{w}_p(k)) - \eta \| \nabla F^u(\mathbf{w}_p(k)) \|_2^2 + \frac{\eta^2 (\lambda + B_x^2 P_s^2)}{2} \| \nabla F^u(\mathbf{w}_p(k)) \|_2^2 \\
    &= F^u(\mathbf{w}_p(k)) - \eta \left( 1 - \frac{\eta_k (\lambda + B_x^2 P_s^2)}{2} \right)  \| \nabla F^u(\mathbf{w}_p(k)) \|_2^2, 
    \end{aligned}
\end{equation}
where inequality $(a)$ is due to the update rule $\mathbf{w}_p(k+1) = \mathbf{w}_p(k) - \eta \nabla F^u(\mathbf{w}_p(k))$.

Let $\mathbf{w}_\star = \arg\min_\mathbf{w} F^u(\mathbf{w})$.
By choosing $\eta = \frac{2}{(\lambda + B_x^2 P_s^2)}$, we have
\begin{equation}\label{eq:convex_learn_rate_selection}
    \Big( F^u(\mathbf{w}_p(k+1)) - F^u(\mathbf{w}_\star) \Big) - \Big( F^u(\mathbf{w}_p(k)) - F^u(\mathbf{w}_\star) \Big) \leq - \frac{1}{2(\lambda + B_x^2 P_s^2)} \| \nabla F^u(\mathbf{w}_p(k)) \|_2^2.
\end{equation}

Since function $F^u$ is convex, we know the following inequality holds: 
\begin{equation}
    \begin{aligned}
    F^u(\mathbf{w}_p(k+1)) - F^u(\mathbf{w}_\star) 
    &\leq \langle \nabla F^u(\mathbf{w}_p(k)), \mathbf{w}_p(k) - \mathbf{w}_\star \rangle \\
    &\leq \| F^u(\mathbf{w}_p(k)) \|_2 \| \mathbf{w}_p(k) - \mathbf{w}_\star \|_2. 
    \end{aligned}
\end{equation}

By plugging it back to Eq.~\ref{eq:convex_learn_rate_selection}, we have
\begin{equation}
    \begin{aligned}
    & F^u(\mathbf{w}_p(k+1)) - F^u(\mathbf{w}_\star) \\
    &\quad \leq (F^u(\mathbf{w}_p(k)) - F^u(\mathbf{w}_\star)) \left( 1 - \frac{F^u(\mathbf{w}_p(k)) - F^u(\mathbf{w}_\star)}{2 (\lambda + B_x^2 P_s^2) \| \mathbf{w}_p(k) - \mathbf{w}_\star \|_2^2 }\right) \\
    &\quad\underset{(a)}{\leq} (F^u(\mathbf{w}_p) - F^u(\mathbf{w}_\star)) \left( 1 - \frac{F^u(\mathbf{w}_p(T)) - F^u(\mathbf{w}_\star)}{2 (\lambda + B_x^2 P_s^2) \| \mathbf{w}_p - \mathbf{w}_\star \|_2^2 }\right),
    \end{aligned}
\end{equation}
where inequality $(a)$ is due to $1 \leq k \leq K$. 
Since $F^u(\mathbf{w}_p(k)) - F^u(\mathbf{w}_\star) \leq 2(\lambda + B_x^2 P_s^2) \| \mathbf{w}_p - \mathbf{w}_\star \|_2$, we have
\begin{equation}
    \frac{1}{F^u(\mathbf{w}_p(k+1)) - F^u(\mathbf{w}_\star)} \geq \frac{1}{F^u(\mathbf{w}_p(k)) - F^u(\mathbf{w}_\star)} + \frac{1}{2(\lambda + B_x^2 P_s^2) \| \mathbf{w}_p - \mathbf{w}_\star \|_2^2 }.
\end{equation}

Telescoping from $k=T, \ldots, T+K$, we get 
\begin{equation}
    \frac{1}{F^u(\tilde{\mathbf{w}}_p) - F^u(\mathbf{w}_\star)} \geq \frac{1}{F^u(\mathbf{w}_p) - F^u(\mathbf{w}_\star)} + \frac{T}{2(\lambda + B_x^2 P_s^2) \| \mathbf{w}_p - \mathbf{w}_\star \|_2^2 },
\end{equation}
which implies
\begin{equation}\label{eq:almost_to_the_end}
    \frac{1}{F^u(\tilde{\mathbf{w}}_p) - F^u(\mathbf{w}_\star)} \leq \frac{2 (\lambda + B_x^2 P_s^2) (F^u(\mathbf{w}_p - F^u(\mathbf{w}_\star)) \| \mathbf{w}_p - \mathbf{w}_\star \|_2^2 }{T (F^u(\mathbf{w}_p - F^u(\mathbf{w}_\star)) + 2(\lambda + B_x^2 P_s^2) \| \mathbf{w}_p - \mathbf{w}_\star \|_2^2  }.
\end{equation}

By using the smoothness at $\mathbf{w}_\star$, we have
\begin{equation}
    F^u(\mathbf{w}_p) - F^u(\mathbf{w}_\star) \leq \frac{(\lambda + B_x^2 P_s^2)}{2} \| \mathbf{w}_p - \mathbf{w}_\star \|_2^2.
\end{equation}

Plugging back to Eq.~\ref{eq:almost_to_the_end}, suppose $T$ is large enough and $\mathbf{w}_u(T) \approx \mathbf{w}_\star$, we have
\begin{equation}
    \begin{aligned}
    F^u(\tilde{\mathbf{w}}_p) - F^u(\mathbf{w}_\star) &\leq \frac{2(\lambda + B_x^2 P_s^2)\| \mathbf{w}_p - \mathbf{w}_\star \|_2^2}{K+4} \\
    &\approx \mathcal{O}\left( \frac{(\lambda + B_x^2 P_s^2)\| \mathbf{w}_p - \mathbf{w}_u(T) \|_2^2}{K} \right).
    \end{aligned}
\end{equation}

\section{Proof on Proposition~\ref{proposition:closeness_of_projector_vs_inflence}}\label{section:proof_of_closeness_of_projector_vs_inflence}

The influence-based unlearning approach~\cite{guo2020certified} unlearn by using second-order gradient update on the weight parameters.
To apply~\cite{guo2020certified} onto a $L$-layer linear GNN, due to the node dependency, we have to unlearn all the $L$-hop neighbors of the deleted nodes.
Therefore, the generalization of~\cite{guo2020certified} to graph requires updating the weight parameters by
\begin{equation}\label{eq:proj_vs_second_order_1}
    \mathbf{w}_p = \mathbf{w}(T) - \left[\nabla^2 F(\mathbf{w}(T),~\mathcal{V}_\text{remain}^L)\right]^{-1} \nabla F(\mathbf{w}(T),~\mathcal{V}_\text{delete}^L),
\end{equation}
where $\mathcal{V}_\text{delete}^L = \text{unique}\{v_j~|~\text{SPD}(v_i,v_j)<L,~v_i\in\mathcal{V}_\text{delete}\}$ denotes the set of nodes that has shortest path distance (\text{SPD}) less than $L$ to nodes in $\mathcal{V}_\text{delete}$,  $\mathcal{V}_\text{remain}^L = \mathcal{V} \setminus \mathcal{V}_\text{delete}^L$, $\nabla^2 F(\mathbf{w},~\mathcal{V}_\text{remain}^L)$ denote computing the Hessain on $\mathcal{V}_\text{remain}^L$, and $\nabla F(\mathbf{w},~\mathcal{V}_\text{delete}^L)$ denote computing the gradient on $\mathcal{V}_\text{remain}^L$.

To prove Proposition~\ref{proposition:closeness_of_projector_vs_inflence}, we need to first analyze the upper bound on $\| \mathbf{w}_p - \mathbf{w} (T)\|_2$ for \textsc{Influence}~\cite{guo2020certified} by
\begin{equation}\label{eq:proj_vs_second_order_2}
    \begin{aligned}
    \| \mathbf{w}_p - \mathbf{w}(T) \|_2 
    &= \left\|  \left[\nabla^2 F(\mathbf{w}(T),~\mathcal{V}_\text{remain}^L)\right]^{-1} \nabla F(\mathbf{w}(T),~\mathcal{V}_\text{delete}^L) \right\|_2 \\
    &\leq \left\|  \left[\nabla^2 F(\mathbf{w}(T),~\mathcal{V}_\text{remain}^L)\right]^{-1} \right\|_2 \left\| \nabla F(\mathbf{w}(T),~\mathcal{V}_\text{delete}^L) \right\|_2.
    \end{aligned}
\end{equation}

Let us first upper bound $\left\| \nabla F(\mathbf{w}(T),~\mathcal{V}_\text{delete}^L) \right\|_2$ by
\begin{equation}\label{eq:proj_vs_second_order_3}
    \begin{aligned}
     &\left\| \nabla F(\mathbf{w}(T),~\mathcal{V}_\text{delete}^L) \right\|_2 \\
     &\quad = \left\| \sum_{j\in\mathcal{V}} \left( \frac{1}{|\mathcal{V}_\text{delete}|}\sum_{i\in\mathcal{V}_\text{delete}} - y_i \sigma(- y_i \mathbf{w}^\top(T) \mathbf{h}_i) [\mathbf{P}^L]_{ij} \right) \mathbf{x}_j + \lambda \mathbf{w}(T) \right\|_2 \\
     &\quad \leq \left\| \sum_{j\in\mathcal{V}} \left( \frac{1}{|\mathcal{V}_\text{delete}|}\sum_{i\in\mathcal{V}_\text{delete}} - y_i \sigma(- y_i \mathbf{w}^\top(T) \mathbf{h}_i) [\mathbf{P}^L]_{ij} \right) \mathbf{x}_j \right\|_2 + \lambda \| \mathbf{w}(T) \|_2 \\
     &\quad \underset{(a)}{\leq} B_x  |\mathcal{V}| + \lambda \| \mathbf{w}(T) \|_2, 
    \end{aligned}
\end{equation}
where $(a)$ is due to $|y_i| = 1,~0<\sigma(x)<1$, and $ [\mathbf{P}^L]_{ij} \leq 1$.
Meanwhile, from Eq.~\ref{eq:recursive_weight}, we can upper bound $\| \mathbf{w}(T) \|_2$ by
\begin{equation}
    \begin{aligned}
     \| \mathbf{w}(T) \|_2
    &= \left\| \sum_{k=0}^{T-1} \eta (1-\eta\lambda)^{T-1-k} \left[ \sum_{j\in\mathcal{V}} \left( \frac{1}{|\mathcal{V}_\text{train}|}\sum_{i\in\mathcal{V}_\text{train}} - y_i \sigma(- y_i \mathbf{w}^\top(k) \mathbf{h}_i) [\mathbf{P}^L]_{ij} \right) \mathbf{x}_j \right] \right\|_2 \\
    &\leq \eta T \times B_x |\mathcal{V}|
    % &\underset{(a)}{\leq} \| \mathbf{w}(t) \|_2 + \eta B_x \cdot \max_{i\in \mathcal{V}_\text{train}} [\mathbf{P}^L]_{ij} \\
    % &\underset{(b)}{\leq} \| \mathbf{w}(t) \|_2 + \eta B_x
    \end{aligned}
\end{equation}
By plugging the result back, we have
\begin{equation}
    \left\| \nabla F(\mathbf{w}(T),~\mathcal{V}_\text{delete}^L) \right\|_2 \leq (1+\lambda \eta T) B_x |\mathcal{V}|.
\end{equation}
Knowing that $\left\| \nabla^2 F(\mathbf{w}(T),~\mathcal{V}_\text{delete}^L) \right\|_2 > \lambda$ due to the strongly convexity of objective function $F(\mathbf{w})$, we have
\begin{equation}
    \left( \left\| \nabla^2 F(\mathbf{w}(T),~\mathcal{V}_\text{delete}^L) \right\|_2 \right)^{-1} \leq \frac{1}{\lambda}
\end{equation}

Therefore, we know that
\begin{equation}~\label{eq:influence_diff}
    \| \mathbf{w}_p - \mathbf{w}(T) \|_2 \leq \frac{(1+\lambda \eta T)B_x|\mathcal{V}|}{\lambda}.
\end{equation}
By comparing Eq.~\ref{eq:projector_diff} and Eq.~\ref{eq:influence_diff}, we know that if
\begin{equation}
    \delta  < \left(\frac{1}{\lambda \eta T} + 1 \right) B_x \times \frac{|\mathcal{V}|}{|\mathcal{V}_\text{delete}|}
\end{equation}
the solution of \our if provable closer to the retraining from scratch than~\cite{guo2020certified}.
Moreover, the above discussion also holds for~\cite{golatkar2020eternal} by replacing the variable $\mathcal{V}_\text{delete}$ in Eq.~\ref{eq:proj_vs_second_order_1}, ~\ref{eq:proj_vs_second_order_2}, and~\ref{eq:proj_vs_second_order_3} as $\mathcal{V}_\text{remain}$.

% \begin{equation}
% \frac{(1+\lambda \eta T)B_x|\mathcal{V}|}{\lambda}  = \delta \eta T |\mathcal{V}_\text{delete}|
% \end{equation}

\section{Proof of Proposition~\ref{prop:linear_as_expressive_as_nonlinear}} \label{section:linear_as_expressive}

The proof is an application of the proof of Theorem 4.1 in~\cite{wang2022powerful} to the linear-GNN structure $g_{\mathbf{w}}(\mathbf{L}, \mathbf{X}) = \sum_{\ell=1}^n  (\mathbf{P}^{\ell-1} \mathbf{X}) \mathbf{w}_\ell $ that we used in the experiment. 
Please notice that $g_{\mathbf{w}}(\mathbf{L}, \mathbf{X})$ is equivalent to first concatenating all polynomial graph convolutions then apply a single weight vector
\begin{equation*}
    g_{\mathbf{w}}(\mathbf{L}, \mathbf{X}) = [\mathbf{X}~||~\mathbf{P}\mathbf{X}~||\ldots||~\mathbf{P}^{n-1} \mathbf{X}] \mathbf{w},
\end{equation*}
where $[\mathbf{A}~||~\mathbf{B}] \in \mathbb{R}^{n\times 2d}$ is concatenating matrices $\mathbf{A},\mathbf{B}\in\mathbb{R}^{n\times d}$ along their feature dimension.
We assume $\mathbf{y} = f(\mathbf{P}, \mathbf{X}) \in \mathbb{R}^{n\times 1}$ as the target function we want to approximate by linear-GNN.

Let us define $\mathbf{U},\bm{\lambda}$ as the eigenvectors and eigenvalues of graph propagation matrix $\mathbf{P}$. Then, the linear-GNN could be written as
\begin{equation*}
    g_{\mathbf{w}}(\mathbf{L}, \mathbf{X}) = \sum_{\ell=1}^n (\mathbf{P}^{\ell-1} \mathbf{X})\mathbf{w}_\ell = \sum_{\ell=1}^n \mathbf{U} \Lambda^{\ell-1} \mathbf{U} \mathbf{X} \mathbf{w}_\ell,~\text{where}~[\Lambda^{\ell-1}]_{i,i} = \lambda^{\ell-1}_i
\end{equation*}

Since we assume all rows in $\mathbf{U} \mathbf{X}$ is non-zero vectors, we know that there always exists a set of $\mathbf{w}_\ell,~\forall \ell\in\{1,\ldots,k\}$ such that all elements in $\mathbf{U} \mathbf{X} \mathbf{w}_\ell,~\forall \ell\in\{1,\ldots,k\}$ is non-zero. For example, we can select $\mathbf{w}_\ell$ from $\mathbb{R}^d$ that is not orthogonal to all vectors in $\mathbf{U}\mathbf{X}$ and not equal to zero vector.
Let us denote $\mathbf{z}_\ell = \mathbf{U}\mathbf{X} \mathbf{w}_\ell \in \mathbb{R}^{n\times 1}$, then our linear-GNN could be written as
\begin{equation*}
    g_{\mathbf{w}}(\mathbf{L}, \mathbf{X}) = \sum_{\ell=1}^n \mathbf{U} \left(\bm{\lambda}^{\ell-1} \cdot \mathbf{z}_\ell \right) = \mathbf{U} \left( \sum_{\ell=1}^n \left(\bm{\lambda}^{\ell-1} \cdot \mathbf{z}_\ell \right) \right),
\end{equation*}
where $\mathbf{a}\cdot \mathbf{b}$ is element-wise dot product between $\mathbf{a},\mathbf{b}$.

In order to use $g_{\mathbf{w}}(\mathbf{L}, \mathbf{X})$ approximate any function $\mathbf{y} = f(\mathbf{P}, \mathbf{X}) \in \mathbb{R}^{n\times 1}$, we have to make sure
\begin{equation*}
    \mathbf{U} \left( \sum_{\ell=1}^n \left(\bm{\lambda}^{\ell-1} \cdot \mathbf{z}_\ell \right) \right) = \mathbf{y} \rightarrow \left( \sum_{\ell=1}^n \left(\bm{\lambda}^{\ell-1} \cdot \mathbf{z}_\ell \right) \right) = \mathbf{U}^\top \mathbf{y},
\end{equation*}

Let us write as $\mathbf{w}_\ell = \mathbf{w} \cdot \alpha_\ell$, then we have the following equality 
\begin{equation*}
    \underbrace{\begin{bmatrix}
        1 & \lambda_1 & \lambda_1^2 & \ldots & \lambda_1^{n-1} \\ 
        1 & \lambda_2 & \lambda_2^2 & \ldots & \lambda_2^{n-1} \\ 
        \vdots & \vdots & \vdots & \vdots & \vdots \\ 
        1 & \lambda_n & \lambda_n^2 & \ldots & \lambda_n^{n-1} 
    \end{bmatrix}}_{\mathbf{M} \in \mathbb{R}^{n\times n}} 
    \begin{bmatrix}
        \alpha_1 \\ 
        \alpha_2 \\ 
        \vdots \\ 
        \alpha_n
    \end{bmatrix} = 
    \begin{bmatrix}
        [\mathbf{U}^\top\mathbf{y}]_1 / [\mathbf{U} \mathbf{X} \mathbf{w}]_1 \\ 
        [\mathbf{U}^\top\mathbf{y}]_2 / [\mathbf{U} \mathbf{X} \mathbf{w}]_2 \\ 
        \vdots \\ 
        [\mathbf{U}^\top\mathbf{y}]_n / [\mathbf{U} \mathbf{X} \mathbf{w}]_n
    \end{bmatrix}
\end{equation*}

If no elements in $\bm{\lambda}$ is identical, $\mathbf{M}$ is inversible and there is always exists a unique set of $\alpha_i,~i\in\{1,\ldots,n\}$ that satisfy the above equality.

\section{Connection and potential application to SVM unlearning} \label{section:connection_to_svm}

The KKT-based unlearning has been studied in the SVM unlearning~\cite{cauwenberghs2000incremental,karasuyama2009multiple,galmeanu2008implementation,diehl2003svm} dated back to the last two decades. Due to the great similarity between SVM and logistic regression, it is interesting to compare it with our projection-based unlearning. Although generalizing the KKT-based unlearning from SVM  to logistic regression is non-trivial, the other way around is possible according to Proposition~\ref{prop:svm_linear_span}. 

\begin{proposition} \label{prop:svm_linear_span}
When training SVM using primal gradient descent with the initial solution $\mathbf{w}_0 \in \text{span}\{\mathbf{x}_1, \ldots, \mathbf{x}_N\}$ (e.g., Pegasos~\cite{shalev2011pegasos}) or using dual coordinate ascent (e.g.,SMO~\cite{platt1998sequential}), the corresponding primal solution after $t$ iterations always satisfy $\mathbf{w}_t\in \text{span}\{\mathbf{x}_1, \ldots, \mathbf{x}_N\}$.
\end{proposition}

Therefore, we limit our following discussion to linear SVM unlearning, where the primal objective function is defined as 
\begin{equation}
    \smash{f_\text{SVM}(\mathbf{w})=\frac{\lambda}{2}\| \mathbf{w}\|_2^2 + \sum_{i=1}^N \max(0, 1-y_i \mathbf{w}^\top \mathbf{x}_i)},
\end{equation}
and its dual objective function is defined as 
\begin{equation}
    \begin{aligned}
    &g_\text{SVM}(\bm{\alpha}) =  - \frac{1}{2\lambda} \sum_{i=1}^N \sum_{j=1}^N \alpha_i\alpha_j y_i y_j \langle \mathbf{x}_i, \mathbf{x}_j \rangle + \sum_{i=1}^N \alpha_i, \\
    &\text{subject to}~~\sum_{i=1}^N \alpha_i y_i = 0,~ \alpha_i \in [0, 1].
    \end{aligned}
\end{equation}

\subsection{Existing SVM unlearning and its limitation}
SVM unlearning~\cite{cauwenberghs2000incremental} investigates how to maintain the KKT optimality condition when data are slightly changed. They propose to quickly identify the dual variable $\alpha_i$ for each data point $\mathbf{x}_i$ and solve the linear system that maintain the KKT optimality condition.
In practice, they require maintaining a $N\times N$ kernel matrix for enlarging or shrinking, which is memory consuming and engineering effort prohibitive when $N$ is large~\cite{laskov2006incremental}. 
Moreover, their method requires multiple iterations to unlearn a single data point, which could potentially be inefficient.
To see this, let us first classify all data points into three categories according to its geometric position to the marginal hyperplanes:
\begin{itemize}
    \item Outside the marginal hyperplanes $\mathcal{O} = \{i~|~y_i \mathbf{w}_\star^\top \mathbf{x}_i > 1,~\alpha_i = 0\}$, 
    \item On the marginal hyperplanes $\smash{\mathcal{M} = \{i~|~y_i \mathbf{w}_\star^\top \mathbf{x}_i = 1,~ 0\leq \alpha_i \leq 1 \}}$,
    \item Between the marginal hyperplanes $\smash{\mathcal{I} = \{i~|~y_i \mathbf{w}_\star^\top \mathbf{x}_i < 1,~\alpha_i = 1\}}$.
\end{itemize}
Besides, according to the KKT condition, the optimal primal solution $\mathbf{w}_\star$ and its prediction on any data point $\hat{y}_i$ can be expanded as 
\begin{equation}
    \mathbf{w}_\star = \frac{1}{\lambda}\sum_{i=1}^N \alpha_i y_i \mathbf{x}_i
    ~\text{and}~
    \hat{y}_i = \mathbf{w}_\star^\top \mathbf{x}_i = \frac{1}{\lambda}\sum_{j \in \mathcal{M}\cup \mathcal{I}} \alpha_j y_j \langle \mathbf{x}_i, \mathbf{x}_j \rangle.
\end{equation}
To delete data point $(\mathbf{x}_c, y_c)$, \cite{cauwenberghs2000incremental} needs to decrease the corresponding Lagrangian parameters $\alpha_c$ to $0$, meanwhile keep the optimal conditions of other parameters satisfied, i.e., for any $i \in \mathcal{M}$ we have 
\begin{equation}
    \Delta \alpha_c y_c \langle \mathbf{x}_i, \mathbf{x}_c \rangle + \sum\nolimits_{j\in\mathcal{M}} \Delta \alpha_j y_j \langle \mathbf{x}_i, \mathbf{x}_j \rangle= 0,
\end{equation}
where  $\Delta \alpha_j$ denote the amount of the change of variable $\alpha_j$.
The update direction of each dual variable can be obtained by solving the linear system with respect to each $\Delta\alpha_j$. The update step size is selected as the largest step length under the condition that no element moves across $\mathcal{M}, \mathcal{O}$, and $ \mathcal{I}$.
When any $\alpha_j,~\forall j\in\mathcal{M} \cup \{c\}$ is increased to $1$ or decreased to $0$, we have to move one point from one set to another, and repeat the above process multiple iterations until stable.
Since every iteration only one element is moving across $\mathcal{M}, \mathcal{O}$, and $ \mathcal{I}$, we have to solve $|\mathcal{M}|$ linear equations multiple iterations\footnote{However, the number of iterations is unknown, which could be extremely large when comparing to $|\mathcal{M}|$.}, each of which requires to inverse a $|\mathcal{M}|\times|\mathcal{M}|$ matrix, which could result in computation overhead of $\mathcal{O}(|\mathcal{M}|^3)$ for unknown number of iterations, which is not guaranteed to be faster than re-training the data from scratch~\cite{tsai2014incremental}.

\subsection{\our for SVM unlearning}
As an alternative, under the assumption that slightly dataset change only cause minor change on the optimal weight parameters, 
% which is often the cause if the number of data to be deleted is small,  
we propose to apply \our directly to primal solution of SVM and then fine-tune for several iterations using gradient descent methods (e.g., Pegasos~\cite{shalev2011pegasos}).
% Then, the unlearned solution is considered as an good initial solution that close to the optimal solution that reduces the number of training iterations.
We note that our primal SVM unlearning method shares the same spirit with~\cite{tsai2014incremental}, in which they propose an approximate unlearning method that directly finetune the primal solution on the new dataset (without the deleted data points), therefore the sensitive information are not guaranteed to be perfectly removed.  In contrast, our projection-based method could provide such guarantee for~\cite{tsai2014incremental}.
We leave this an an interesting future direction which could explore the idea in the future.

% \clearpage
% \input{supplementary/discussion_on_privacy}

% \input{supplementary/linear_as_expressive_as_nonlinear}
\end{document}